\documentclass[10pt,journal,compsoc]{IEEEtran}
\usepackage{graphicx}
\usepackage{amsmath,amssymb}

\usepackage{cite}

\usepackage[table]{xcolor}
\usepackage{amsthm}

\newtheorem{theorem}{Theorem}
\newtheorem{proposition}{Proposition}

\providecommand{\eref}[1]{\eqref{#1}}  
\providecommand{\cref}[1]{Chapter~\ref{#1}}

\providecommand{\fref}[1]{Figure~\ref{#1}}
\providecommand{\tref}[1]{Table~\ref{#1}}

\providecommand{\R}{\ensuremath{\mathbb{R}}}

\providecommand{\bydef}{\overset{\text{def}}{=}}

\renewcommand{\vec}[1]{\ensuremath{\boldsymbol{#1}}}
\providecommand{\mat}[1]{\ensuremath{\boldsymbol{#1}}}

\providecommand{\calF}{\mathcal{F}}
\providecommand{\calG}{\mathcal{G}}

\providecommand{\calI}{\mathcal{I}}

\providecommand{\calT}{\mathcal{T}}

\providecommand{\mB}{\mat{B}}

\providecommand{\mD}{\mat{D}}

\providecommand{\mG}{\mat{G}}
\providecommand{\mH}{\mat{H}}
\providecommand{\mI}{\mat{I}}

\providecommand{\mL}{\mat{L}}

\providecommand{\mP}{\mat{P}}

\providecommand{\mT}{\mat{T}}

\providecommand{\va}{\vec{a}}
\providecommand{\vb}{\vec{b}}

\providecommand{\vr}{\vec{r}}

\providecommand{\vu}{\vec{u}}
\providecommand{\vv}{\vec{v}}

\providecommand{\vx}{\vec{x}}
\providecommand{\vy}{\vec{y}}
\providecommand{\vz}{\vec{z}}

\providecommand{\mDelta}{\mat{\Delta}}

\providecommand{\mSigma}{\mat{\Sigma}}

\providecommand{\valpha}{\vec{\alpha}}

\providecommand{\vmu}{\vec{\mu}}

\providecommand{\Jtilde}{\widetilde{J}}

\providecommand{\mDtilde}{\mat{\widetilde{D}}}

\providecommand{\mLtilde}{\mat{\widetilde{L}}}

\providecommand{\vzero}{\vec{0}}
\providecommand{\vone}{\vec{1}}

\newcommand{\subjectto}{\mathop{\mathrm{subject\, to}}}
\newcommand{\argmin}[1]{\mathop{\underset{#1}{\mbox{argmin}}}}

\newcommand{\minimize}[1]{\mathop{\underset{#1}{\mathrm{minimize}}}}

\usepackage{amssymb}
\usepackage{pdfpages}
\usepackage{enumitem}
\usepackage{bbm}
\usepackage{multirow}
\usepackage{algorithm,algpseudocode}

\usepackage{array}
\newcommand{\PreserveBackslash}[1]{\let\temp=\\#1\let\\=\temp}
\newcolumntype{C}[1]{>{\PreserveBackslash\centering}p{#1}}
\newcolumntype{R}[1]{>{\PreserveBackslash\raggedleft}p{#1}}
\newcolumntype{L}[1]{>{\PreserveBackslash\raggedright}p{#1}}

\begin{document}

\title{Automatic Foreground Extraction from Imperfect Backgrounds using Multi-Agent Consensus Equilibrium}
\author{Xiran Wang,~\IEEEmembership{Student Member,~IEEE}, Jason Juang and Stanley H. Chan,~\IEEEmembership{Senior Member,~IEEE}
\thanks{X. Wang and S. Chan are with the School of Electrical and Computer Engineering, Purdue University, West Lafayette, IN 47907, USA. Email: \texttt{\{ wang470, stanchan\}@purdue.edu}. This work is supported, in part, by the National Science Foundation under grants CCF-1763896 and CCF-1718007.}
\thanks{J. Juang is with HypeVR Inc., San Diego, CA 92108, USA. Email: \texttt{jason@hypevr.com}}
}
\graphicspath{{./pix/}}

\IEEEtitleabstractindextext{\begin{abstract}
Extracting accurate foreground objects from a scene is an essential step for many video applications. Traditional background subtraction algorithms can generate coarse estimates, but generating high quality masks requires professional softwares with significant human interventions, e.g., providing trimaps or labeling key frames. We propose an automatic foreground extraction method in applications where a static but imperfect background is available. Examples include filming and surveillance where the background can be captured before the objects enter the scene or after they leave the scene. Our proposed method is very robust and produces significantly better estimates than state-of-the-art background subtraction, video segmentation and alpha matting methods. The key innovation of our method is a novel information fusion technique. The fusion framework allows us to integrate the individual strengths of alpha matting, background subtraction and image denoising to produce an overall better estimate. Such integration is particularly important when handling complex scenes with imperfect background. We show how the framework is developed, and how the individual components are built. Extensive experiments and ablation studies are conducted to evaluate the proposed method.
\end{abstract}

}

\maketitle

\section{Introduction}
Extracting accurate foreground objects is an essential step for many video applications in filming, surveillance, environment monitoring and video conferencing \cite{Bouwmans_Garcia_2020,Bouwmans_Porikli_2014}, as well as generating ground truth for performance evaluation \cite{Wang_Luo_Jodoin_2017}. As video technology improves, the volume and resolution of the images have grown significantly over the past decades. Manual labeling has become increasingly difficult; even with the help of industry-grade production softwares, e.g., NUKE, producing high quality masks in large volume is still very time-consuming. The standard solution in the industry has been chroma-keying \cite{Shimoda_Hayashi_Kanatsugu_1989} (i.e., using green screens). However, setting up green screens is largely limited to indoor filming. When moving to outdoor, the cost and manpower associated with the hardware equipment is enormous, not to mention the uncertainty of the background lighting. In addition, some dynamic scenes prohibit the usage of green screens, e.g., capturing a moving car on a road. Even if we focus solely on indoor filming, green screens still have limitations for 360-degree cameras where cameras are surrounding the object. In situations like these, it is unavoidable that the equipments become part of the background.

\begin{figure*}[!]
\centering
\includegraphics[width=\linewidth]{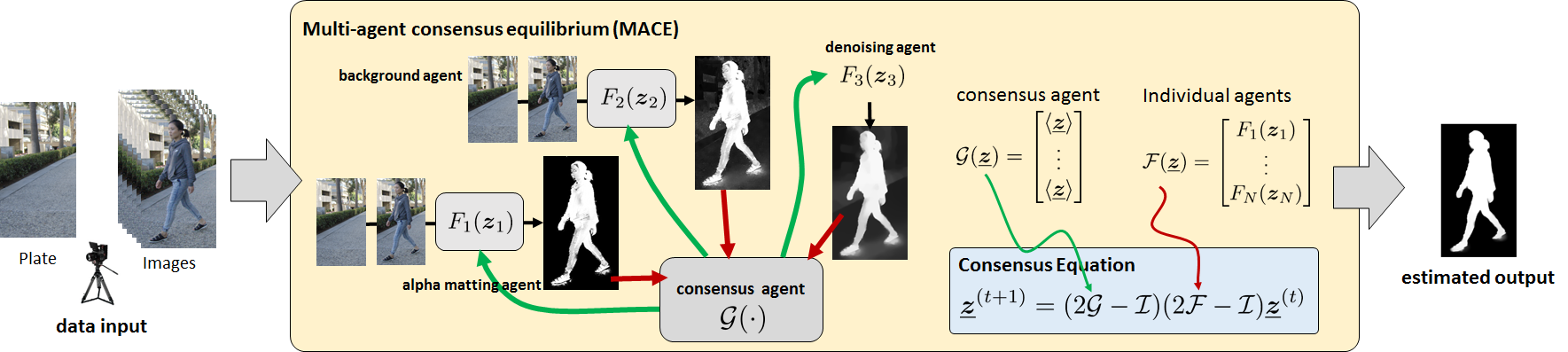}
\caption{Conceptual illustration of the proposed MACE framework. Given the input image and the background plate, the three MACE agents individually computes their estimates. The consensus agent aggregates the information and feeds back to the agents to ask for updates. The iteration continues until all the agents reach a consensus which is the final output.}
\label{fig: MACE illustration}
\end{figure*}

This paper presents an alternative solution for the aforementioned foreground extraction task. Instead of using a green screen, we assume that we have captured the background image before the object enters or after it leaves the scene. We call this background image as the \emph{plate} image. Using the plate image is significantly less expensive than using the green screens. However, plate images are never perfect. The proposed method is to robustly extract the foreground masks in the presence of imperfect plate images.

\subsection{Main Idea}
The proposed method, named Multi-Agent Consensus Equilibrium (MACE), is illustrated in \fref{fig: MACE illustration}. At the high level, MACE is an information fusion framework where we construct a strong estimator from several weak estimators. By calling individual estimators as \emph{agents}, we can imagine that the agents are asserting forces to pull the information towards themselves. Because each agent is optimizing for itself, the system is never stable. We therefore introduce a \emph{consensus agent} which averages the individual signals and broadcast the feedback information to the agents. When the individual agents receive the feedback, they adjust their internal states in order to achieve an overall equilibrium. The framework is theoretically guaranteed to find the equilibrium state under mild conditions.

In this paper, we present a novel design of the agents, $F_1, F_2, F_3$. Each agent has a different task: Agent 1 is an alpha mating agent which takes the foreground background pair and try to estimate the mask using the alpha matting equation. However, since alpha matting sometimes has false alarms, we introduce Agent 2 which is a background estimation agent. Agent 2 provides better edge information and hence the mask. The last agent, Agent 3, is a denoising agent which promotes smoothness of the masks so that there will not be isolated pixels.

As shown in \fref{fig: MACE illustration}, the collection of the agents is the operator $\calF$ which takes some signal and updates it. There is $\calG$ which is the consensus agent. Its goal is to take averages of the incoming signals and broadcast the average as a feedback. The red arrows represent signals flowing into the consensus agent, and the green arrows represent signals flowing out to the individual agents. The iteration is given by the consensus equation. When the iteration terminates, we obtain the final estimate.

\subsection{Contributions}
While most of the existing methods are based on developing a single estimator, this paper proposes to use a consensus approach. The consensus approach has several advantages, which correspond to the contributions of this paper.

\begin{itemize}
\item MACE is a training-free approach. We do not require training datasets like those deep learning approaches. MACE is an optimization-based approach. All steps are transparent, explainable, interpretable, and can be debugged.
\item MACE is fully automatic. Unlike classical alpha matting algorithms where users need to feed manual scribbles, MACE does not require human in the loop. As will be shown in the experiments, MACE can handle a variety of imaging conditions, motion, and scene content.
\item MACE is guaranteed to converge under mild conditions which will be discussed in the theory section of this paper.
\item MACE is flexible. While we propose three specific agents for this problem and we have shown their necessity in the ablation study, the number and the type of agents can be expanded. In particular, it is possible to use deep learning methods as agents in the MACE framework.
\item MACE offers the most robust result according to the experiments conducted in this paper. We attribute this to the complementarity of the agents when handling difficult situations.
\end{itemize}

In order to evaluate the proposed method, we have compared against more than 10 state-of-the-art video segmentation and background subtraction methods, including several deep neural network solutions. We have created a database with ground truth masks and background images. The database will be release to the general public on our project website. We have conducted an extensive ablation study to evaluate the importance of individual components.

\section{Background}
\subsection{Related Work}

\begin{figure*}[!]
\centering
\includegraphics[width=\linewidth]{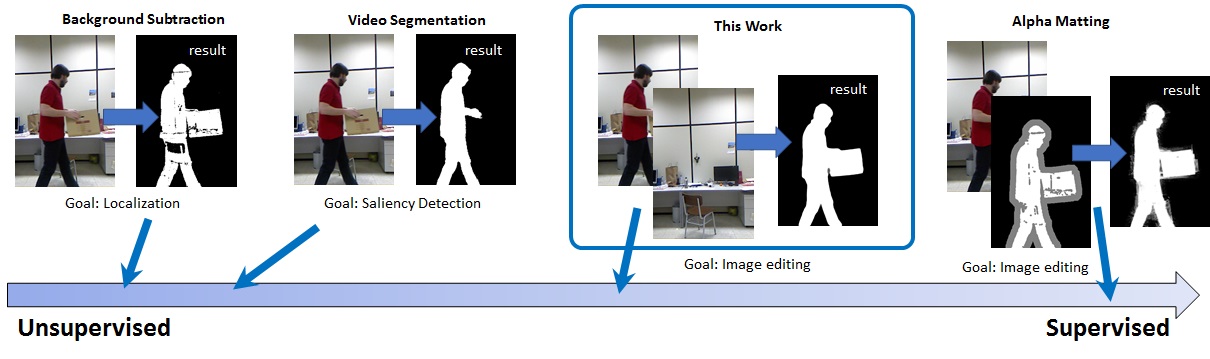}

\vspace{-4ex}
\caption{The landscape of the related problems. We group the methods according to the level of supervision they need during the inference stage. Background Subtraction (e.g., \cite{Hofmann_Tiefenbacher_Rigoll_2012}) aims at locating objects in a video, and most of the available methods are un-supervised. Video Segmentation (e.g., \cite{Wang_Song_Zhao_2019}) aims at identifying the salient objects. The training stage is supervised, but the inference stage is mostly un-supervised. Alpha matting (e.g., \cite{Cho_Tai_Kweon_2016}) needs a high-quality trimap to refine the uncertain regions of the boundaries, making it a supervised method. Our proposed method is in the middle. We do not need a high-quality trimap but we assume an imperfect background image.}
\label{fig: comparison with existing}
\vspace{-2ex}
\end{figure*}

Extracting foreground masks has a long history in computer vision and image processing. Survey papers in the field are abundant, e.g., \cite{Bouwmans_2011, Javed_2018, Bouwmans_Javed_Sultana_2019}. However, there is a subtle but important difference between the problem we are studying in this paper with the existing literature, as illustrated in \fref{fig: comparison with existing} and Table~\ref{table: comparison with existing}. At the high level, these differences can be summarized in three forms: (1) Quality of the output masks. (2) Requirement of the input (3) Automation and supervision. We briefly describe the comparison with existing methods.

\begin{itemize}[leftmargin=*]
\item \textbf{Alpha Matting}. Alpha matting is a supervised method. Given a user labeled ``trimap'', the algorithm uses a linear color model to predict the likelihood of a pixel being foreground or background as shown in \fref{fig: comparison with existing}. A few better known examples include Poisson matting \cite{Sun_Jia_Tang_2004}, closed-form matting \cite{Levin_Lischinski_Weiss_2008}, shared matting \cite{Gastal_Oliveira_2010}, Bayesian matting \cite{Chuang_Curless_Salesin_2001}, and robust matting \cite{Wang_Cohen_2007}. More recently, deep neural network based approaches are proposed, e.g., \cite{Xu_Price_Huang_2017}, \cite{Cho_Tai_Kweon_2016}, \cite{Tang_Aksoy_Oztireli_2019} and \cite{Xi_Chen_Qian_2019}.
    The biggest limitation of alpha matting is that the trimaps have to be \emph{error-free}. As soon there is a false alarm of miss in the trimap, the resulting mask will be severely distorted. In video setting, methods such as \cite{Marki_Perazzi_Wang_2016,Ramakanth_Babu_2014,Jain_Grauman_2014} suffer similar issues of error-prone trimaps due to temporal propagation. Two-stage methods such as \cite{Hsieh_Lee_2013} requires initial segmentation \cite{Rother_Kolmogorov_Blake_2004} to provide the trimap and suffer the same problem.  Other methods \cite{Cho_Yamasaki_Aizawa_2011,Wang_Finger_Yang_2007} require additional sensor data, e.g., depth, which is not always available.

    \vspace{2ex}
\item \textbf{Background Subtraction}. Background subtraction is unsupervised. Existing background subtraction methods range from the simple frame difference method to the more sophisticated mixture models \cite{Zivkovic_2004}, contour saliency \cite{Davis_Sharma_2005}, dynamic texture \cite{Mahadevan_Vasconcelos_2008}, feedback models \cite{Hofmann_Tiefenbacher_Rigoll_2012} and attempts to unify several approaches \cite{Barnich_Marc_2011}. Most background subtraction methods are used to track objects instead of extracting the alpha mattes. They are fully-automated and are real time, but the foreground masks generated are usually of low quality.

    It is also important to mention that there work about initializing the background image. These methods are particularly useful when the pure background plate image is not available. Bouwmans et al. has a comprehensive survey about the approaches \cite{Bouwmans_Maddalena_Petrosino_2017}. In addition, a number of methods should be noted, e.g., \cite{Colombari_2006, Javed_Mahmood_Bouwmans_2018, Javed_Bouwmans_Jung_2017, Sultana_Mahmood_Javed_2019, Laugraud_2018, Laugraud_2017}.

    \vspace{2ex}
\item \textbf{Unsupervised Video Segmentation}. Unsupervised video segmentation, as it is named, is unsupervised and fully automatic. The idea is to use different saliency cues to identify objects in the video, and then segments them out. Early approaches include key-segments \cite{Lee_Grauman_2011}, graph model \cite{Ma_Latecki_2012}, contrast based saliency \cite{Perazzi_Krahenbuhl_Pritch_2012}, motion cues \cite{Papazouglou_Ferrari_2013}, non-local statistics \cite{Faktor_Irani_2014}, co-segmentation \cite{Wang_Shen_Li_2015}, convex-optimization \cite{Jang_Lee_Kim_2016}. State-of-the-art video segmentation methods are based on deep neural networks, such as using short connection \cite{Hou_Cheng_Hu_2017}, pyramid dilated networks \cite{Song_Wang_Zhao_2018}, super-trajectory \cite{Wang_Shen_Porikli_2018}, video attention \cite{Wang_Song_Zhao_2019}, and feature pyramid \cite{Dong_Gao_Sun_2018, Dong_Gao_Pirbhulal_2020}. Readers interested in the latest developments of video segmentation can consult the tutorial by Wang et al. \cite{Wang_Shen_Xie_2019}. Online tools for video segmentation are also available \cite{Dehghan_Zhang_Siam_2019}. One thing to note is that most of the deep neural network solutions require post-processing methods such as conditional random field \cite{Krahenbuhl_Koltun_2011} to fine tune the masks. If we directly use the network output, the results are indeed not good. In contrast, our method does not require any post-processing.
\end{itemize}

\begin{table}[t]
\caption{Objective and Assumptions of (a) Alpha matting, (b) Background Subtraction, (c) Unsupervised Video Segmentation, and (d) Our work.}
\scriptsize{
\begin{tabular}{L{1cm}C{1.2cm}C{1.5cm}C{1.2cm}C{1.2cm}}
\hline
\hline
 Method       & Alpha  & Bkgnd   &Video &Ours\\
              & Matting        & Subtraction        & Segmentation\\
\hline
  Goal        & foreground     & object             &saliency              &foreground\\
              & estimation     & detection          &detection             &estimation\\
\hline
  Input       &  image+trimap  & video              &video                 &image+plate\\
  Accuracy    &  high          & low (binary)       &medium                &high\\
  Automatic   &  semi          & full               &full                  &full\\
  Supervised  &  semi          & no                 &no                    &semi\\
   \hline
\end{tabular}}
\label{table: comparison with existing}
\end{table}

Besides the above mentioned papers, there are some existing methods using the consensus approach, e.g., the early work of Wang and Suter \cite{Wang_Suter_2006}, Han et al. \cite{Han_Cai_Wang_2013}, and more recently the work of St. Charles et al. \cite{Charles_Bilodeau_Begevin_2016}. However, the notion of consensus in these papers are more about making votes for the mask. Our approach, on the other hand, are focusing on information exchange across agents. Thus, we are able to offer theoretical guarantees whereas the existing consensus approaches are largely rule-based heuristics.

\begin{figure*}[!]
\centering
\begin{tabular}{ccccccc}
I & \includegraphics[width=0.14\linewidth]{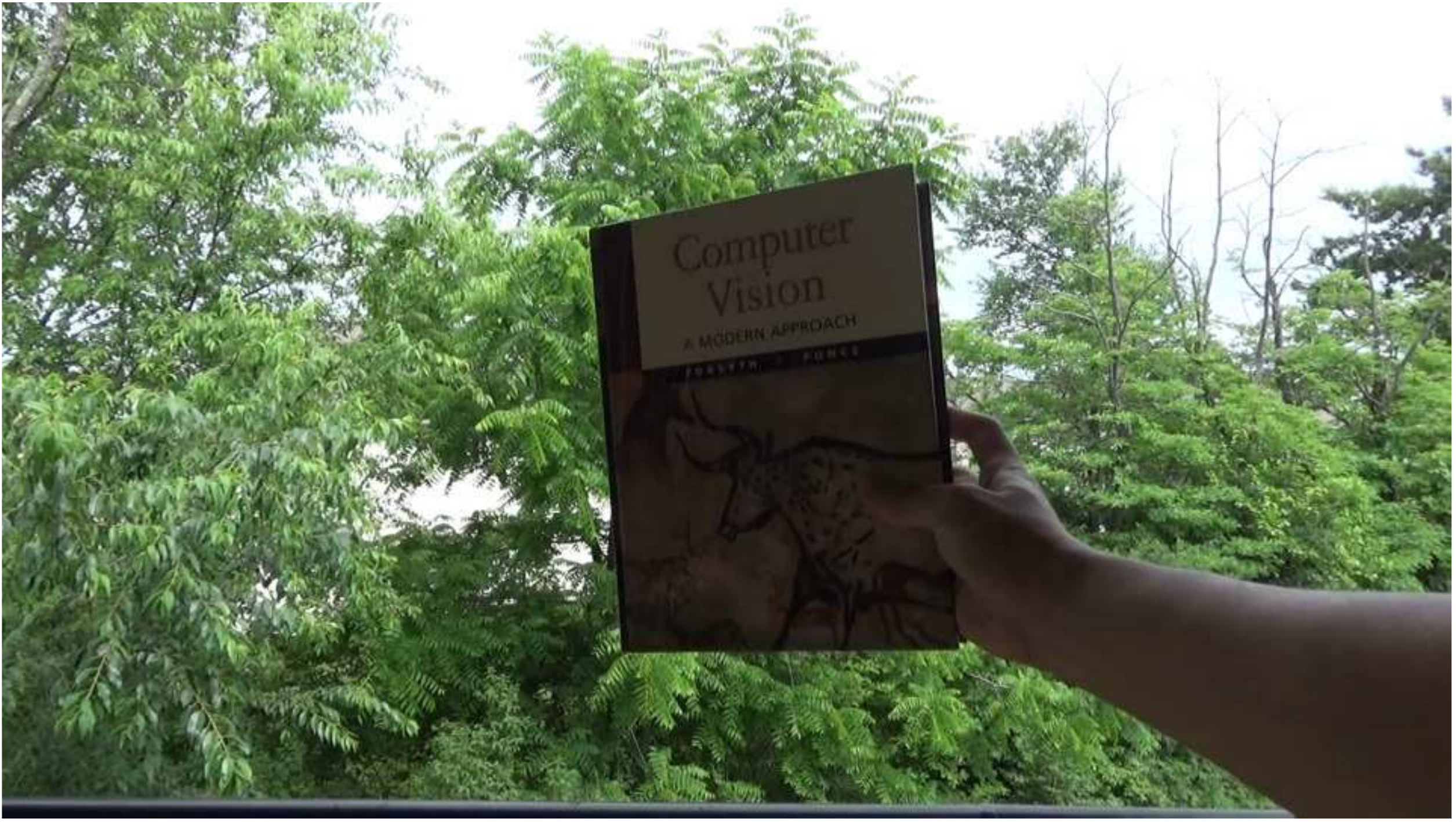}&
\hspace{-2ex}
\includegraphics[width=0.14\linewidth]{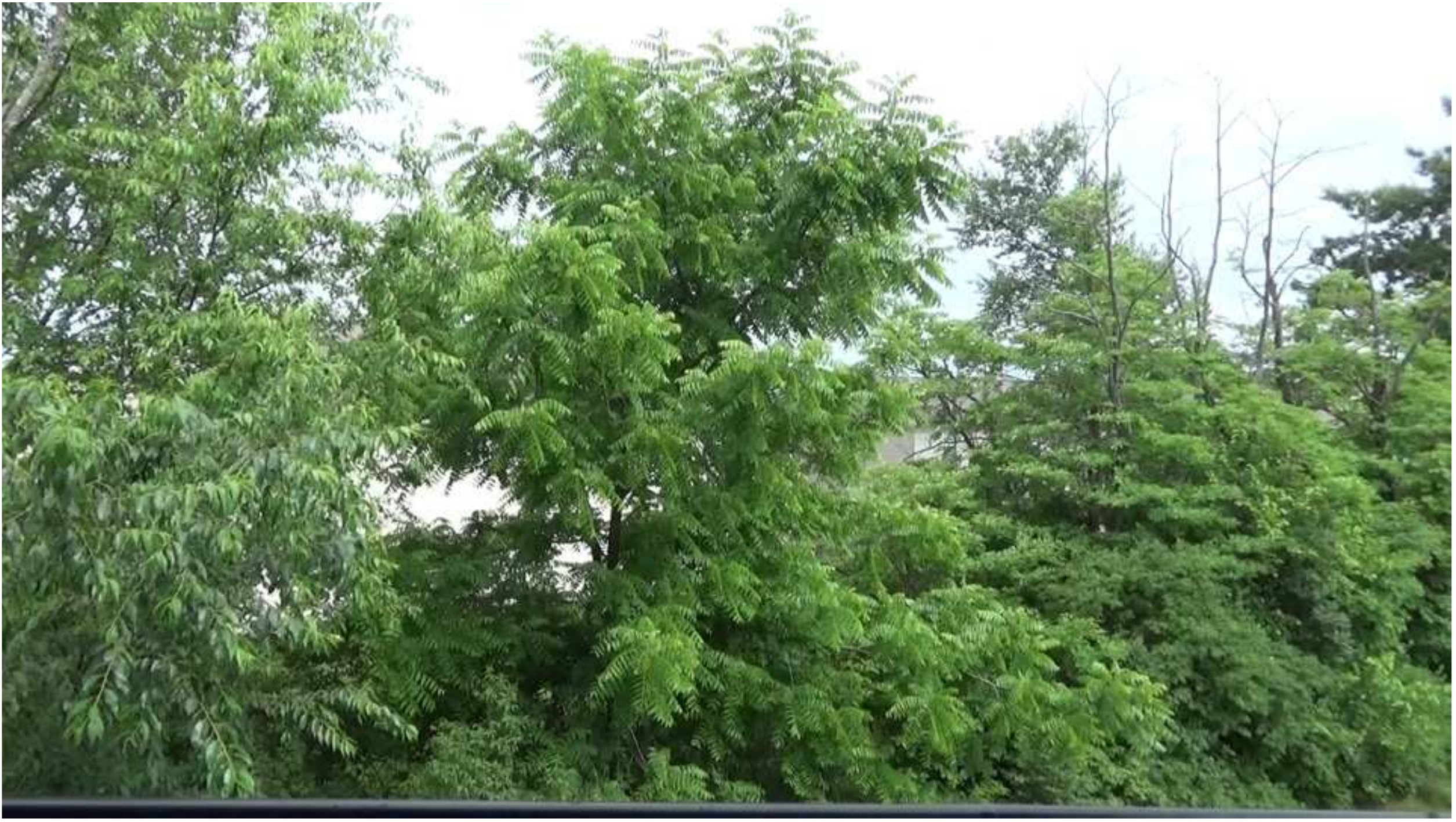}&
\hspace{-2ex}
\includegraphics[width=0.14\linewidth]{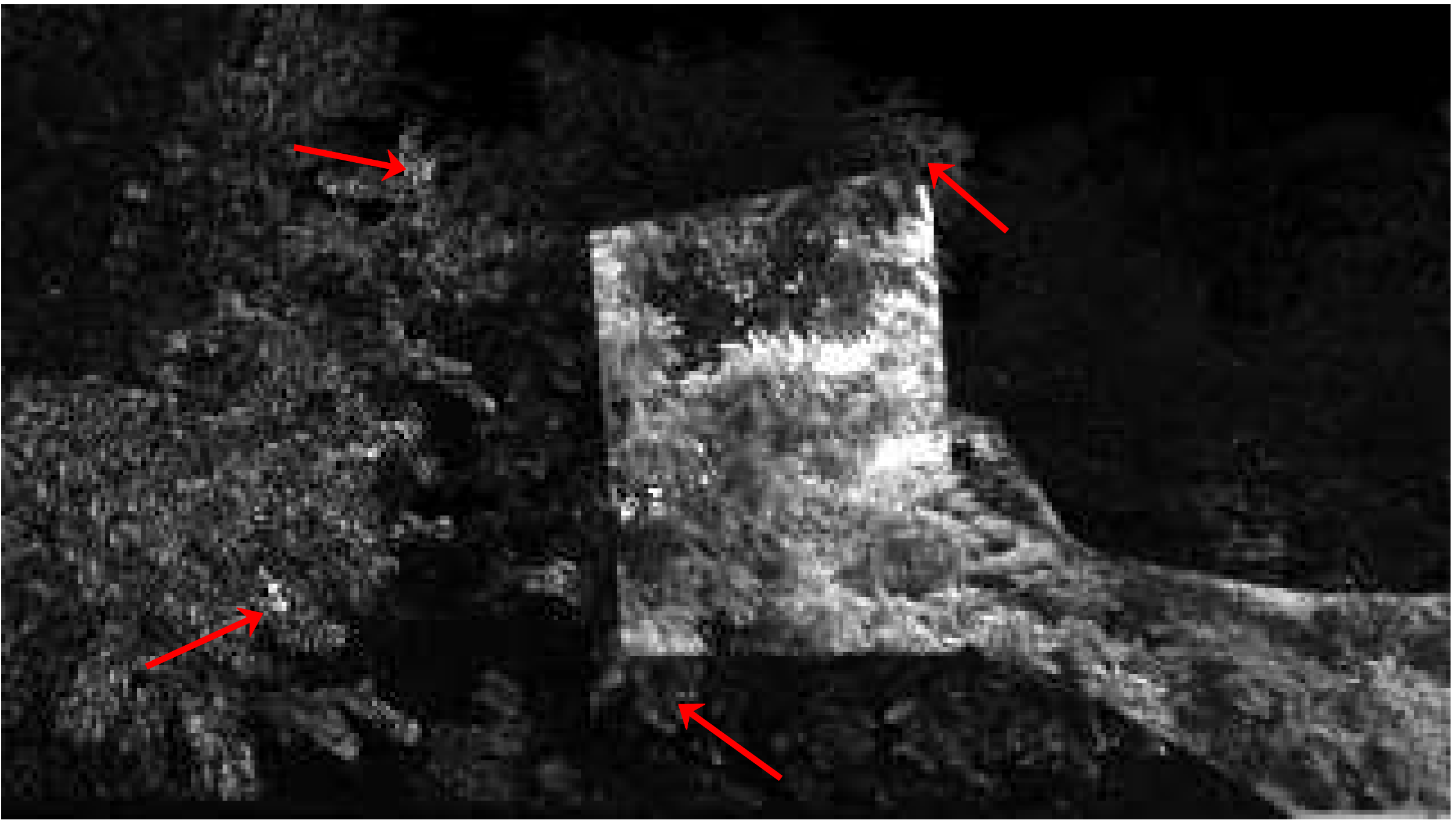}&
\hspace{-2ex}
\includegraphics[width=0.14\linewidth]{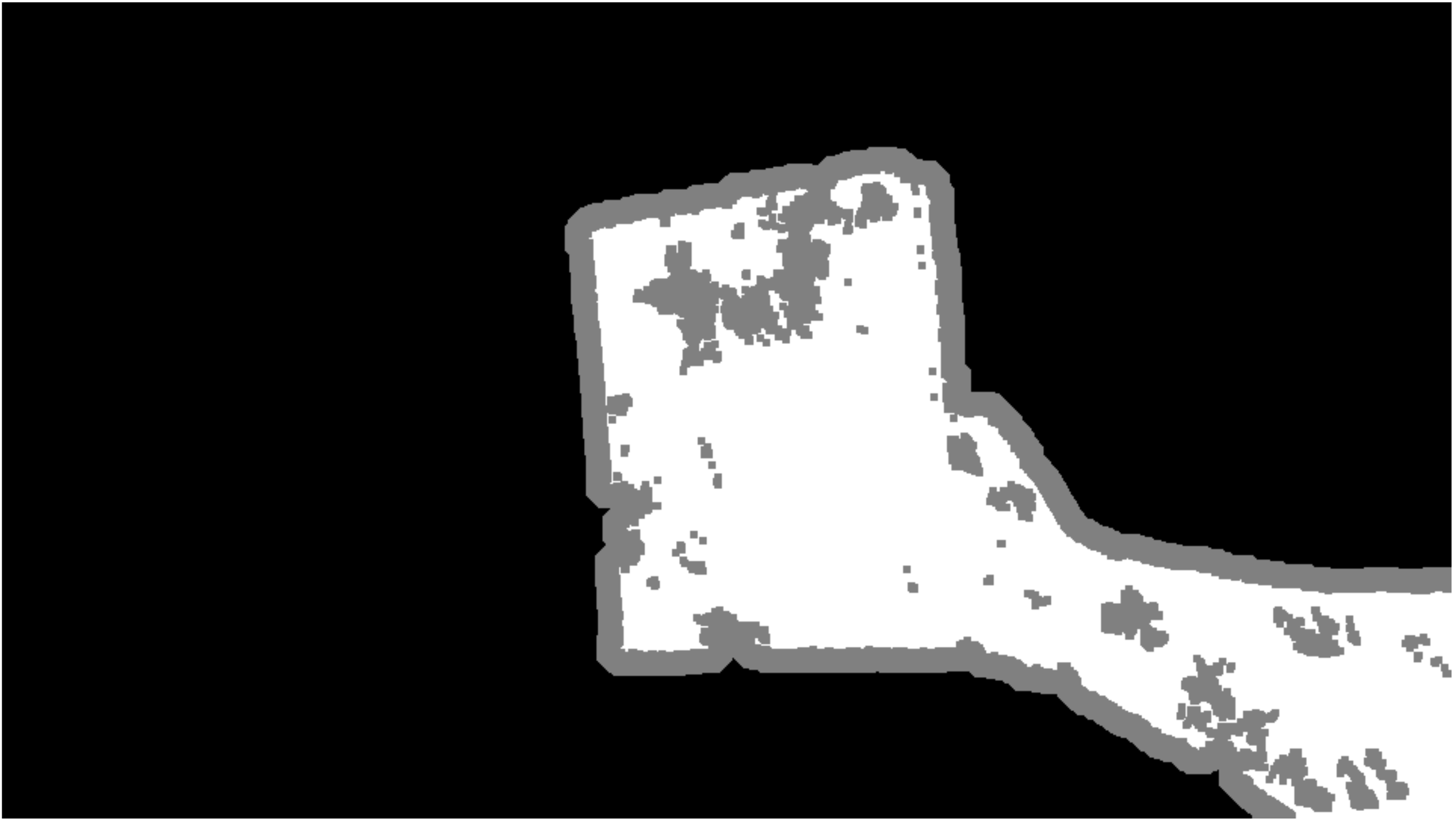}&
\hspace{-2ex}
\includegraphics[width=0.14\linewidth]{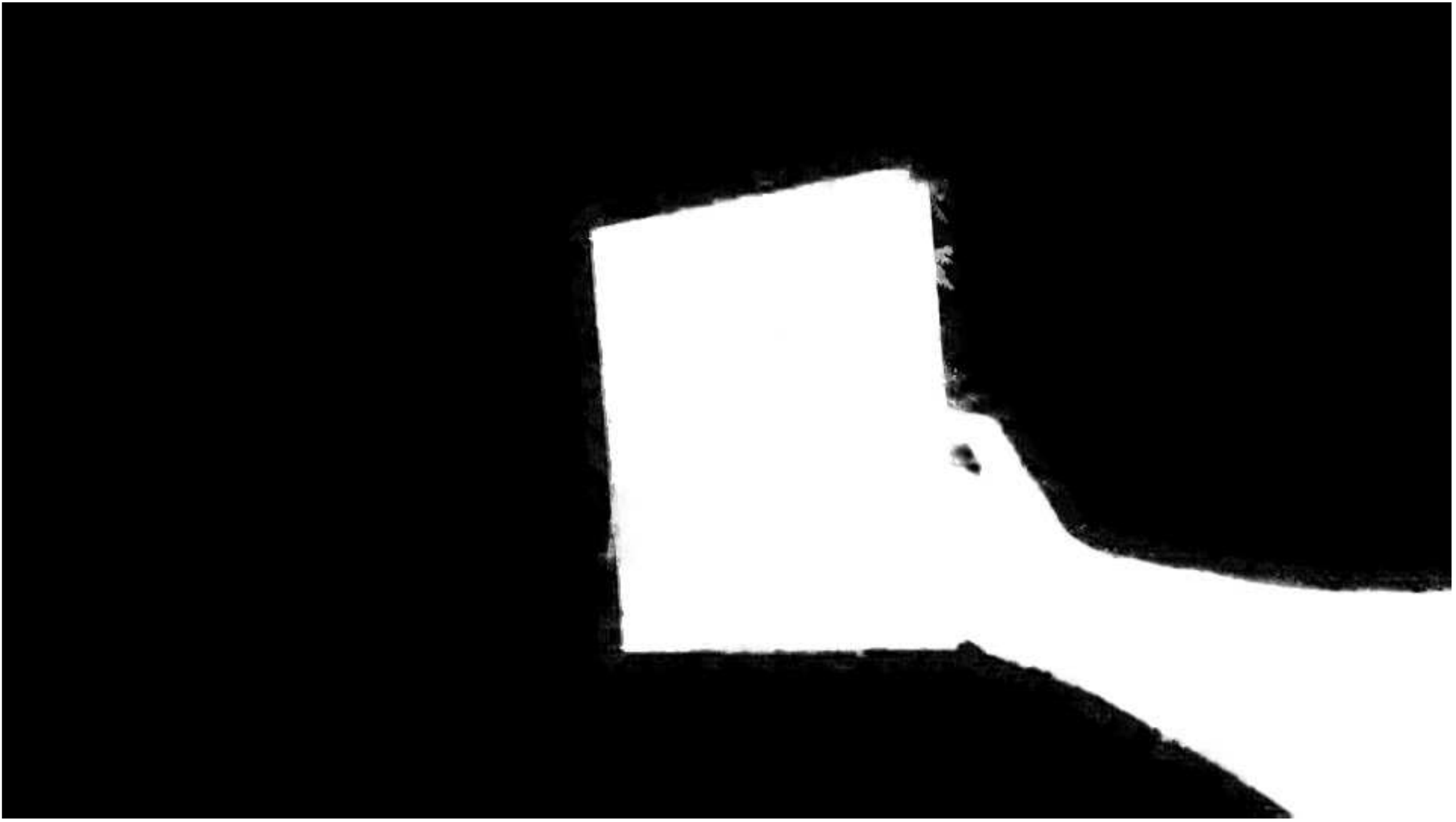}&
\hspace{-2ex}
\includegraphics[width=0.14\linewidth]{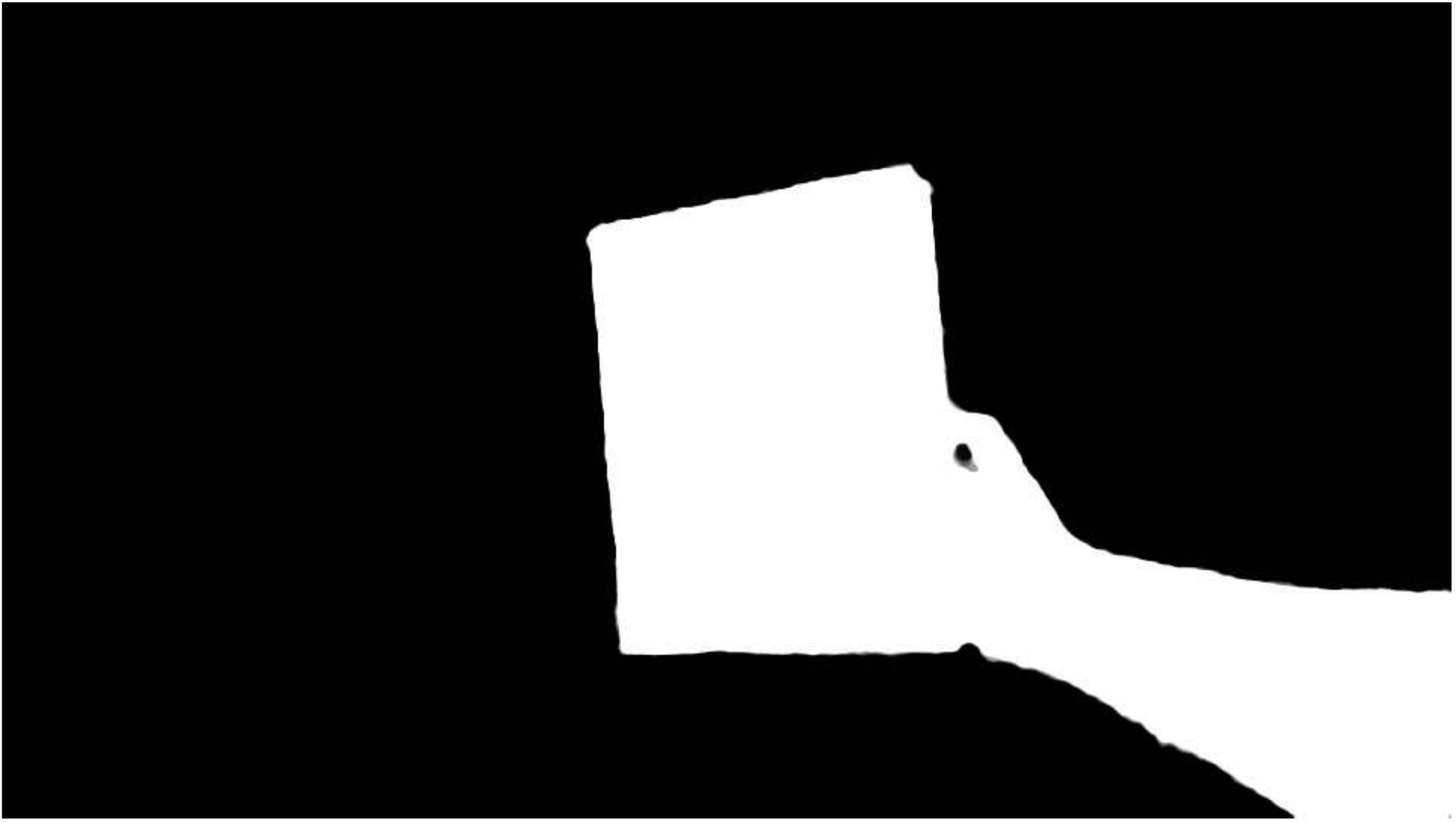}\\
II & \includegraphics[width=0.14\linewidth]{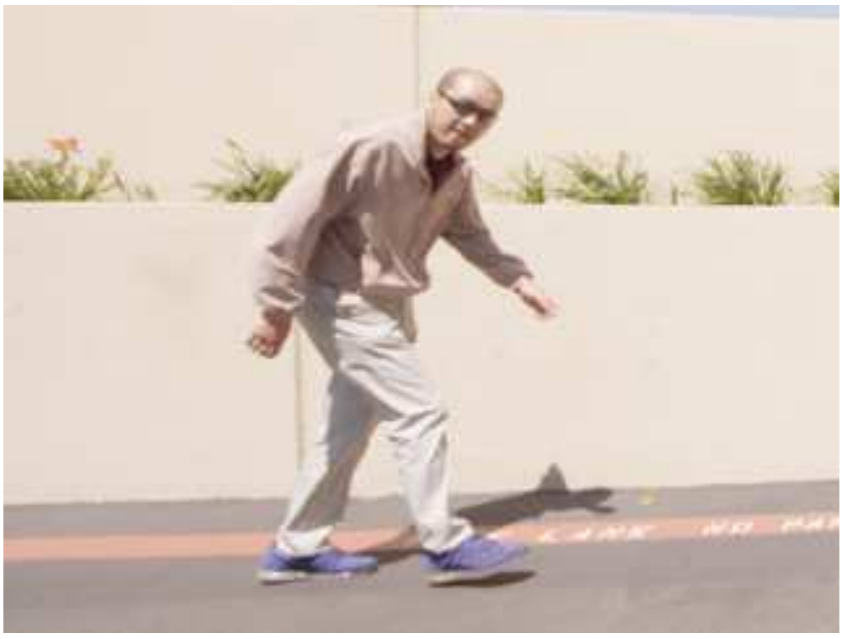}&
\hspace{-2ex}
\includegraphics[width=0.14\linewidth]{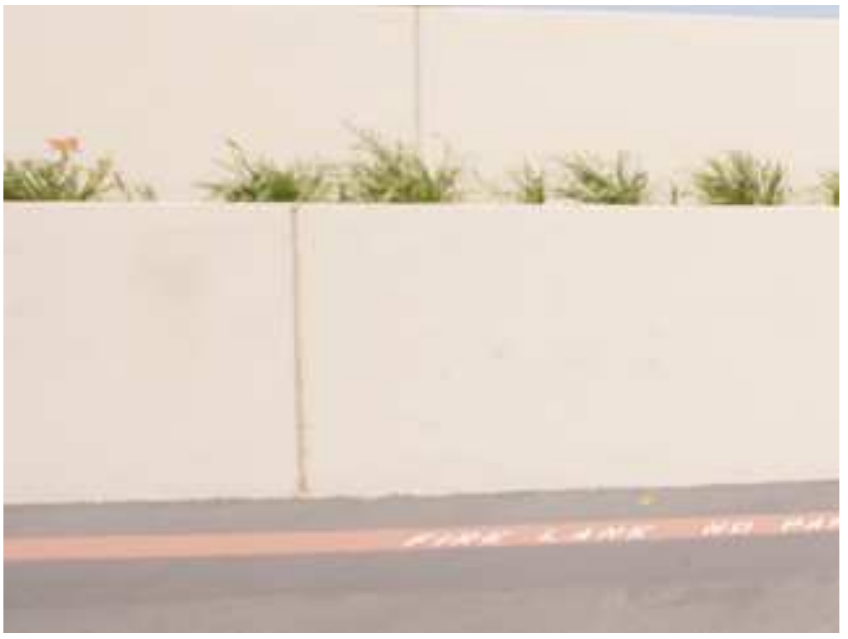}&
\hspace{-2ex}
\includegraphics[width=0.14\linewidth]{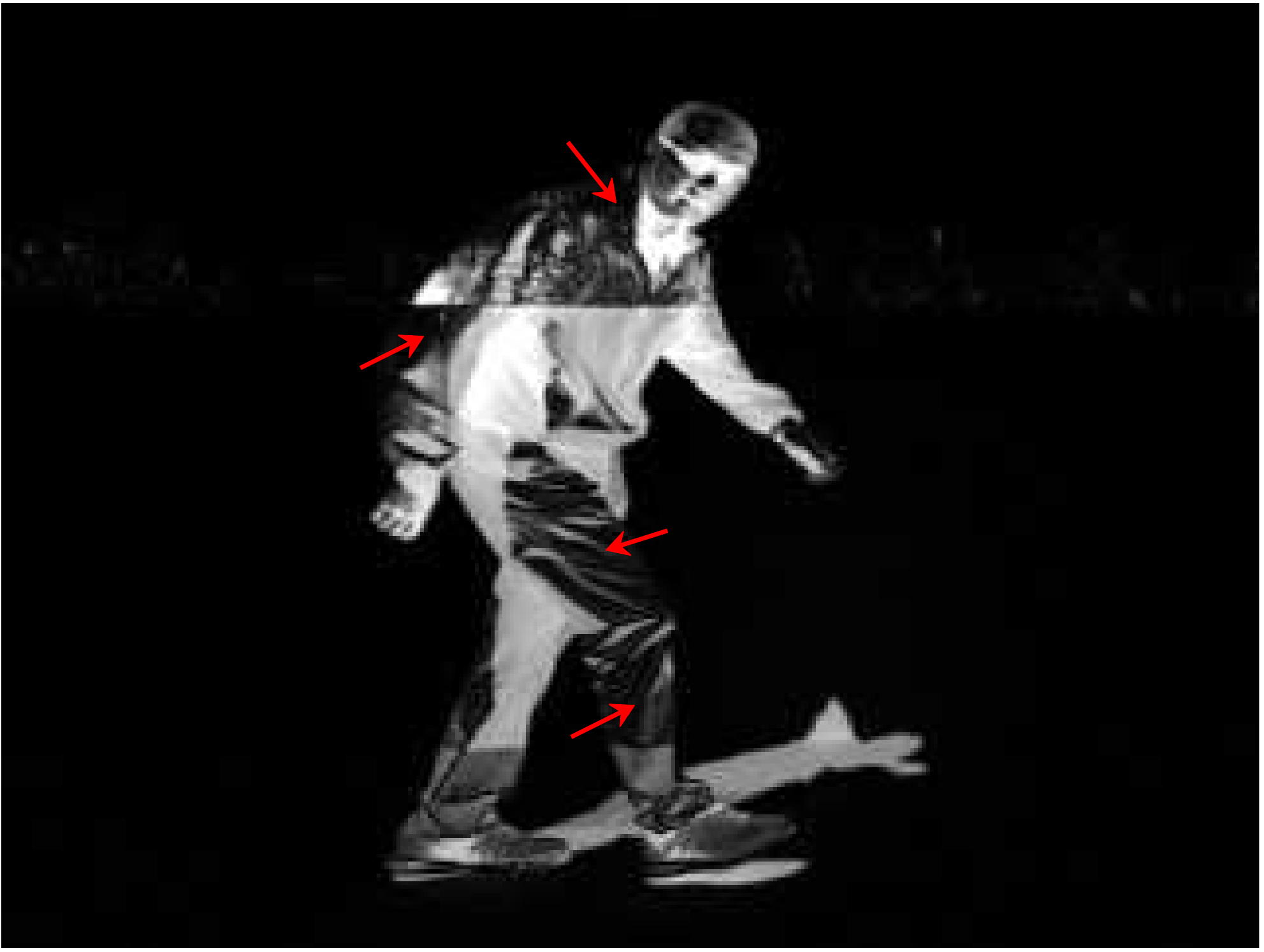}&
\hspace{-2ex}
\includegraphics[width=0.14\linewidth]{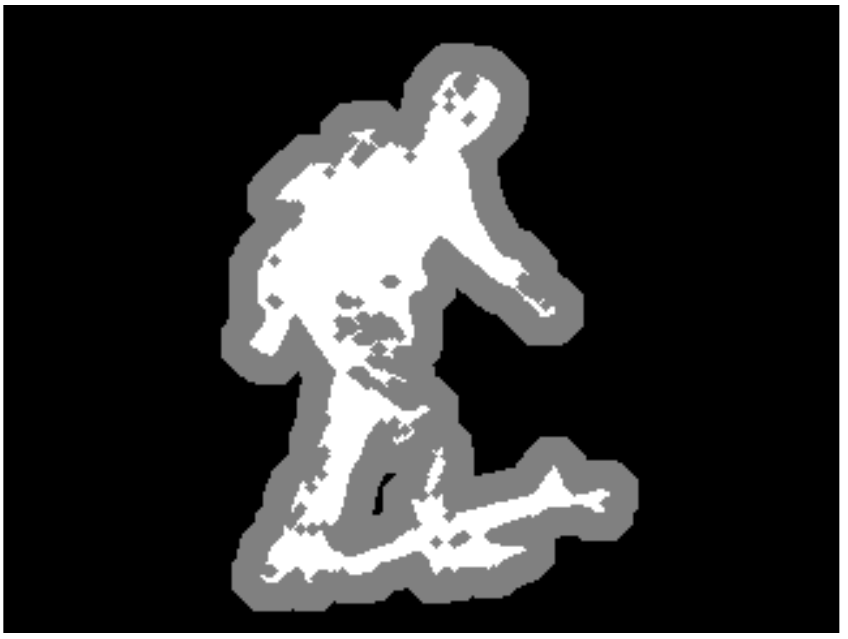}&
\hspace{-2ex}
\includegraphics[width=0.14\linewidth]{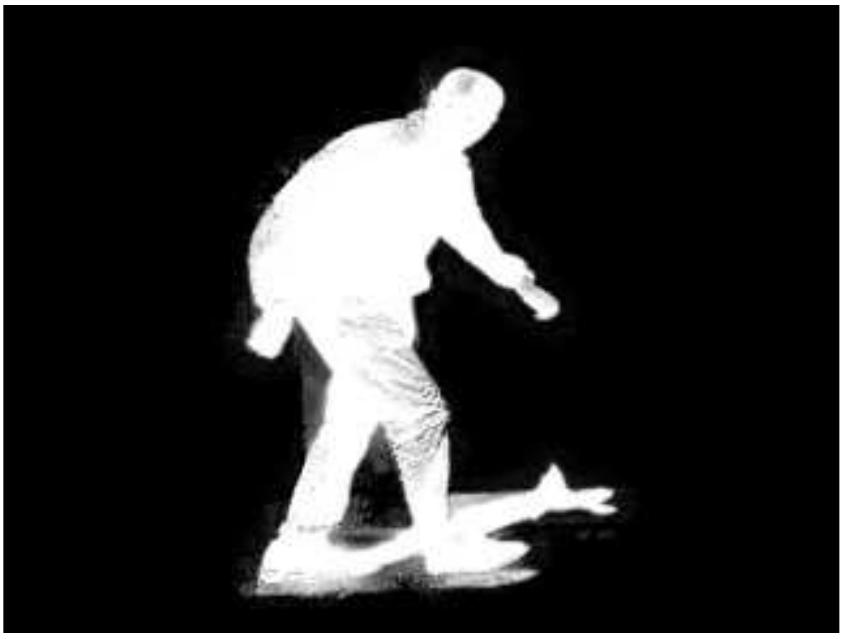}&
\hspace{-2ex}
\includegraphics[width=0.14\linewidth]{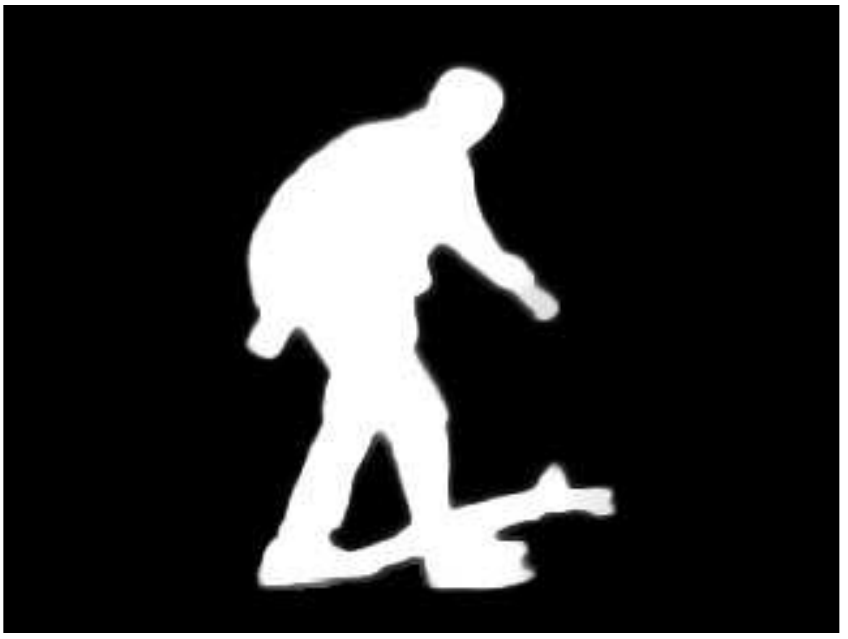}\\
III & \includegraphics[width=0.14\linewidth]{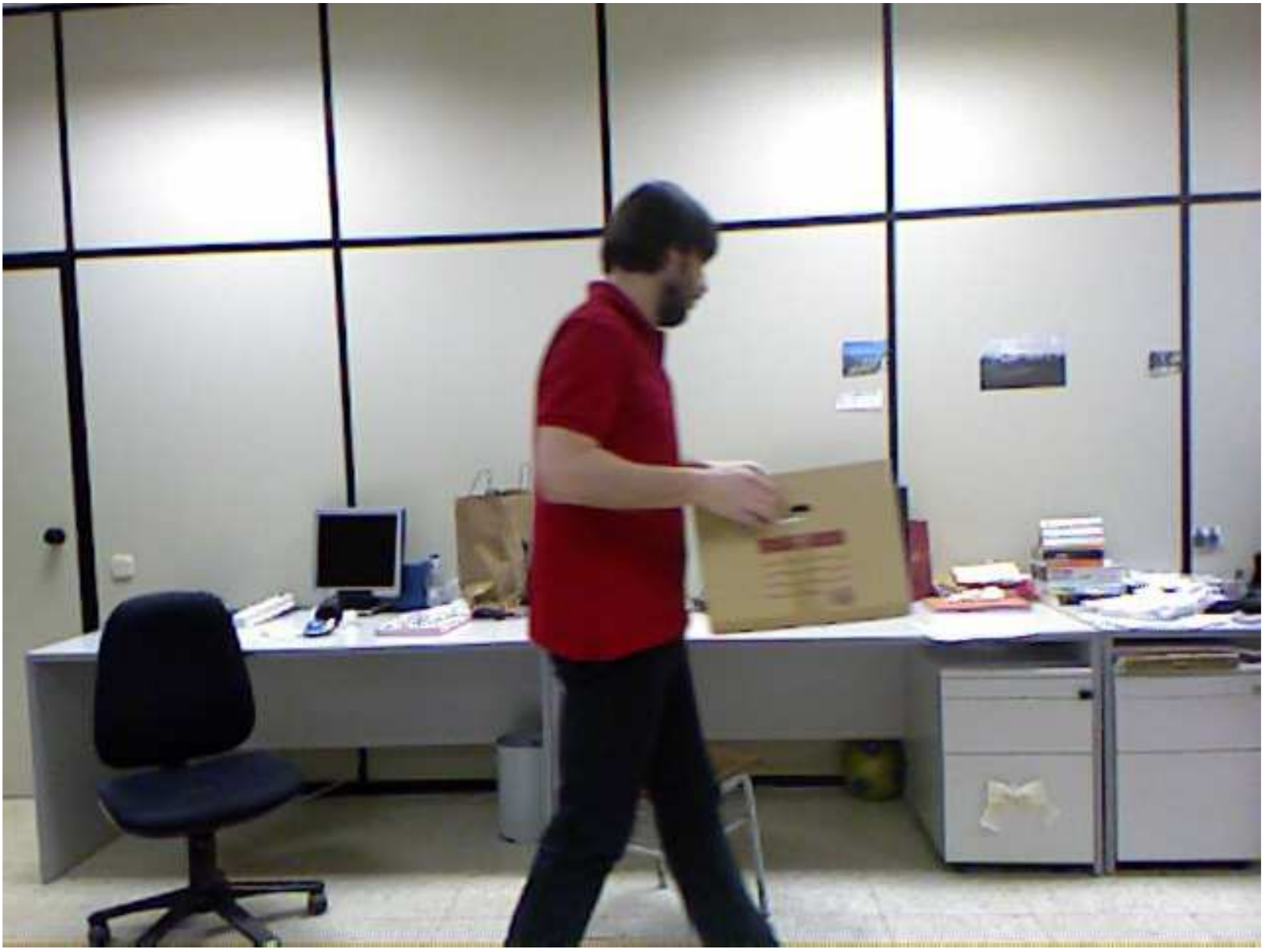}&
\hspace{-2ex}
\includegraphics[width=0.14\linewidth]{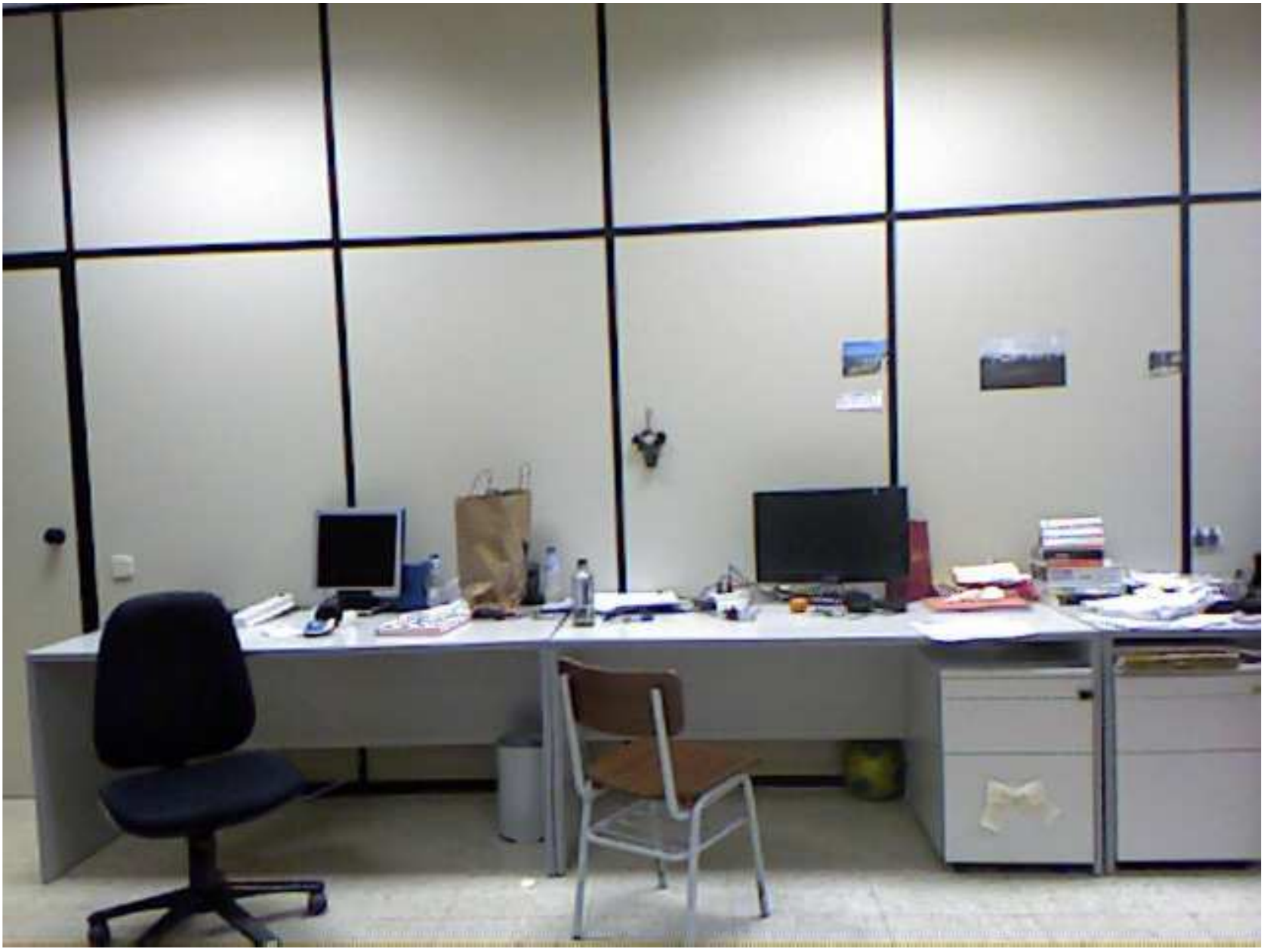}&
\hspace{-2ex}
\includegraphics[width=0.14\linewidth]{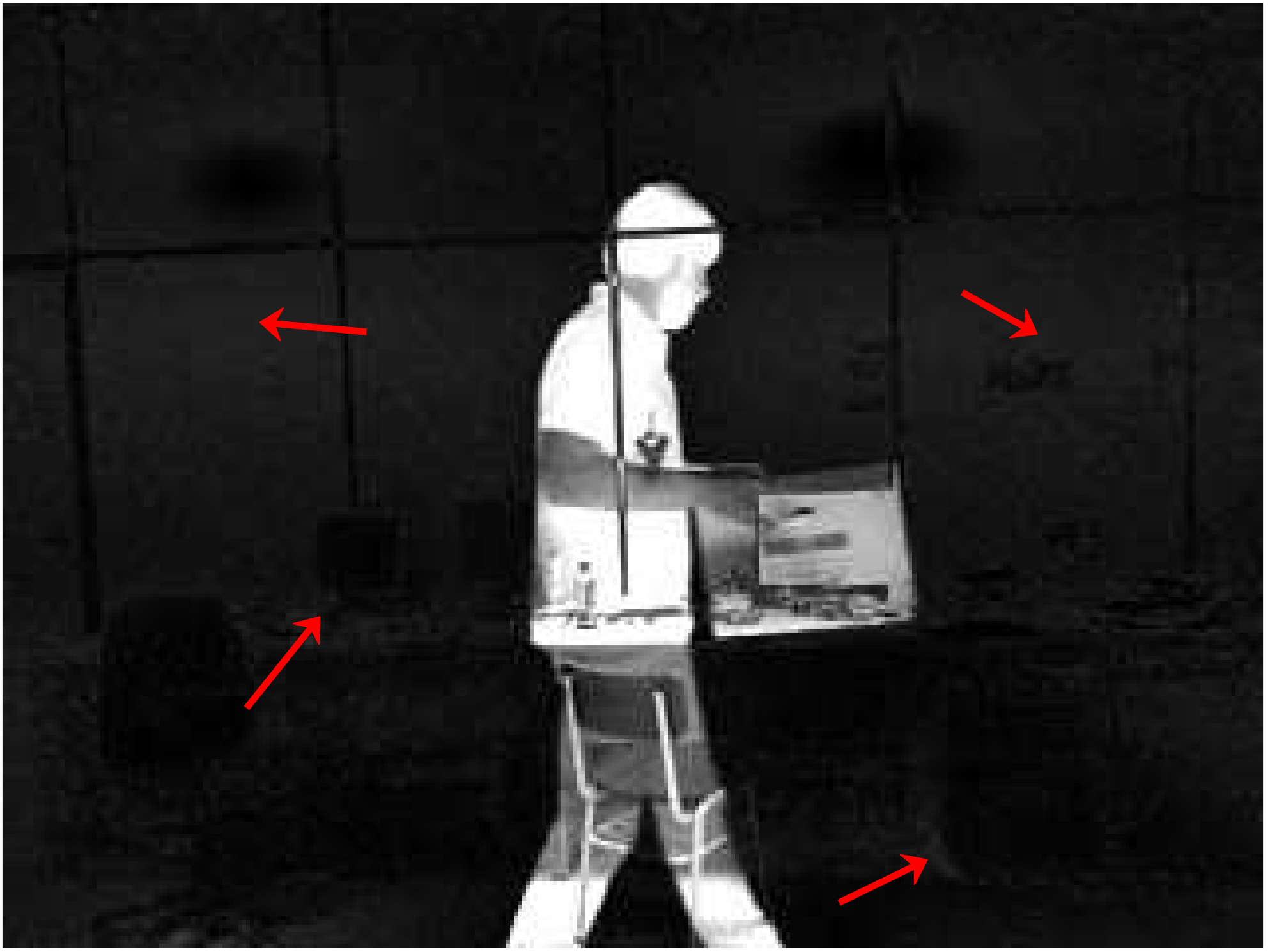}&
\hspace{-2ex}
\includegraphics[width=0.14\linewidth]{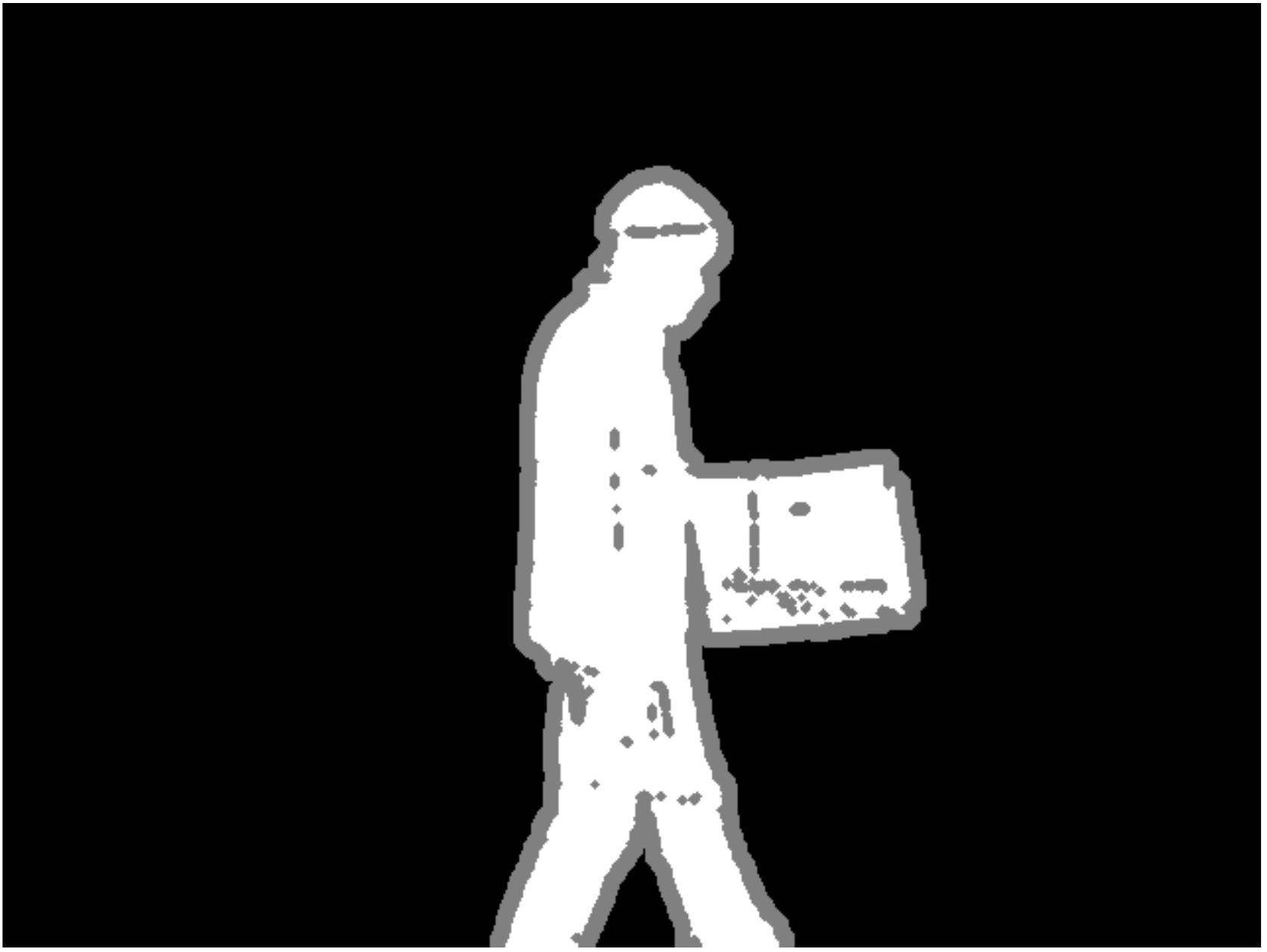}&
\hspace{-2ex}
\includegraphics[width=0.14\linewidth]{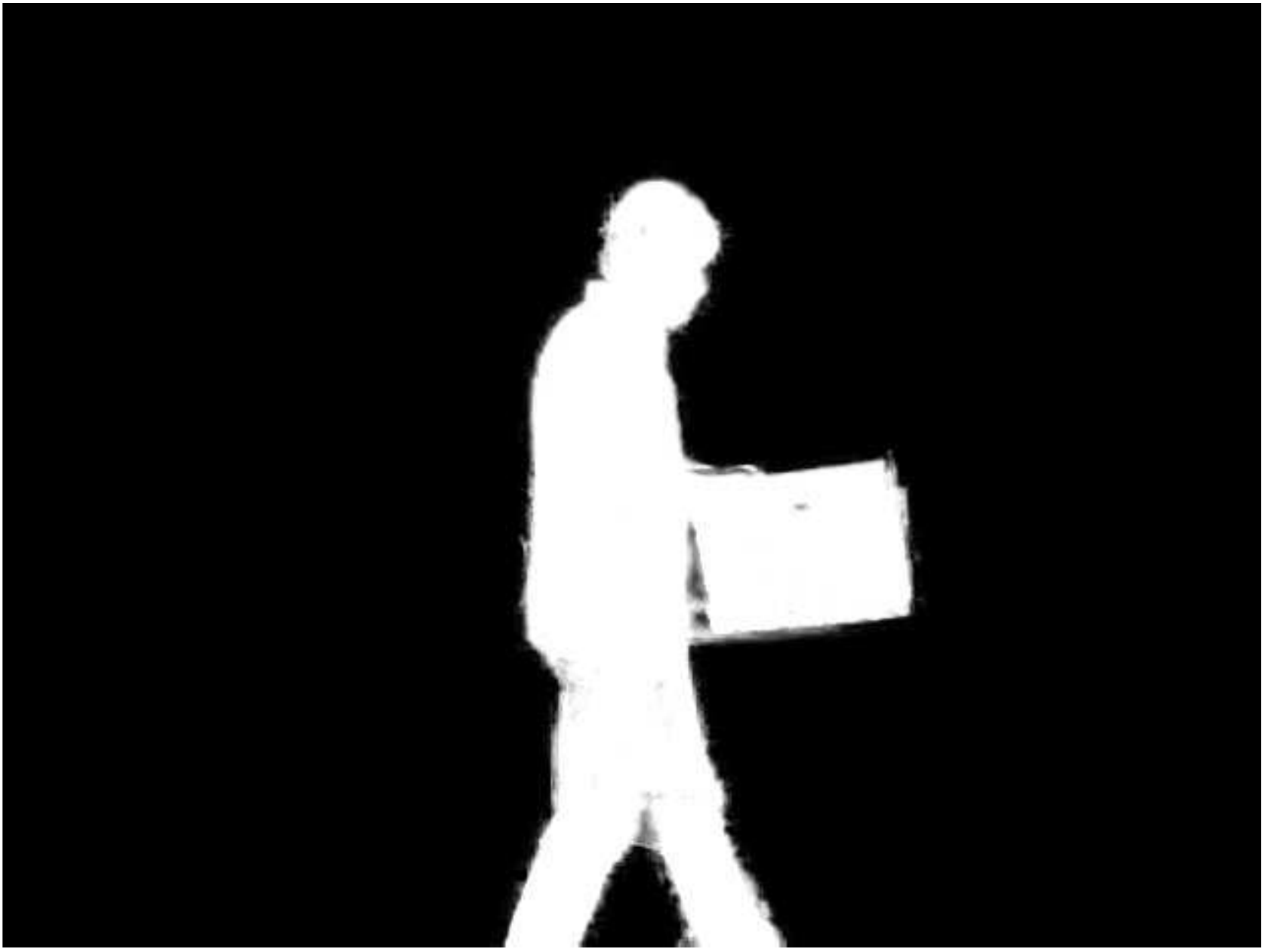}&
\hspace{-2ex}
\includegraphics[width=0.14\linewidth]{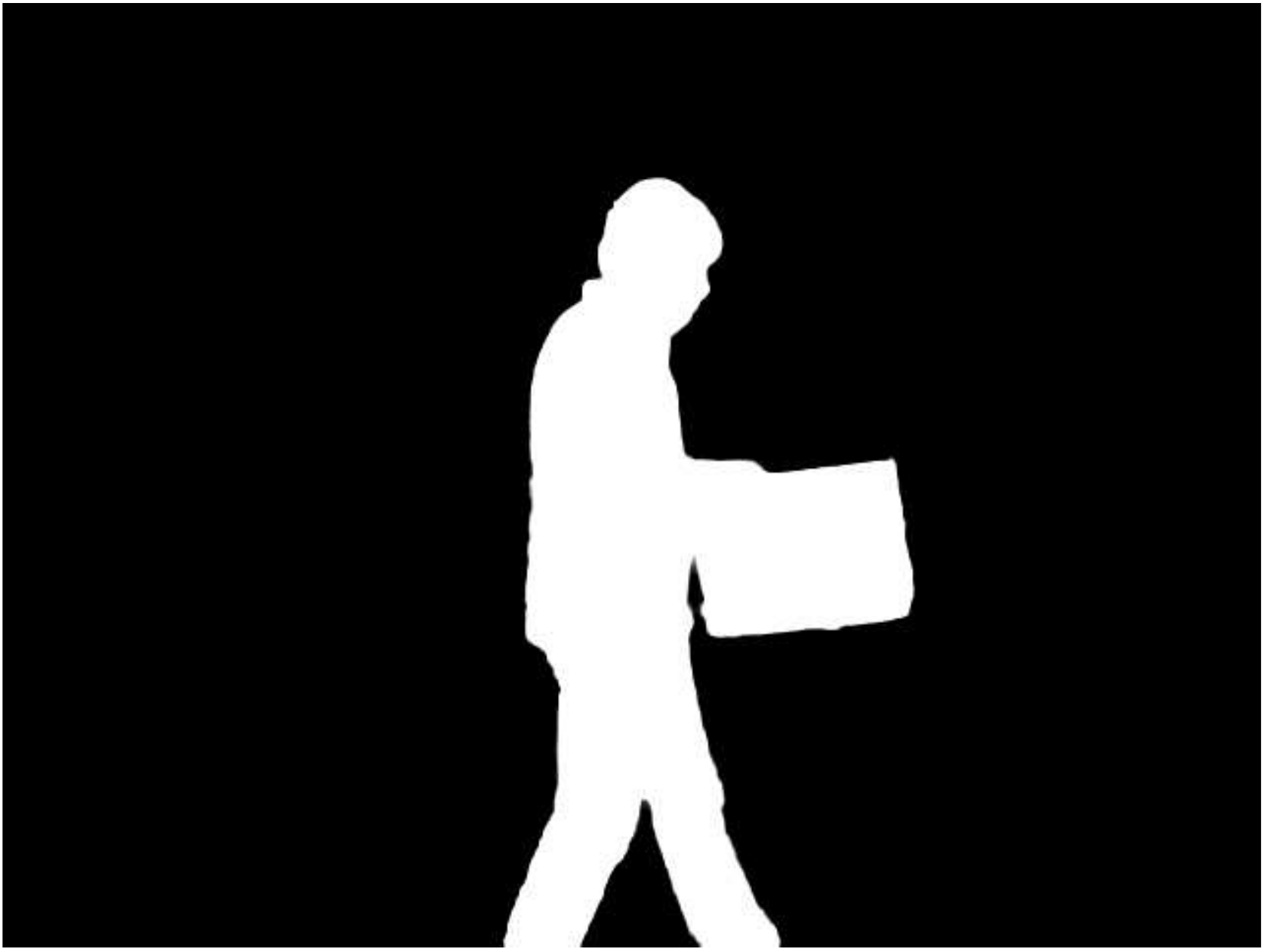}\\
&
(a) Input &
\hspace{-2ex}
(b) Plate &
\hspace{-2ex}
(c) Frame diff. &
\hspace{-2ex}
(d) Trimap  &
\hspace{-2ex}
(e)  DCNN  &
\hspace{-2ex}
(f) Ours
\end{tabular}
\caption{Three common issues of automatic foreground extraction. Case I: Vibrating background. Notice the small vibration of the leaves in the background. Case II. Similar foreground / background color. Notice the missing parts of the body of the man, and the excessive large uncertainty region of the trimap. Case III. Auto-exposure. Notice the false alarm in the background of the frame difference map. We compare our method with DCNN \cite{Cho_Tai_Kweon_2016}, a semi-supervised alpha matting method using the generated trimaps. The video data of Case III is from \cite{Camplani_Maddalena_Alcover_2017}.}
\label{fig:vibrant background}
\end{figure*}

We emphasize that the problem we study in this paper does \emph{not} belong to any of the above categories. Existing alpha matting algorithms have not been able to handle imperfect plate images, whereas background subtraction is targeting a completely different objective. Saliency based unsupervised video segmentation is an overkill, in particular those deep neural network solutions. More importantly, currently there is no training sets for plate image. This makes learning-based methods impossible. In contrast, the proposed method does not require training. Note that the plate image assumption in our work is the practical reality for many video applications. The problem remains challenging because the plate images are imperfect.

\subsection{Challenges of Imperfect Background}
Before we discuss the proposed method, we should explain the difficulty of an imperfect plate. If the plate were perfect (i.e., static and matches perfectly with the target images), then a standard frame difference with morphographic operations (e.g., erosion / dilation) would be enough to provide a trimap, and thus a sufficiently powerful alpha matting algorithm would work. When the plate image is imperfect, then complication arises because the frame difference will be heavily corrupted.

The imperfectness of the plate images comes from one or more of the following sources:
\begin{itemize}[leftmargin=*]
\item \textbf{Background vibration}. While we assume that the plate does not contain large moving objects, small vibration of the background generally exists. \fref{fig:vibrant background} Case I shows an example where the background tree vibrates.
\item \textbf{Color similarity}. When foreground color is very similar to the background color, the trimap generated will have false alarms and misses. \fref{fig:vibrant background} Case II shows an example where the cloth of the man has a similar color to the wall.
\item \textbf{Auto-exposure}. If auto-exposure is used, the background intensity will change over time. \fref{fig:vibrant background} Case III shows an example where the background cabinet becomes dimmer when the man leaves the room.
\end{itemize}

\begin{figure*}[t]
\centering
\begin{tabular}{ccccc}
\includegraphics[width=0.18\linewidth]{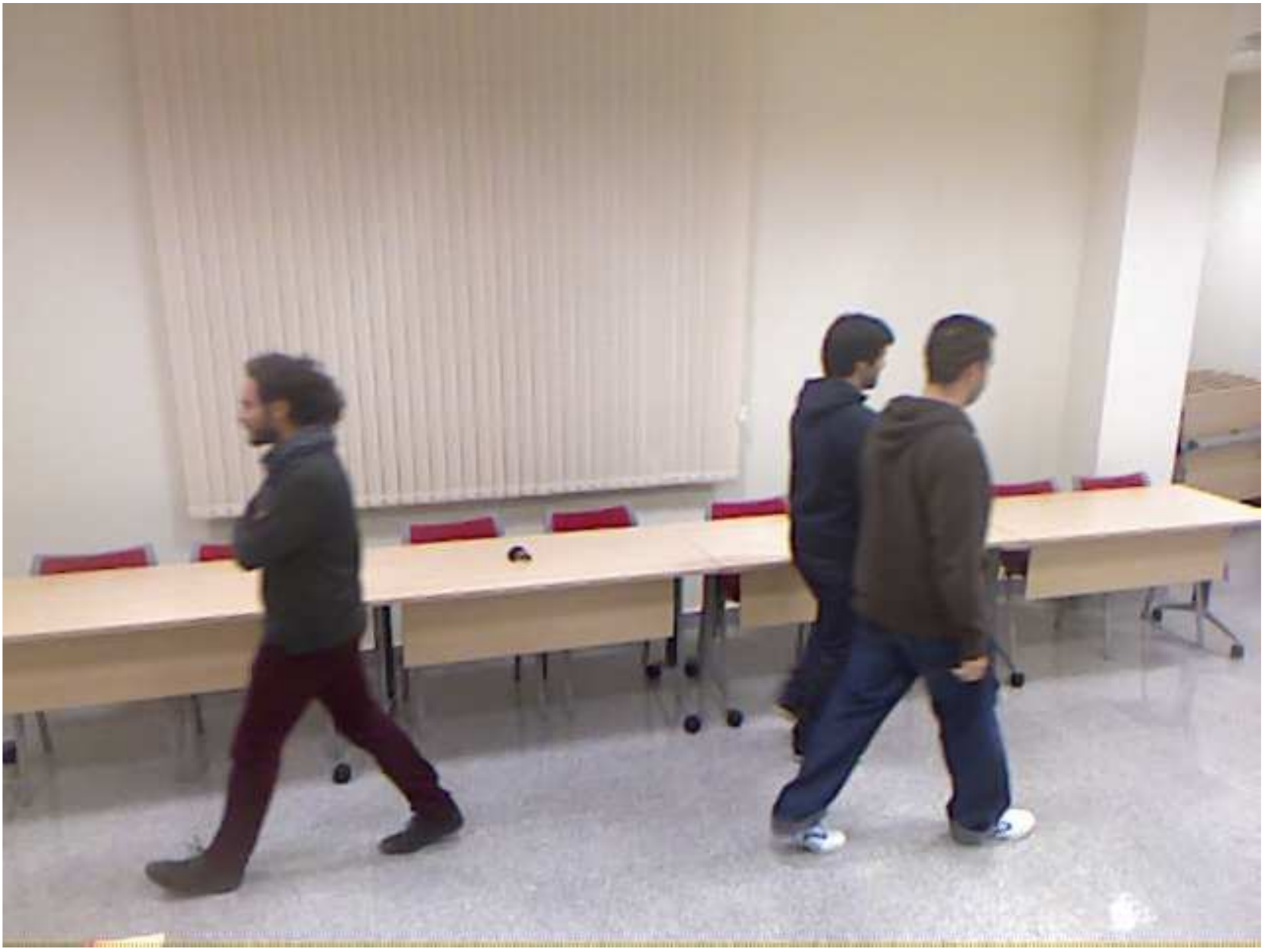}&
\hspace{-2ex}\includegraphics[width=0.18\linewidth]{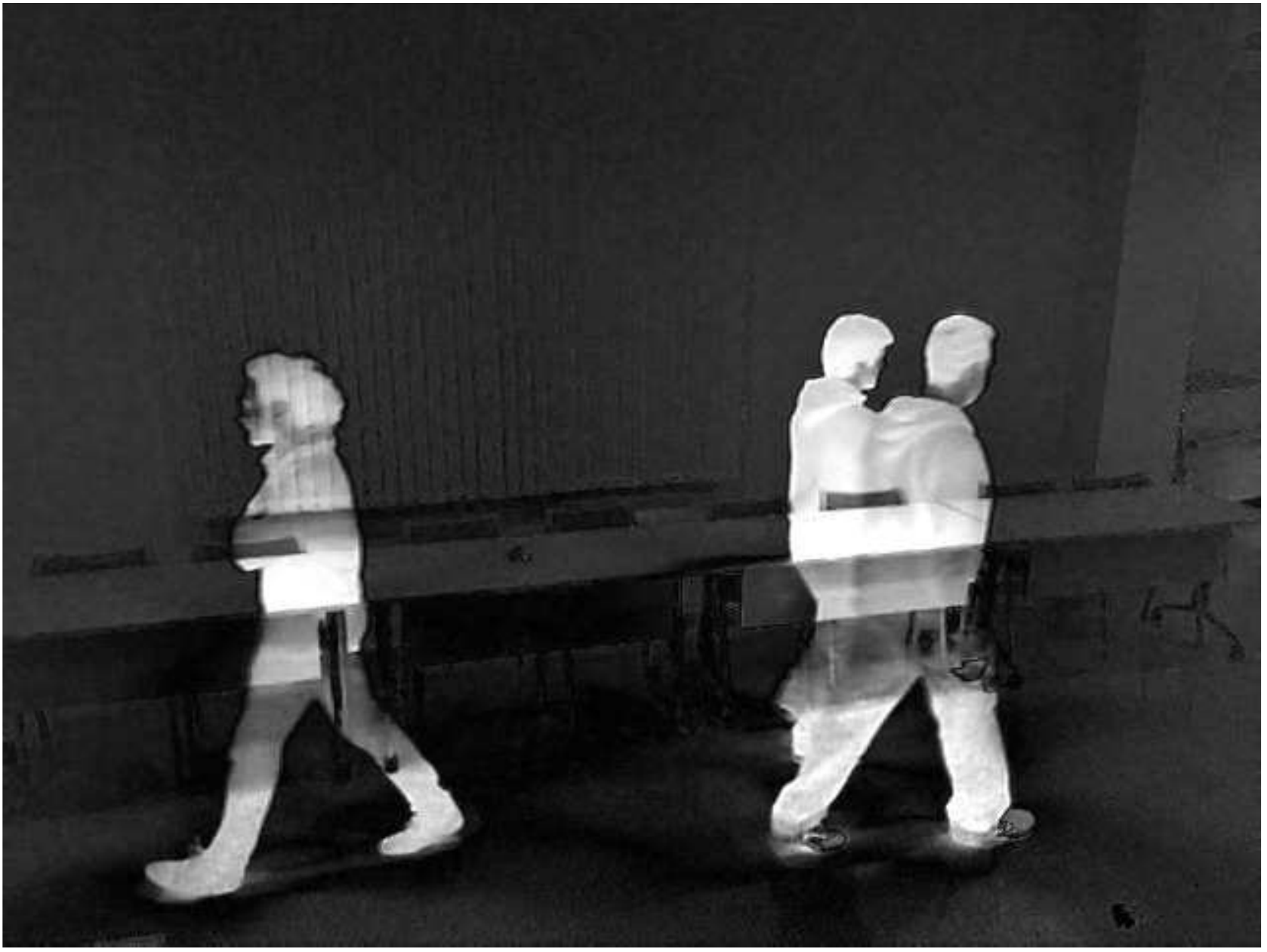}&
\hspace{-2ex}\includegraphics[width=0.18\linewidth]{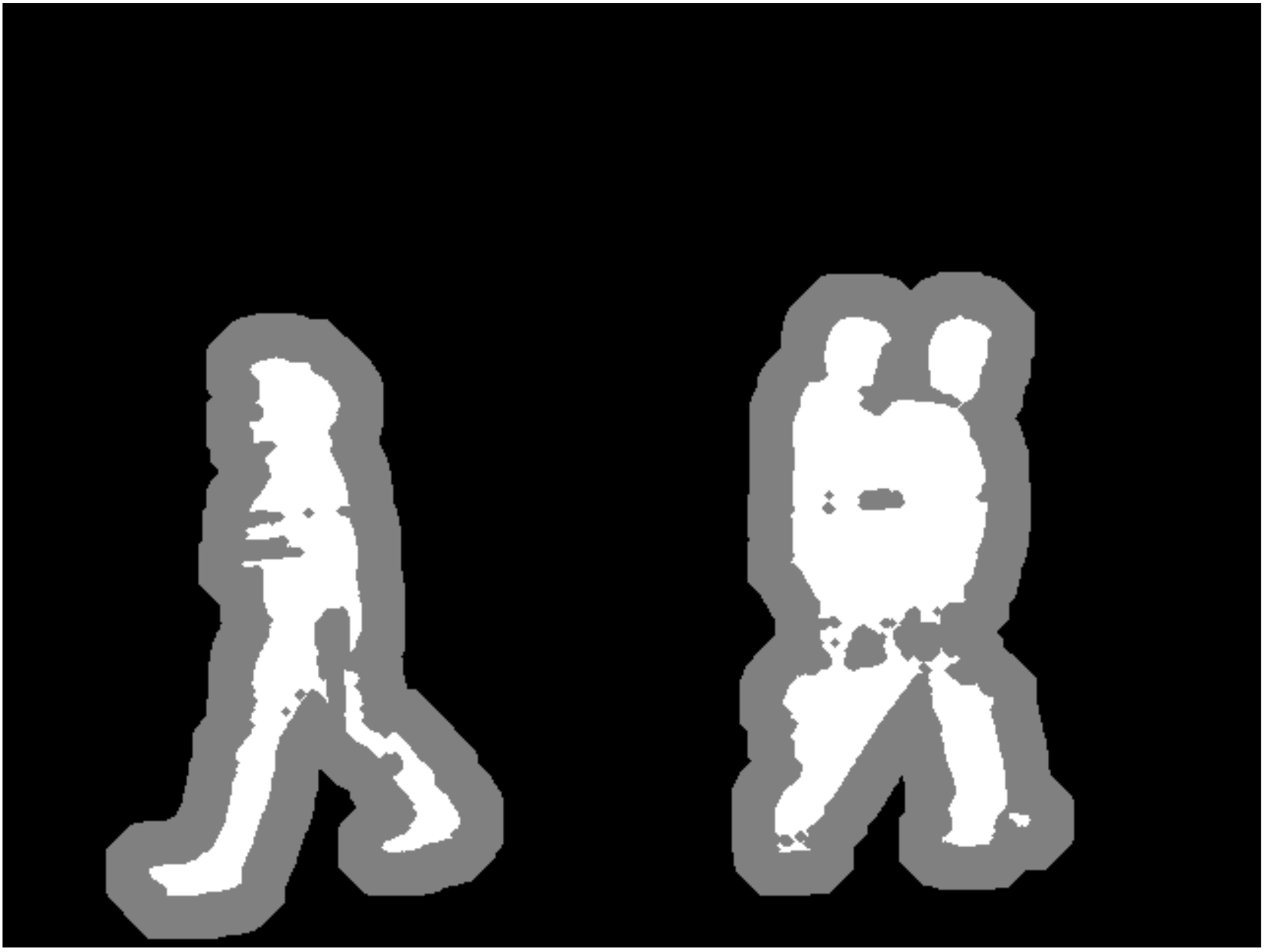}&
\hspace{-2ex}\includegraphics[width=0.18\linewidth]{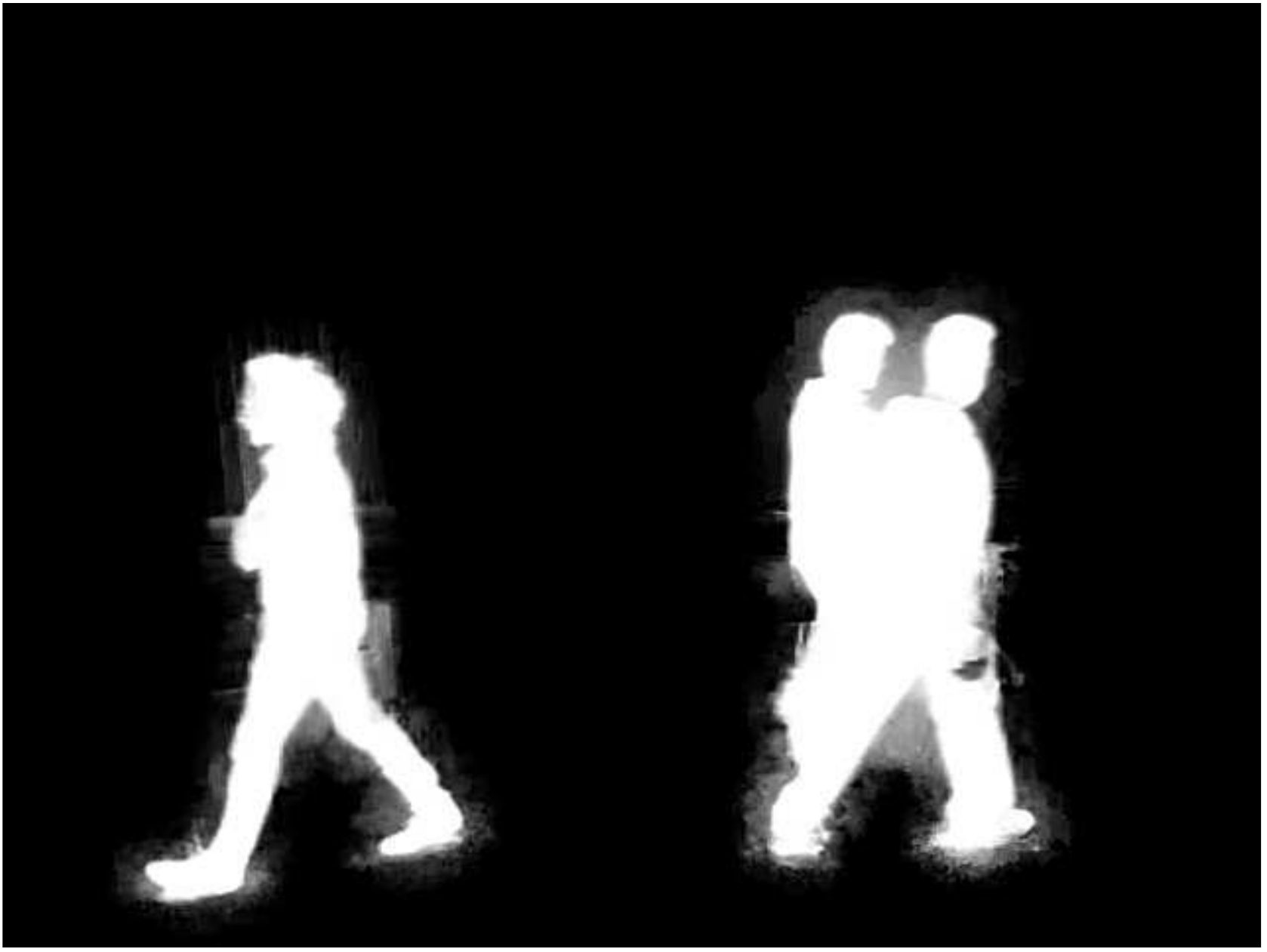}&
\hspace{-2ex}\includegraphics[width=0.18\linewidth]{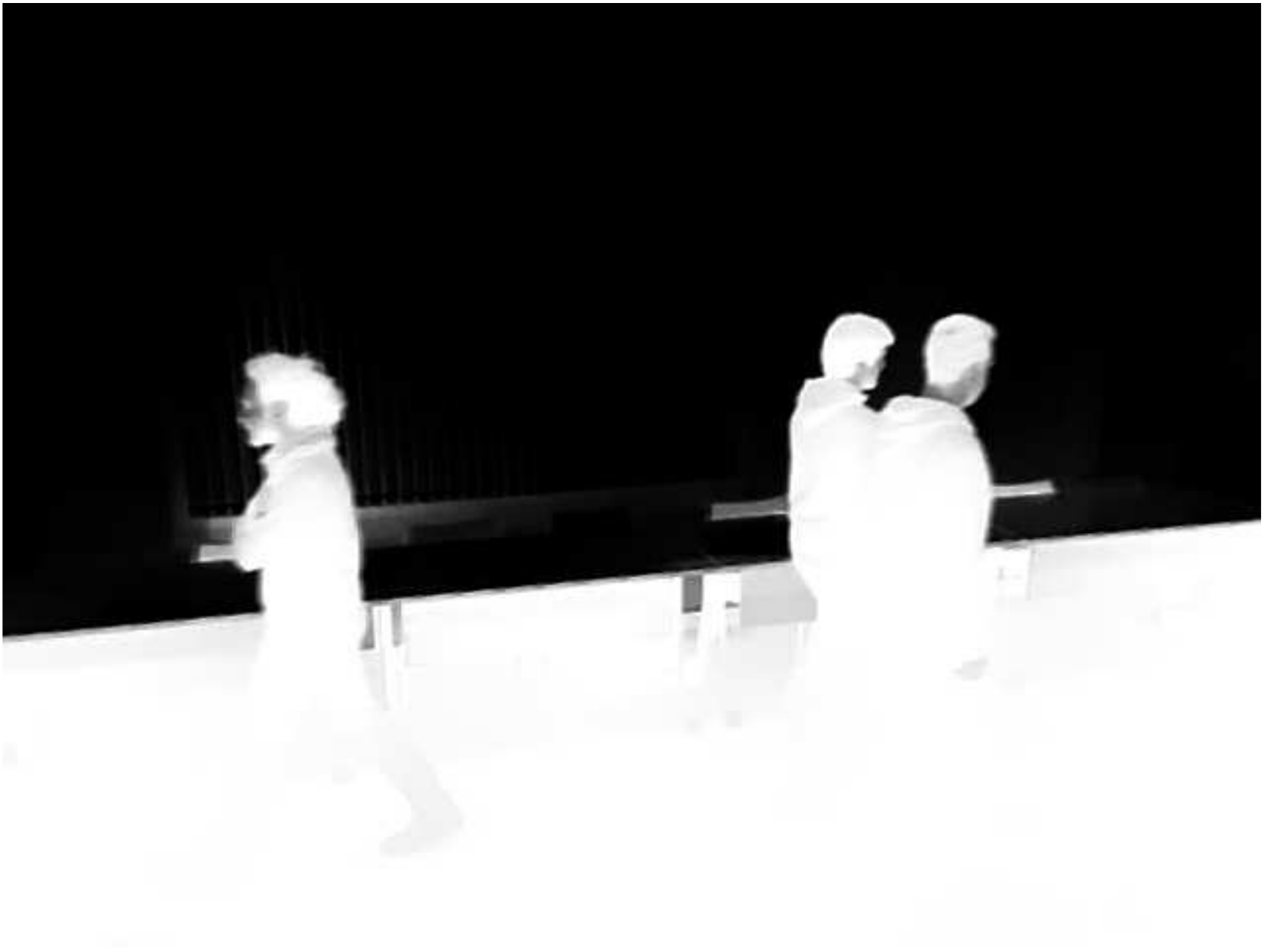}\\
(a) Input &
\hspace{-2ex} (b) Frame difference &
\hspace{-2ex} (c) Trimap  &
\hspace{-2ex} (d) Closed-form   &
\hspace{-2ex} (e) Spectral matting\\
 &
\hspace{-2ex}  &
\hspace{-2ex} &
\hspace{-2ex} matting \cite{Levin_Lischinski_Weiss_2008}  &
\hspace{-2ex} \cite{Levin_Rav-Acha_Lischinski_2008}\\
\includegraphics[width=0.18\linewidth]{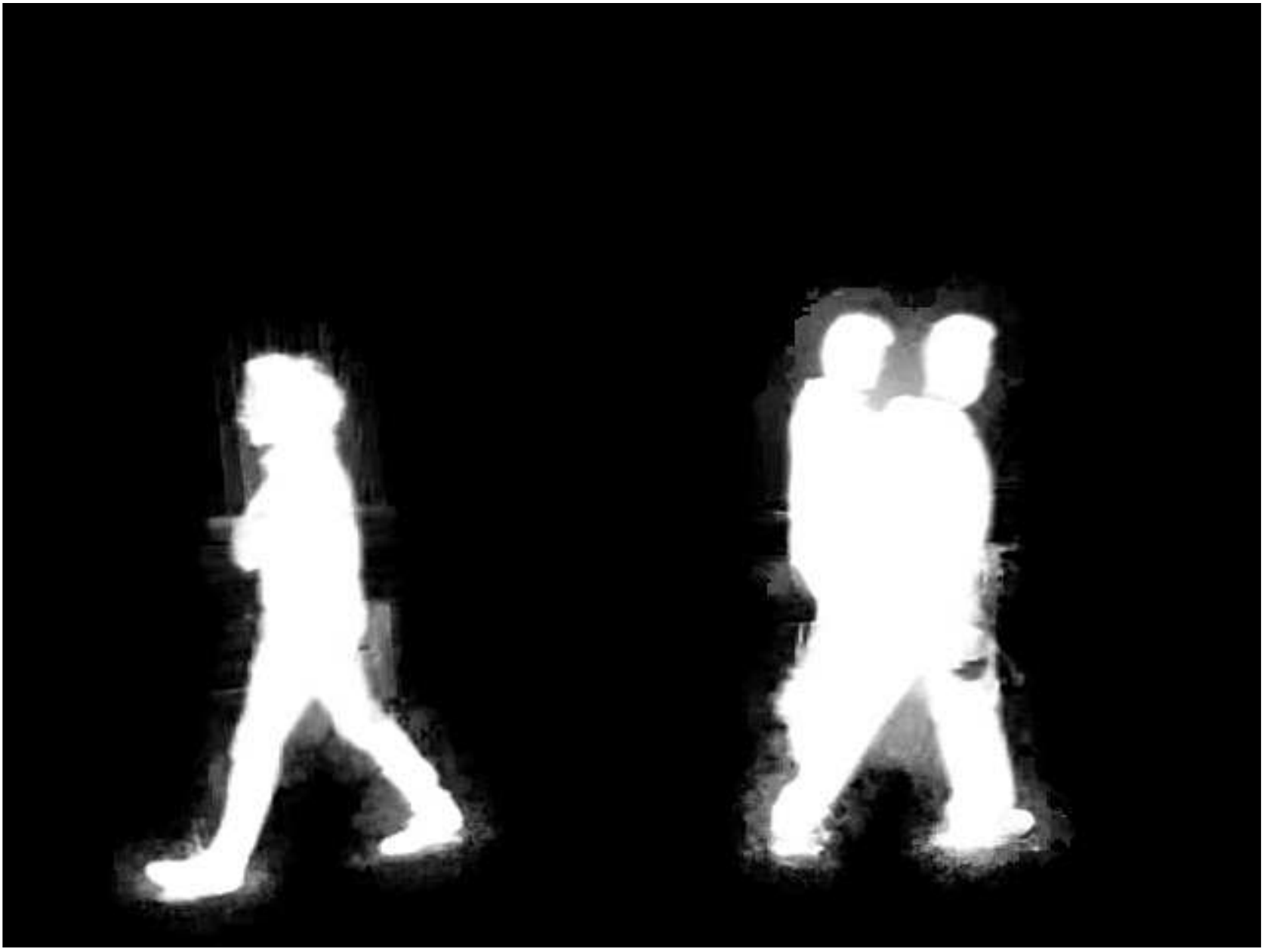}&
\hspace{-2ex}\includegraphics[width=0.18\linewidth]{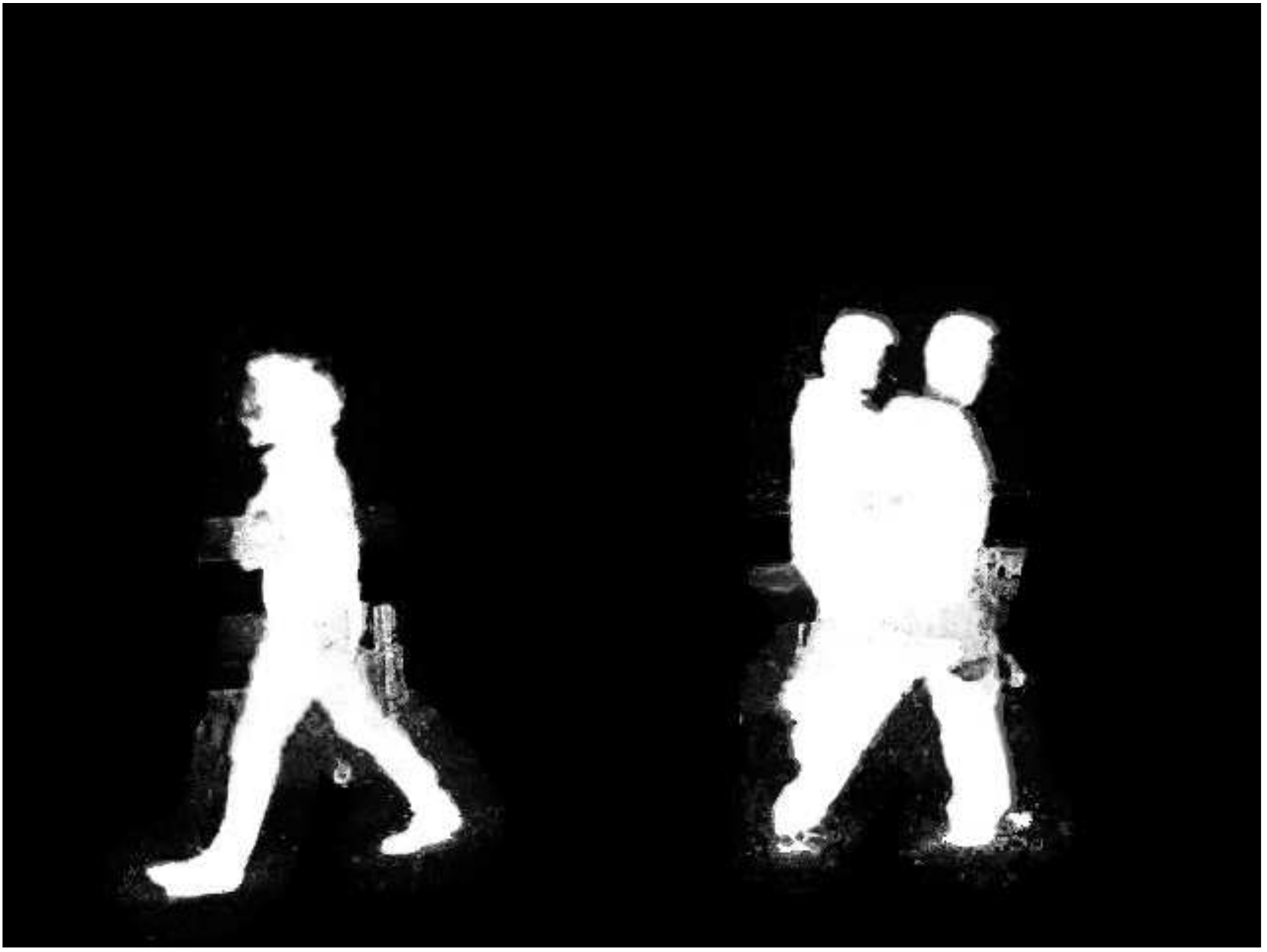}&
\hspace{-2ex}\includegraphics[width=0.18\linewidth]{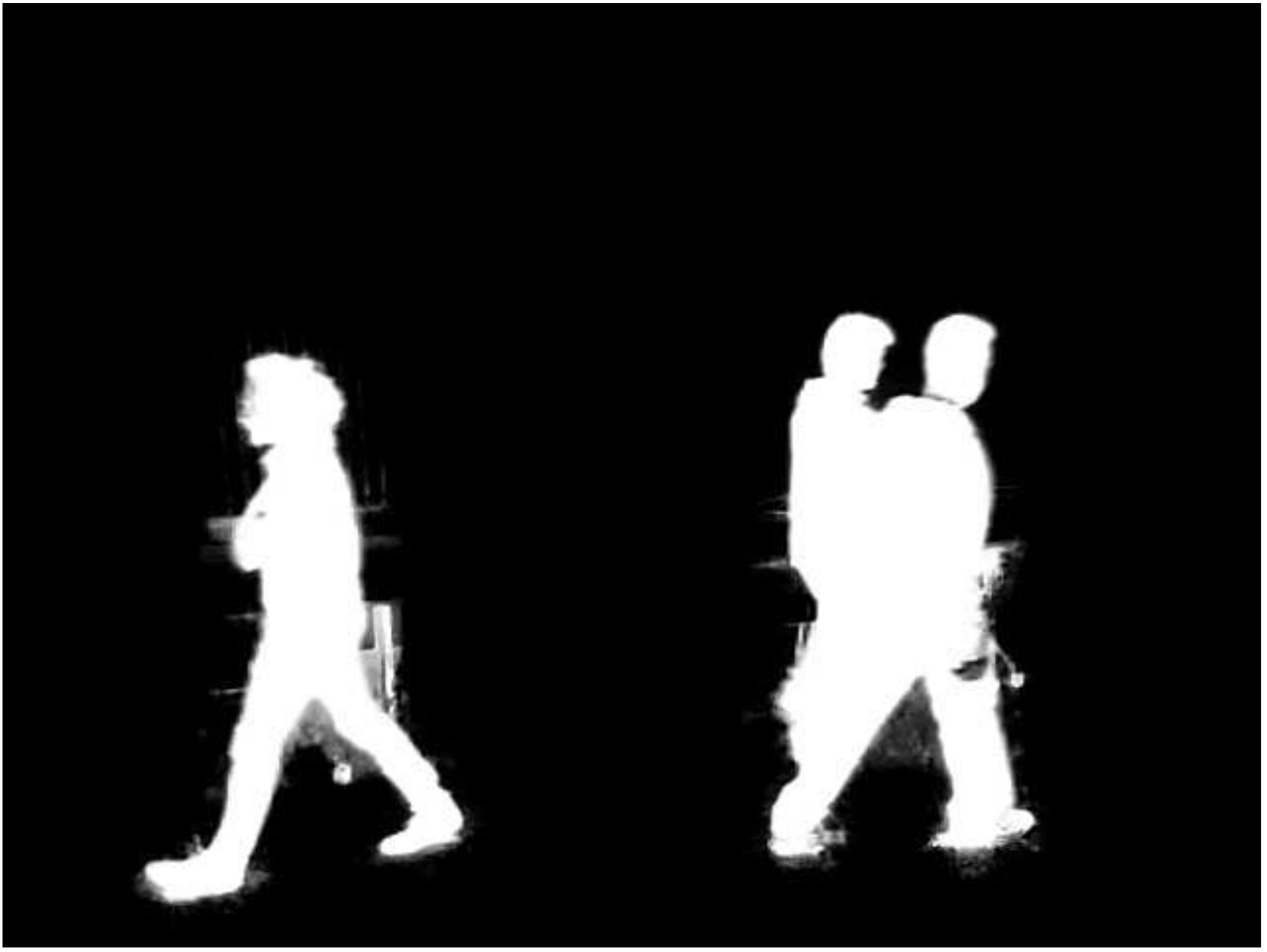}&
\hspace{-2ex}\includegraphics[width=0.18\linewidth]{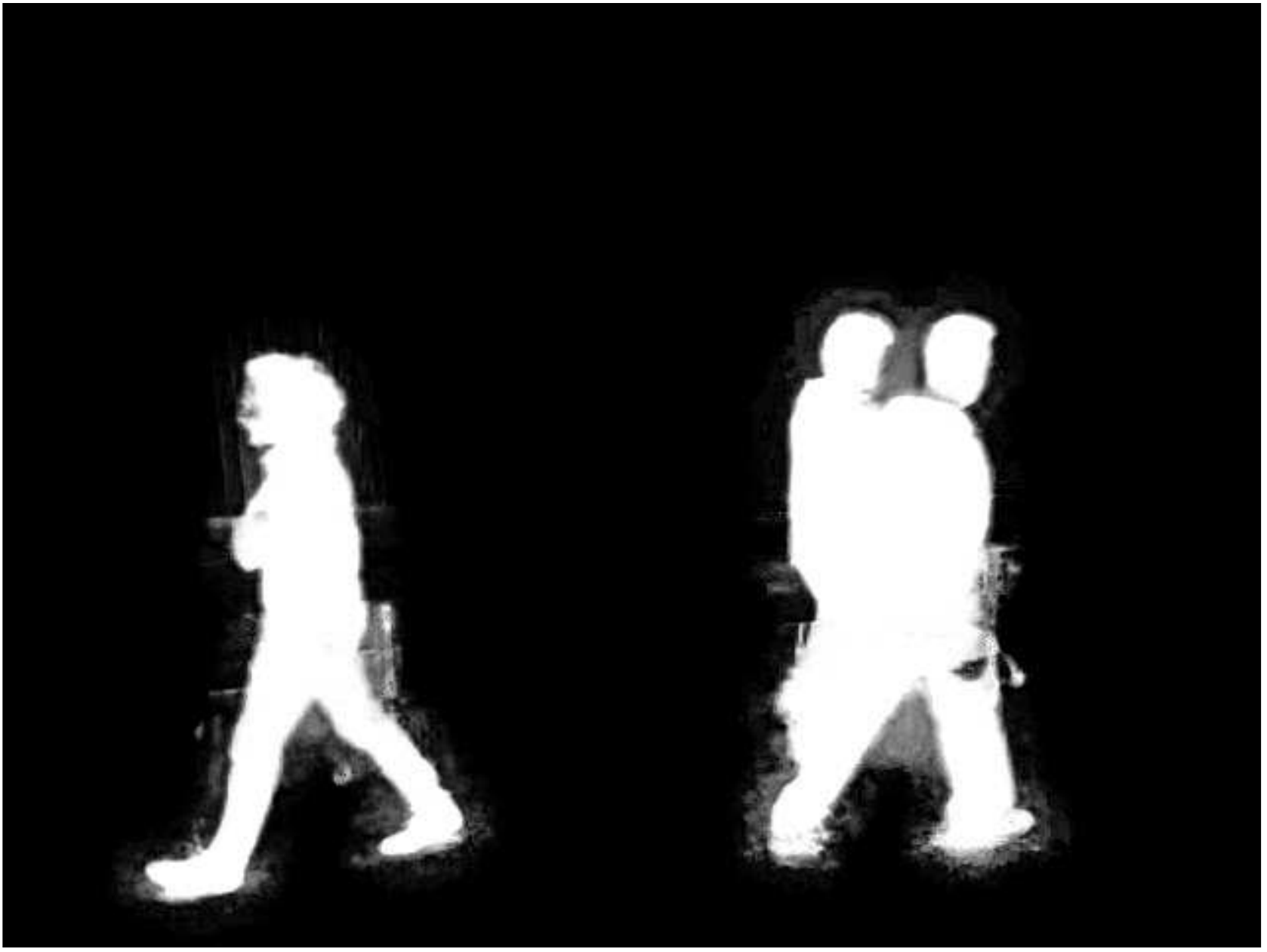}&
\hspace{-2ex}\includegraphics[width=0.18\linewidth]{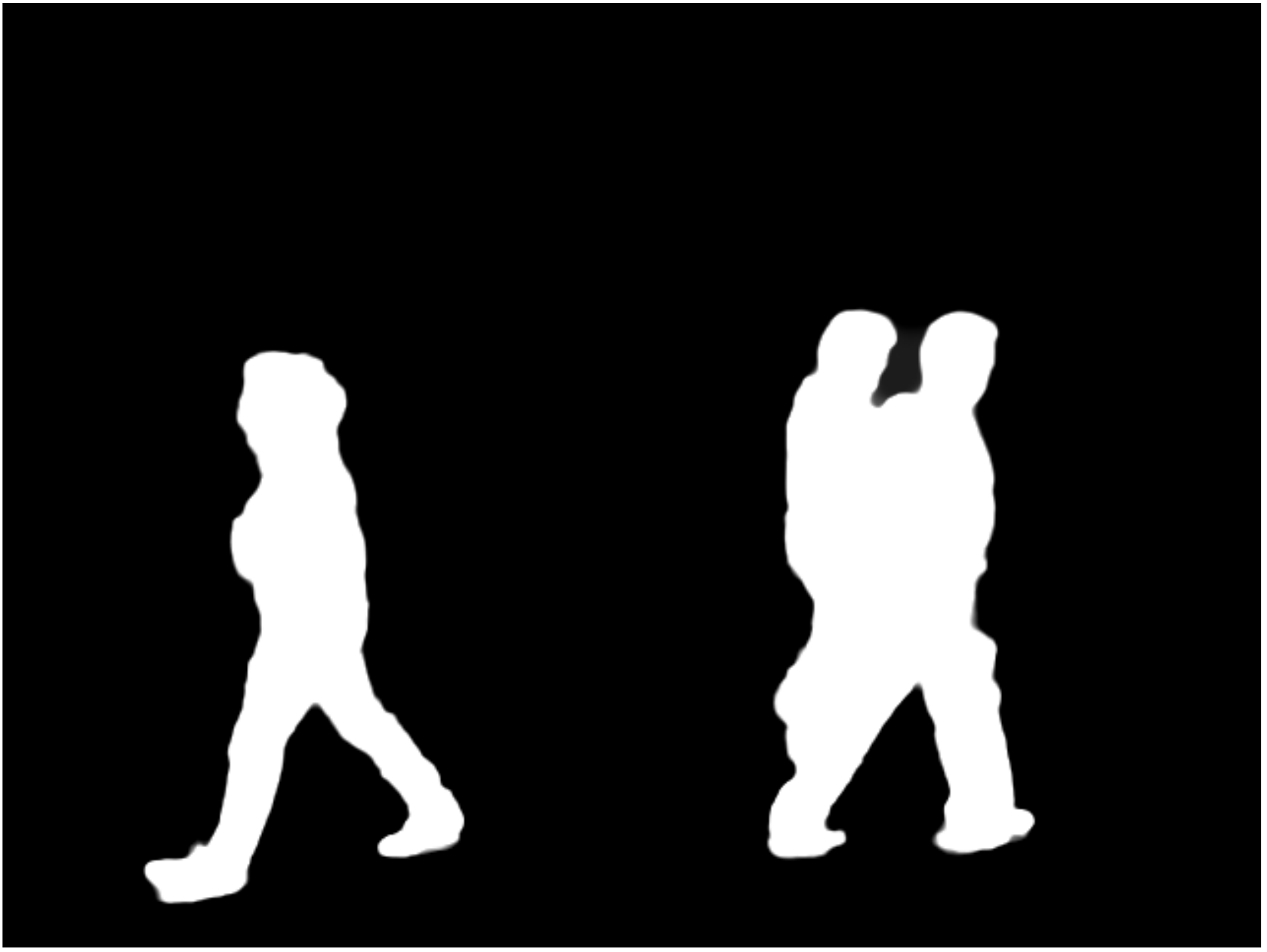}\\
\hspace{-2ex} (f) Learning-based &
\hspace{-2ex} (g) K-nearest   &
\hspace{-2ex} (h) Comprehensive  &
\hspace{-2ex} (i) DCNN \cite{Cho_Tai_Kweon_2016} &
\hspace{-2ex} (j) Ours\\
\hspace{-2ex} matting \cite{Zheng_Kambhamettu_2009} &
\hspace{-2ex} neighbors\cite{Chen_Li_Tang_2013} &
\hspace{-2ex} sampling \cite{Shahrian_Rajan_Price_2013}  &
\hspace{-2ex}  &
\hspace{-2ex} (without trimap)\\
\end{tabular}
\caption{Comparison with existing alpha-matting algorithms on real images with a frame-difference based trimap. (a) Input image. (b) Frame difference. (c) Trimap generated by morphographic operation (dilation / erosion) of the binary mask. (d) - (i) Alpha matting algorithms available on \texttt{alphamatting.com}. (j) Proposed method. This sequence is from the dataset of \cite{Camplani_Maddalena_Alcover_2017}. }
\label{fig:failure example multi people}
\vspace{-2ex}
\end{figure*}

As shown in the examples, error in frame difference can be easily translated to false alarms and misses in the trimap. While we can increase the uncertainty region of the trimap to rely more on the color constancy model of the alpha matting, in general the alpha matting performs worse when the uncertainty region grows. We have also tested more advanced background estimation algorithms, e.g., \cite{Hofmann_Tiefenbacher_Rigoll_2012} in OpenCV. However, the results are similar or sometimes even worse. \fref{fig:failure example multi people} shows a comparison using various alpha matting algorithms.

\section{Multi-Agent Consensus Equilibrium}
Our proposed method is an information fusion technique. The motivation for adopting a fusion strategy is that for complex scenarios, no single estimator can be uniformly superior in all situations. Integrating weak estimators to construct a stronger one is likely more robust and can handle more cases. The weak estimators we use in this paper are the alpha matting, background subtraction and image denoising. We present a principled method to integrate these estimators.

The proposed fusion technique is based on the Multi-Agent Consensus Equilibrium (MACE), recently developed by Buzzard et al. \cite{Buzzard_Chan_Sreehari_2017}. Recall the overview diagram shown in \fref{fig: MACE illustration}. The method consists of three individual agents which perform specific tasks related to our problem. The agents will output an estimate based on their best knowledge and their current state. The information is aggregated by the consensus agents and broadcast back to the individual agents. The individual agents update their estimates until all three reaches a consensus which is the final output. We will discuss the general principle of MACE in this section, and describe the individual agents in the next section.

\subsection{ADMM}
The starting point of MACE is the alternating direction method of multipliers (ADMM) algorithm \cite{Boyd_Parikh_Chu_2011}. The ADMM algorithm aims at solving a constrained minimization:
\begin{equation}
\minimize{\vx_1, \vx_2} \; f_1(\vx_1) + f_2(\vx_2), \quad \subjectto \quad \vx_1 = \vx_2,
\label{eq:constrained problem}
\end{equation}
where $\vx_i \in \R^n$, and $f_i: \R^n \rightarrow \R$ are mappings, typically a forward model describing the image formation process and a prior distribution of the latent image. ADMM solves the problem by solving a sequence of subproblems as follows:
\begin{subequations}
\begin{align}
\vx_{1}^{(k+1)} &= \argmin{\vv\in \R^n} \;\; f_1(\vv) +\frac{\rho}{2} \|\vv - (\vx_{2}^{(k)}-\vu^{(k)})\|^2, \label{eq:ADMM2,x}\\
\vx_{2}^{(k+1)} &= \argmin{\vv\in \R^n} \;\; f_2(\vv) + \frac{\rho}{2}\|\vv - (\vx_{1}^{(k+1)}+\vu^{(k)})\|^2,\label{eq:ADMM2,v}\\
\vu^{(k+1)} &= \vu^{(k)} + (\vx_{1}^{(k+1)} - \vx_{2}^{(k+1)}). \label{eq:ADMM2,u}
\end{align}
\end{subequations}
In the last equation \eref{eq:ADMM2,u}, the vector $\vu^{(k)} \in \R^n$ is the Lagrange multiplier associated with the constraint. Under mild conditions, e.g., when $f_1$ and $f_2$ are convex, close, and proper, global convergence of the algorithm can be proved \cite{Sreehari_Venkatakrishnan_Wohlberg_2015}. Recent studies show that ADMM converges even for some non-convex functions \cite{Chan_Wang_elgendy_2017}.

When $f_1$ and $f_2$ are convex, the minimizations in \eref{eq:ADMM2,x} and \eref{eq:ADMM2,v} are known as the proximal maps of $f_1$ and $f_2$, respectively \cite{Parikh_Boyd_2014}. If we define the proximal maps as
\begin{equation}
F_i(\vz) = \argmin{\vv\in \R^n} \;\; f_i(\vv) +\frac{\rho}{2} \|\vv - \vz\|^2,
\end{equation}
then it is not difficult to see that at the optimal point, \eref{eq:ADMM2,x} and \eref{eq:ADMM2,v} become
\begin{subequations}
\begin{align}
F_1(\vx^* - \vu^*) &= \vx^* \label{eq:CE:x1},\\
F_2(\vx^* + \vu^*) &= \vx^* \label{eq:CE:x2},
\end{align}
\end{subequations}
where $(\vx^*,\vu^*)$ are the solutions to the original constrained optimization in \eref{eq:constrained problem}. \eref{eq:CE:x1} and \eref{eq:CE:x2} shows that the solution $(\vx^*,\vu^*)$ can now be considered as a fixed point of the system of equations.

Rewriting \eref{eq:ADMM2,x}-\eref{eq:ADMM2,u} in terms of \eref{eq:CE:x1} and \eref{eq:CE:x2} allows us to consider agents $F_i$ that are not necessarily proximal maps, i.e., $f_i$ is not convex or $F_i$ may not be expressible as optimizations. One example is to use an off-the-shelf image denoiser for $F_i$, e.g., BM3D, non-local means, or neural network denoisers. Such algorithm is known as the Plug-and-Play ADMM \cite{Sreehari_Venkatakrishnan_Wohlberg_2015, Chan_Wang_elgendy_2017,Venkatakrishnan_Bouman_Wohlberg_2013} (and variations thereafter \cite{Wang_Chan_2017,Buzzard_Chan_Sreehari_2017}).

\subsection{MACE and Intuition}
MACE generalizes the above ADMM formulation. Instead of minimizing a sum of two functions, MACE minimizes a sum of $N$ functions $f_1,\ldots,f_N$:
\begin{equation}
\minimize{\vx_1,\ldots,\vx_N} \;\; \sum_{i=1}^N f_i(\vx_i), \quad \vx_1 = \ldots = \vx_N.
\label{eq: MACE}
\end{equation}
In this case, the equations in \eref{eq:CE:x1}-\eref{eq:CE:x2} are generalized to
\begin{equation}
\begin{array}{ll}
F_i(\vx^* + \vu_i^*) &= \vx^*, \quad \mbox{for } i = 1,\ldots,N\\
\sum_{i=1}^N \vu_i^* &= 0.
\end{array}
\label{eq:CE equation}
\end{equation}
What does \eref{eq:CE equation} buy us? Intuitively, \eref{eq:CE equation} suggests that in a system containing $N$ agents, each agent will create a tension $\vu_i^* \in \R^n$. For example, if $F_1$ is an inversion step whereas $F_2$ is a denoising step, then $F_1$ will not agree with $F_2$ because $F_1$ tends to recover details but $F_2$ tends to smooth out details. The agents $F_1,\ldots,F_N$ will reach an equilibrium state where the sum of the tension is zero. This explains the name ``consensus equilibrium'', as the the algorithm is seeking a consensus among all the agents.

How does the equilibrium solution look like? The following theorem, shown in \cite{Buzzard_Chan_Sreehari_2017}, provides a way to connect the equilibrium condition to a fixed point of an iterative algorithm.
\begin{theorem}[MACE solution \cite{Buzzard_Chan_Sreehari_2017}]
\label{thm:main}
 Let $\underline{\vu}^* \bydef [\vu_1^*; \ldots; \vu_N^*]$. The consensus equilibrium $(\vx^*,\underline{\vu}^*)$ is a solution to the MACE equation \eref{eq:CE equation} if and only if the points $\vv^*_i \bydef \vx^* + \vu_i^*$ satisfy
\begin{align}
\frac{1}{N} \sum_{i=1}^N \vv^*_i &= \vx^* \label{eq:CE condition 1}\\
(2\calG-\calI)(2\calF-\calI)\underline{\vv}^* &= \underline{\vv}^*, \label{eq:CE condition 2}
\end{align}
where $\underline{\vv}^* \bydef [\vv_1^*; \ldots; \vv_N^*] \in \R^{nN}$, and $\calF, \calG: \R^{nN} \rightarrow \R^{nN}$ are mappings defined as
\begin{equation}
\calF(\underline{\vz}) =
\begin{bmatrix}
F_1(\vz_1)\\
\vdots\\
F_N(\vz_N)
\end{bmatrix}
,\quad
\mbox{and}
\quad
\calG(\underline{\vz}) =
\begin{bmatrix}
\langle \underline{\vz} \rangle \\
\vdots\\
\langle \underline{\vz} \rangle
\end{bmatrix},
\label{eq:F and G}
\end{equation}
where $\langle \underline{\vz} \rangle \bydef \frac{1}{N}\sum_{i=1}^N \vz_i$ is the average of $\underline{\vz}$.
\end{theorem}

\begin{algorithm}[h]
\caption{MACE Algorithm}
\begin{algorithmic}[1]
\State Initialize $\underline{\vv}^t = [\vv_1^t,\ldots,\vv_N^t]$.
\For{$t = 1,\ldots,T$}
    \State \% Perform agent updates, $(2\calF-\calI)(\underline{\vv}^t)$
    \State
    \begin{equation}
    \begin{bmatrix}
    \vz_1^t\\
    \vdots\\
    \vz_N^t
    \end{bmatrix} =
    \begin{bmatrix}
    2F_1(\vv_1^t) - \vv_1^t\\
    \vdots\\
    2F_N(\vv_N^t) - \vv_N^t
    \end{bmatrix}
    \end{equation}
    \State
    \State \% Perform the data aggregation $(2\calG-\calI)(\underline{\vz}^t)$
    \State
    \begin{equation}
    \begin{bmatrix}
    \vv_1^{t+1}\\
    \vdots\\
    \vv_N^{t+1}
    \end{bmatrix}
    =
    \begin{bmatrix}
    2\langle \underline{\vz}^{t} \rangle - \vz_1^t\\
    \vdots \\
    2\langle \underline{\vz}^{t} \rangle - \vz_N^t
    \end{bmatrix}
    \end{equation}
\EndFor
\State Output $\langle \underline{\vv}^{T} \rangle$.
\end{algorithmic}
\label{alg:MACE}
\end{algorithm}

Theorem~\ref{thm:main} provides a full characterization of the MACE solution. The operator $\calG$ in Theorem~\ref{thm:main} is a consensus agent that takes a set of inputs $\vz_1,\ldots,\vz_N$ and maps them to their average $\langle \underline{\vz} \rangle$. In fact, we can show that $\calG$ is a projection and that $(2\calG-\calI)$ is its self-inverse\cite{Buzzard_Chan_Sreehari_2017}. As a result, \eref{eq:CE condition 2} is equivalent to $(2\calF-\calI)\underline{\vv}^* = (2\calG-\calI) \underline{\vv}^*$. That is, we want the individual agents $F_1,\ldots,F_N$ to match with the consensus agent $\calG$ such that the equilibrium holds: $(2\calF-\calI)\underline{\vv}^* = (2\calG-\calI) \underline{\vv}^*$.

The algorithm of the MACE is illustrated in Algorithm~\ref{alg:MACE}. According to \eref{eq:CE condition 2}, $\underline{\vv}^*$ is a fixed point of the set of equilibrium equations. Finding the fixed point can be done by iteratively updating $\underline{\vv}^{(t)}$ through the procedure
\begin{equation}
\underline{\vv}^{(t+1)} = (2\calG-\calI)(2 \calF-\calI)\underline{\vv}^{(t)}.
\label{eq:CE iteration}
\end{equation}
Therefore, the algorithmic steps are no more complicated than updating the individual agents $(2\calF-\calI)$ in parallel, and then aggregating the results through $(2\calG-\calI)$.

The convergence of MACE is guaranteed when $\calT$ is non-expansive \cite{Buzzard_Chan_Sreehari_2017}summarized in the proposition below.
\begin{proposition}
Let $\calF$ and $\calG$ be defined as \eref{eq:F and G}, and let $\calT \bydef (2\calG-\calI)(2\calF-\calI)$. Then the following results hold:
\begin{enumerate}
\item[(i)] $\calF$ is firmly non-expansive if all $F_i$'s are firmly non-expansive.
\item[(ii)] $\calG$ is firmly non-expansive.
\item[(iii)] $\calT$ is non-expansive if $\calF$ and $\calG$ are firmly non-expansive.
\end{enumerate}
\end{proposition}
\begin{proof}
See Appendix.
\end{proof}

\section{Designing MACE Agents}
After describing the MACE framework, in this section we discuss how each agent is designed for our problem.

\subsection{Agent 1: Dual-Layer Closed-Form Matting}
The first agent we use in MACE is a modified version of the classic closed-form matting. More precisely, we define the agent as
\begin{equation}
F_1(\vz) = \argmin{\valpha} \;\; \valpha^T \mLtilde \valpha + \lambda_1 (\valpha-\vz)^T \mDtilde (\valpha-\vz),
\label{eq:alpha matting agent}
\end{equation}
where $\widetilde{\mL}$ and $\mDtilde$ are matrices, and will be explained below. The constant $\lambda_1$ is a parameter.

\vspace{2ex}

\noindent\textbf{Review of Closed-Form Matting}. To understand the meaning of \eref{eq:alpha matting agent}, we recall that the classical closed-form matting is an algorithm that tries to solve
\begin{align}
&J(\valpha,\va,\vb) \notag \\
&=\sum_{j\in I} \left(\sum_{i\in w_{j}} \left(\alpha_{i}-\sum_{c} a^{c}_{j}I^{c}_{i}-b_{j} \right)^{2}+\epsilon\sum_{c}(a^{c}_{j})^{2} \right)\label{eq:cfm obj}.
\end{align}
Here, $(a^r,a^g,a^b,b)$ are the linear combination coefficients of the color line model $\alpha_i \approx \sum_{c \in \{r,g,b\}} a^c I_i^c + b$, and $\alpha_i$ is the alpha matte value of the $i$th pixel\cite{Levin_Lischinski_Weiss_2008}. The weight $w_j$ is a $3 \times 3$ window of pixel $j$. With some algebra, we can show that the marginalized energy function $J(\valpha) \bydef \min_{\va,\vb} J(\valpha,\va,\vb)$ is equivalent to
\begin{equation}
J(\valpha) \bydef \min_{\va,\vb} J(\valpha,\va,\vb) = \valpha^T \mL \valpha,
\label{eq:matting laplacian}
\end{equation}
where $\mL \in \R^{n \times n}$ is the so-called matting Laplacian matrix. When trimap is given, we can regularize $J(\valpha)$ by minimizing the overall energy function:
\begin{equation}
\widehat{\valpha} = \argmin{\valpha} \; \valpha^T \mL \valpha + \lambda (\valpha-\vz)^T \mD (\valpha-\vz),
\label{eq:alpha matting 1}
\end{equation}
where $\mD$ is a binary diagonal matrix with entries being one for pixels that are labeled in the trimap, and zero otherwise. The vector $\vz \in \R^n$ contains specified alpha values given by the trimap. Thus, for large $\lambda$, the minimization in \eref{eq:alpha matting 1} will force the solution to satisfy the constraints given by the trimap.

\vspace{2ex}
	\noindent\textbf{Dual-Layer Matting Laplacian $\widetilde{\mL}$}. In the presence of the plate image, we have two pieces of complementary information: $\mI \in \R^{n \times 3}$ the color image containing the foreground object, and $\mP \in \R^{n \times 3}$ the plate image. Correspondingly, we have alpha matte $\valpha^I$ for $\mI$ , and the alpha matte $\valpha^P$ for $\mP$. When $\mP$ is given, we can redefine the color line model as
\begin{equation}
\begin{bmatrix}
\alpha^I_i\\
\alpha^P_i
\end{bmatrix}
\approx
\sum_{c \in \{r,g,b\}}
a^c
\begin{bmatrix}
I^{c}_i\\
P^{c}_i
\end{bmatrix}
+ b.
\label{eq:color line 2}
\end{equation}
In other words, we ask the coefficients $(a^r,a^g,a^b,b)$ to fit simultaneously the actual image $\mI$ and the plate image $\mP$. When \eref{eq:color line 2} is assumed, the energy function $J(\valpha,\va,\vb)$ becomes
\begin{align}
\widetilde{J}(\valpha^I,\valpha^P,\va,\vb)=\sum_{j\in I} \Bigg\{ \sum_{i\in w_{j}} \left(\alpha_{i}^I-\sum_{c} a^{c}_{j}I^{c}_{i}-b_{j} \right)^{2} \notag \\
+ \eta \sum_{i\in w_{j}}\left(\alpha_{i}^P-\sum_{c} a^{c}_{j}P^{c}_{i}-b_{j} \right)^{2} +\epsilon\sum_{c}(a^{c}_{j})^{2} \Bigg\}
\label{eq:cfm_obj_2},
\end{align}
where we added a constant $\eta$ to regulate the relative emphasis between $\mI$ and $\mP$.

\begin{theorem}
\label{prop:alpha matting}
The marginal energy function
\begin{align}
\widetilde{J}(\valpha) \bydef \min_{\va,\vb} \widetilde{J}(\valpha,\vzero,\va,\vb)
\end{align}
can be equivalently expressed as $\widetilde{J}(\valpha) = \valpha^{T} \widetilde{\mL} \valpha$,
where $\widetilde{\mL} \in \R^{n\times n}$ is the modified matting Laplacian, with the $(i,j)$th element
\begin{align}
\widetilde{L}_{i,j} = \sum_{k|(i,j)\in w_{k}}  \Bigg\{ \delta_{ij}-\frac{1}{2|w_{k}|} \Bigg( 1+(\mI_{i}-\vmu_{k})^{T} \notag \\
\left( \mSigma_k - n(1+\eta)\vmu_k \vmu_k^T \right)^{-1} (\mI_{j}-\vmu_{k})\Bigg) \Bigg\}.
\label{eq:L tilde}
\end{align}
Here, $\delta_{ij}$ is the Kronecker delta, $\mI_i \in \R^3$ is the color vector at the $i$th pixel. The vector $\vmu_k \in \R^3$ is defined as
\begin{align}
\vmu_k    = \frac{1}{2|w_k|}\sum_{j \in w_k} (\mI_j + \mP_j),
\end{align}
and the matrix $\mSigma_k \in \R^{3 \times 3}$ is
\begin{align}
\mSigma_k &= \frac{1}{2} \Bigg\{ \frac{1}{|w_k|} \sum_{j \in w_k} (\mI_j - \vmu_k)(\mI_j - \vmu_k)^T \notag \\ &+ \frac{1}{|w_k|} \sum_{j \in w_k}(\mP_j - \vmu_k)(\mP_j - \vmu_k)^T \Bigg\}.
\end{align}
\end{theorem}
\begin{proof}
See Appendix.
\end{proof}

Because of the plate term in \eref{eq:cfm_obj_2}, the modified matting Laplacian $\widetilde{\mL}$ is positive definite. See Appendix for proof. The original $\mL$ in \eref{eq:matting laplacian} is only positive semi-definite.

\vspace{2ex}
\noindent\textbf{Dual-Layer Regularization $\mDtilde$}.
The diagonal regularization matrix $\mDtilde$ in \eref{eq:alpha matting agent} is reminiscent to the binary matrix $\mD$ in \eref{eq:alpha matting 1}, but $\mDtilde$ is defined through a sigmoid function applied to the input $\vz$. To be more precise, we define $\widetilde{\mD} \bydef \mbox{diag}(\widetilde{d}_i)$, where
\begin{equation}
\widetilde{d}_i = \mbox{diag}\left\{ \frac{1}{1+\exp\{-\kappa(z_i - \theta)\}} \right\} \in \R^{n \times n},
\label{eq:sigmoid D}
\end{equation}
and $z_i$ is the $i$-th element of the vector $\vz \in \R^n$, which is the argument of $F_1$. The scalar constant $\kappa > 0$ is a user defined parameter specifying the stiffness of the sigmoid function, and $0 < \theta < 1$ is another user defined parameter specifying the center of the transient. Typical values of $(\kappa,\theta)$ for our MACE framework are $\kappa = 30$ and $\theta = 0.8$.

A closer inspection of $\mD$ and $\mDtilde$ reveals that $\mD$ is performing a hard-threshold whereas $\mDtilde$ is performing a soft-threshold. In fact, the matrix $\mD \bydef \mbox{diag}(d_{i})$ has diagonal entries
\begin{equation}
d_{i} =
\begin{cases}
0, &\quad \theta_1 < z_i < \theta_2, \\
1, &\quad \mbox{otherwise}.
\end{cases}
\label{eq:sigmoid D 2}
\end{equation}
for two cutoff values $\theta_1$ and $\theta_2$. This hard-threshold is equivalent to the soft-threshold in \eref{eq:sigmoid D} when $\kappa \rightarrow \infty$.

There are a few reasons why \eref{eq:sigmoid D} is preferred over \eref{eq:sigmoid D 2}, especially when we have the plate image. First, the soft-threshold in \eref{eq:sigmoid D} tolerates more error present in $\vz$, because the values of $\mDtilde$ represent the probability of having foreground pixels. Second, the one-sided threshold in \eref{eq:sigmoid D} ensures that the background portion of the image is handled by the plate image rather than the input $\vz$. This is usually beneficial when the plate is reasonably accurate.

\subsection{Agent 2: Background Estimator}
Our second agent is a background estimator, defined as
\begin{equation}
F_2(\vz) = \argmin{\valpha} \;\; \|\valpha - \vr_0\|^2 + \lambda_2 \|\valpha-\vz\|^2 + \gamma \valpha^T(1-\valpha).
\label{eq:background estimator}
\end{equation}
The reason of introducing $F_2$ is that in $F_1$, the matrix $\mDtilde$ is determined by the current estimate $\vz$. While $\mDtilde$ handles part of the error in $\vz$, large missing pixels and false alarms can still cause problems especially in the interior regions. The goal of $F_2$ is to complement $F_1$ for these interior regions.

\vspace{2ex}
\noindent \textbf{Initial Background Estimate $\vr_0$}. Let us take a look at \eref{eq:background estimator}. The first two terms are quadratic. The interpretation is that given some fixed initial estimate $\vr_0$ and the current input $\vz$, $F_2(\vz)$ returns a linearly combined estimate between $\vr_0$ and $\vz$. The initial estimate $\vr_0$ consists of two parts:
\begin{equation}
\vr_0 = \vr_c \odot \vr_e,
\end{equation}
where $\odot$ means elementwise multiplication. The first term $\vr_c$ is the \emph{color} term, measuring the similarity between foreground and background colors. The second term $\vr_e$ is the \emph{edge} term, measuring the likelihood of foreground edges relative background edges. In the followings we will discuss these two terms one by one.

\vspace{2ex}
\noindent \textbf{Defining the Color Term $\vr_c$}. We define $\vr_c$ by measuring the distance $\|\mI_j - \mP_j\|^2 = \sum_{c \in \{r,g,b\}} (I_j^c - P_j^c)^2$ between a color pixel $\mI_j \in \R^3$ and a plate pixel $\mP_j \in \R^3$. Ideally, we would like $\vr_c$ to be small when $\|\mI_j - \mP_j\|^2$ is large.

In order to improve the robustness of $\|\mI_j - \mP_j\|^2$ against noise and illumination fluctuation, we modify $\|\mI_j - \mP_j\|^2$ by using the bilateral weighted average over a small neighborhood:
\begin{equation}
\Delta_i =\sum_{j \in \Omega_i} w_{ij} \|\mI_{j}-\mP_{j}\|^{2},
\end{equation}
where $\Omega_i$ specifies a small window around the pixel $i$. The bilateral weight $w_{ij}$ is defined as
\begin{equation}
w_{ij} = \frac{\widetilde{w}_{ij}}{\sum_{j} \widetilde{w}_{ij}},
\end{equation}
where
\begin{equation}
\widetilde{w}_{ij} = \exp\left\{-\frac{\|\vx_i - \vx_j\|^{2}}{2 h_{s}^{2}}\right\} \exp\left\{-\frac{\|\mI_i-\mI_j\|^{2}}{2 h_{r}^{2}} \right\}.
\label{eq:bilateral weight}
\end{equation}
Here, $\vx_i$ denotes the spatial coordinate of pixel $i$, $\mI_i \in \R^3$ denotes the $i$th color pixel of the color image $\mI$, and $(h_s,h_r)$ are the parameters controlling the bilateral weight strength. The typical values of hs and hr are both 5.

We now need a mapping which maps the distance $\mDelta \bydef [\Delta_1,\ldots,\Delta_n]^T$ to a vector of numbers $\vr_c$ in $[0,1]^n$ so that the term $\|\valpha - \vr_0\|^2$ makes sense. To this end, we choose a simple Gaussian function:
\begin{equation}
\vr_c = 1-\exp\left\{ - \frac{\mDelta^2 }{ 2\sigma_{\delta}^2} \right\},
\label{eq:mapping for vrc}
\end{equation}
where $\sigma_\delta$ is a user tunable parameter. We tested other possible mappings such as the sigmoid function and the cumulative distribution function of a Gaussian. However, we do not see significant difference compared to \eref{eq:mapping for vrc}. The typical value for $\sigma_\delta$ is 10.

\begin{figure}[t]
\centering
\includegraphics[width=\linewidth]{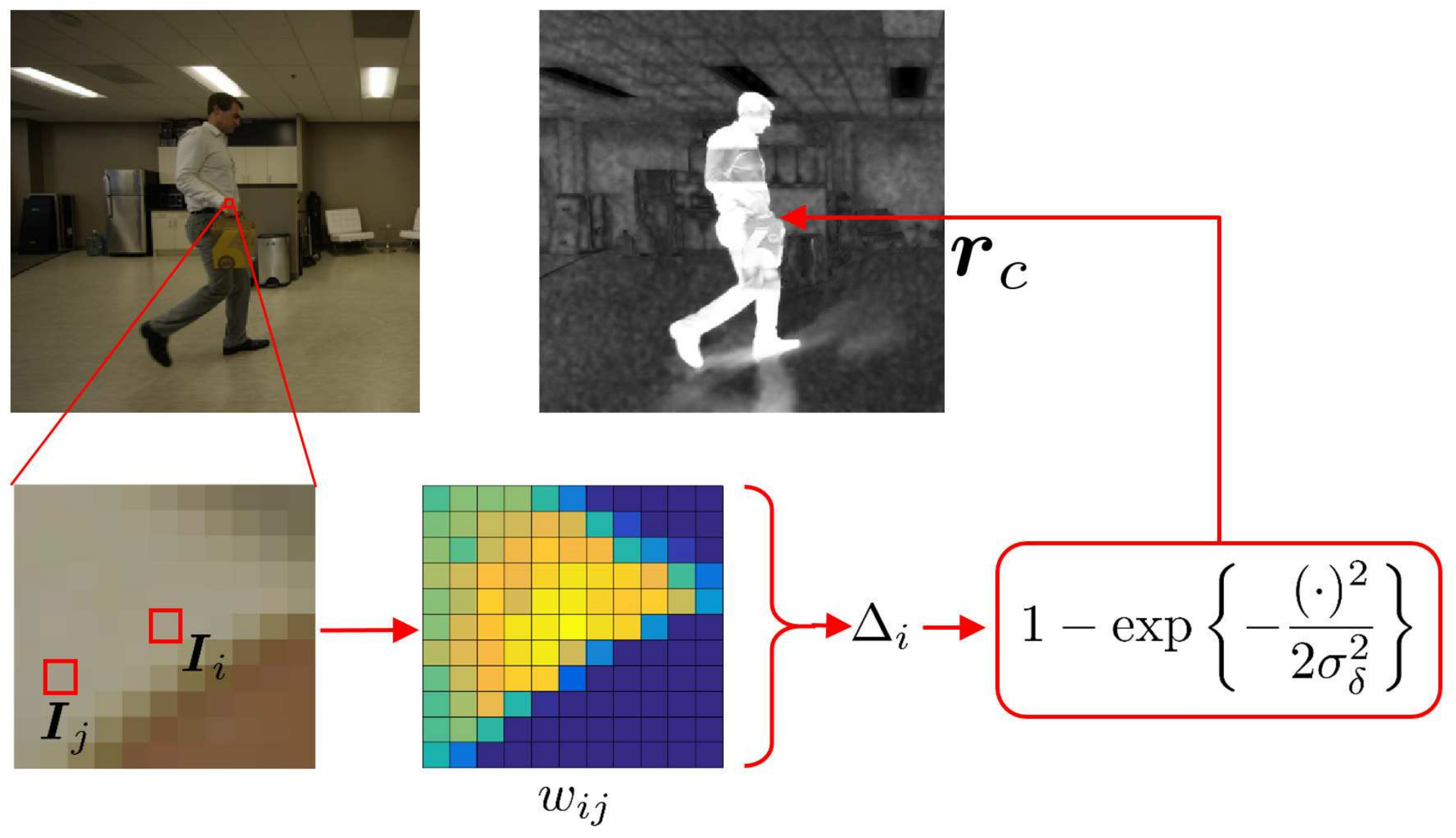}
\vspace{-4ex}
\caption{Illustration of how to construct the estimate $\vr_c$. We compute the distance between the foreground and the background. The distance has a bilateral weight to improve robustness. The actual $\vr_0$ represents the probability of having a foreground pixel.}
 \label{fig:pic1}
\end{figure}

\vspace{2ex}
\noindent \textbf{Defining the Edge Term $\vr_e$}.
The color term $\vr_c$ is able to capture most of the difference between the image and the plate. However, it also generates false alarms if there is illumination change. For example, shadow due to the foreground object is often falsely labeled as foreground. See the shadow near the foot in \fref{fig:pic1}.

In order to reduce the false alarm due to minor illumination change, we first create a ``super-pixel'' mask by grouping similar colors. Our super-pixels are generated by applying a standard flood-fill algorithm \cite{Burtsev_Kuzmin_1993} to the image $\mI$. This gives us a partition of the image $\mI$ as
\begin{equation}
\mI \rightarrow \{\mI^{S_1}, \mI^{S_2}, \ldots, \mI^{S_m}\},
\end{equation}
where $S_1, \ldots, S_m$ are the $m$ super-pixel index sets. The plate image is partition using the same super-pixel indices, i.e., $\mP \rightarrow \{\mP^{S_1}, \mP^{S_2}, \ldots, \mP^{S_m}\}$.

While we are generating the super-pixels, we also compute the gradients of $\mI$ and $\mP$ for every pixel $i = 1,\ldots, n$. Specifically, we define $\nabla\mI_{i} = [\nabla_x \mI_i, \nabla_y \mI_i]^T$ and $\nabla\mP_{i} = [\nabla_x \mP_i, \nabla_y \mP_i]^T$, where $\nabla_x \mI_i \in \R^3$ (and $\nabla_y \mI_i \in \R^3$) are the two-tap horizontal (and vertical) finite difference at the $i$-th pixel. To measure how far $\mI_i$ is from $\mP_i$, we compute
\begin{align}
\theta_i = \|\nabla \mI_i - \nabla \mP_i\|_2.
\end{align}
Thus, $\theta_i$ is small for background regions because $\mI_i \approx \mP_i$, but is large when there is a foreground pixel in $\mI_i$. If we set a threshold operation after $\theta_i$, i.e., set $\theta_i = 1$ if $\theta_i > \tau_\theta$ for some threshold $\tau_\theta$, then shadows can be removed as their gradients are weak.

\begin{figure}[t]
\centering
\includegraphics[width=1\linewidth]{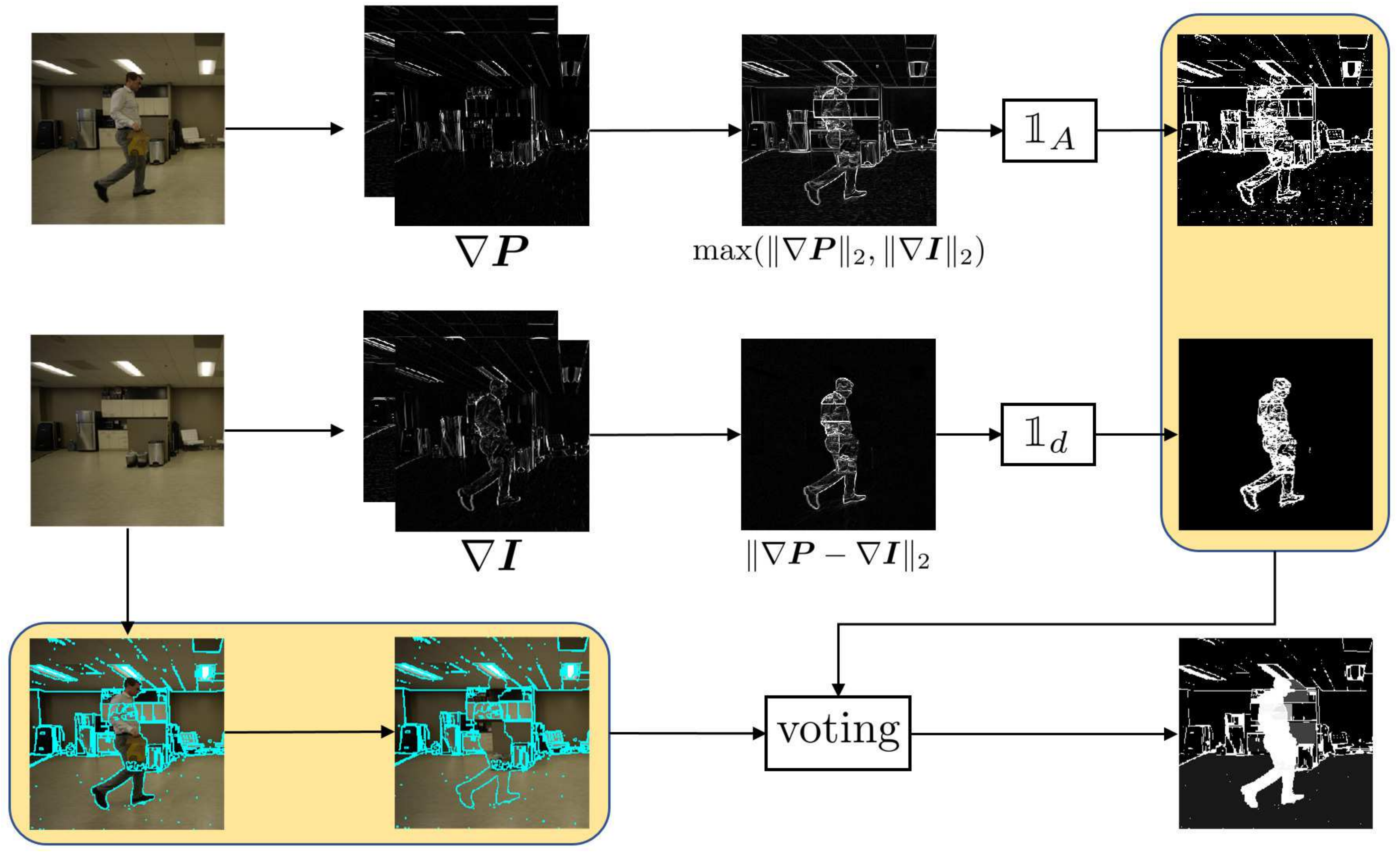}
\vspace{-4ex}
\caption{Illustration of how to construct the estimate $\vr_e$.}
 \label{fig:pic2}
\end{figure}

Now that we have computed $\theta_i$, we still need to map it back to a quantity similar to the alpha matte. To this end, we compute a normalization term
\begin{equation}
A_i = \max\left(\|\nabla \mI_i\|_2, \|\nabla \mP_i \|_2\right),
\end{equation}
and normalize $\mathbbm{1}\{\theta_{i}>\tau_\theta\}$ by
\begin{equation}
(r_{e})_i \bydef \frac{\sum_{j\in S_{i}}\mathbbm{1}\{A_{i} > \tau_A\} \mathbbm{1}\{\theta_{i}>\tau_\theta\}}{\sum_{j\in S_{i}}\mathbbm{1}\{A_{i} > \tau_A\}},
\label{eq:re}
\end{equation}
where $\mathbbm{1}$ denotes the indicator function, and $\tau_A$ and $\tau_\theta$ are thresholds. In essence, \eref{eq:re} says in the $i$-th super-pixel $S_i$, we count the number of edges $\mathbbm{1}\{\theta_{i}>\tau_\theta\}$ that have strong difference between $\mI_i$ and $\mP_i$. However, we do not want to count every pixel but only pixels that already contains strong edges, either in $\mI$ or $\mP$. Thus, we take the weighted average using $\mathbbm{1}\{A_{i} > \tau_A\}$ as the weight. This defines $\vr_e$, as the weighted average $(r_{e})_i$ is shared among all pixels in the super-pixel $S_i$. \fref{fig:pic2} shows a pictorial illustration.

\begin{figure}[t]
\centering
\begin{tabular}{ccc}
\includegraphics[width=0.3\linewidth]{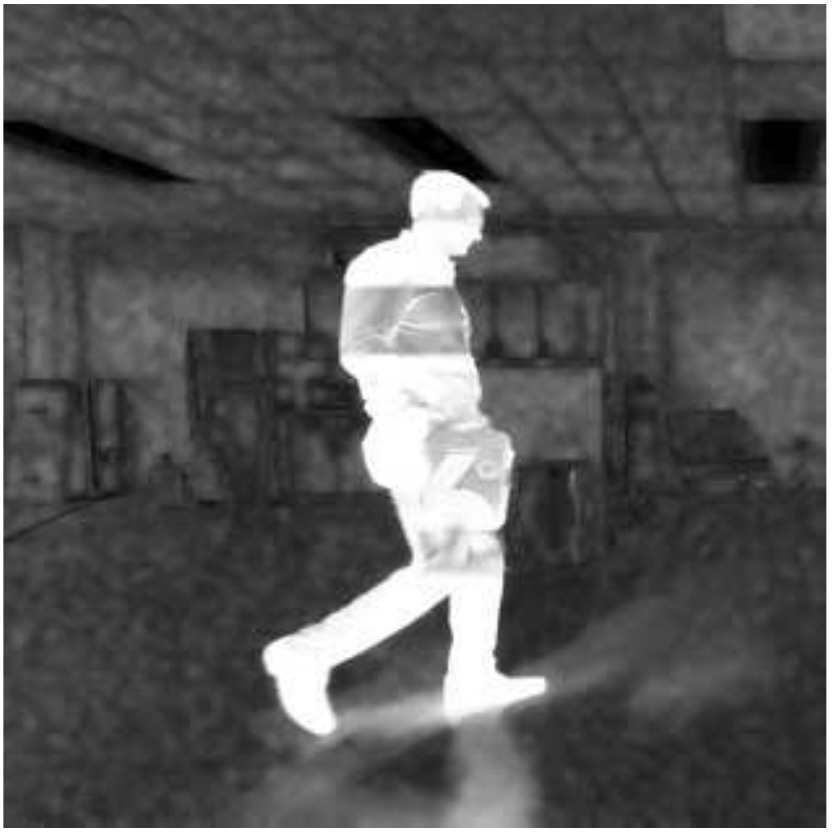}&
\hspace{-2ex}\includegraphics[width=0.3\linewidth]{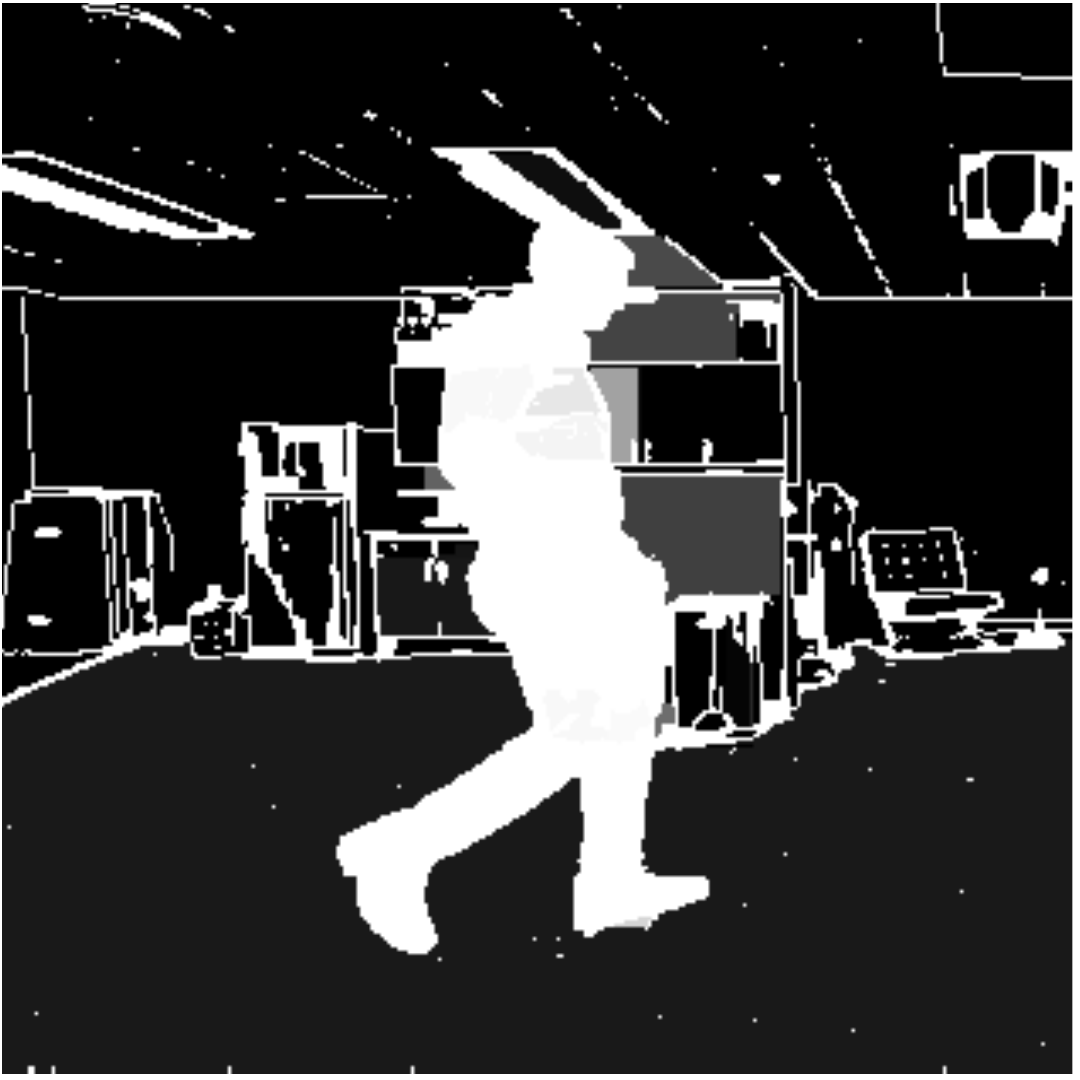}&
\hspace{-2ex}\includegraphics[width=0.3\linewidth]{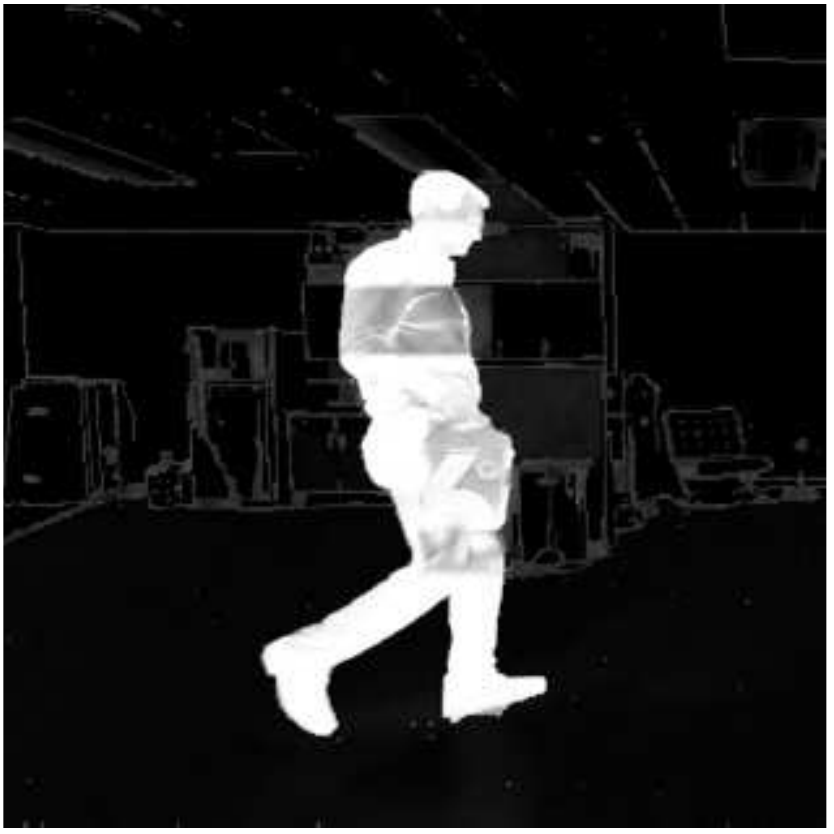}\\
(a) $\vr_c$ &(b) $\vr_e$ &(c) $\vr_0$
\end{tabular}
\caption{Comparison between $\vr_c$, $\vr_e$, and $\vr_0$.}
\label{fig:pic3}
\vspace{-2ex}
\end{figure}

Why is $\vr_e$ helpful? If we look at $\vr_c$ and $\vr_e$ in \fref{fig:pic3}, we see that the foreground pixels of $\vr_c$ and $\vr_e$ coincide but background pixels roughly cancel each other. The reason is that while $\vr_c$ creates weak holes in the foreground, $\vr_e$ fills the gap by ensuring the foreground is marked.

\vspace{2ex}
\noindent \textbf{Regularization $\valpha^T(1-\valpha)$}.
The last term $\valpha^T(1-\valpha)$ in \eref{eq:background estimator} is a regularization to force the solution to either 0 or 1. The effect of this term can be seen from the fact that $\valpha^T(1-\valpha)$ is a symmetric concave quadratic function with a value zero for $\valpha = \vone$ or $\valpha = \vzero$. Therefore, it introduces penalty for solutions that are away from 0 or 1. For $\gamma \le 1$, one can show that the Hessian matrix of the function $f_2(\valpha) = \|\valpha - \vr_0\|^2  + \gamma \valpha^T(1-\valpha)$ is positive semidefinite. Thus, $f_2$ is strongly convex with parameter $\gamma$.

\subsection{Agent 3: Total Variation Denoising}
The third agent we use in this paper is the total variation denoising:
\begin{equation}
F_3(\vz) = \argmin{\valpha} \;\; \|\valpha\|_{\mathrm{TV}}  + \lambda_3 \|\valpha - \vz\|^2,
\label{eq:tv denoising}
\end{equation}
where $\lambda_3$ is a parameter. The norm $\|\cdot\|_{\mathrm{TV}}$ is defined in space-time:
\begin{equation}
\|\vv\|_{\mathrm{TV}} \bydef \sum_{i,j,t} \sqrt{ \beta_x (\nabla_x \vv)^2 + \beta_y (\nabla_y \vv)^2 + \beta_t (\nabla_t \vv)^2},
\label{eq:tv minimization}
\end{equation}
where $(\beta_x,\beta_y,\beta_t)$ controls the relative strength of the gradient in each direction. In this paper, for spatial total variation we set $(\beta_x,\beta_y,\beta_t) = (1,1,0)$, and for spatial-temporal total variation we set $(\beta_x,\beta_y,\beta_t) = (1,1,0.25)$.

\begin{figure}[h]
\centering
\begin{tabular}{cccc}
\includegraphics[width=0.23\linewidth]{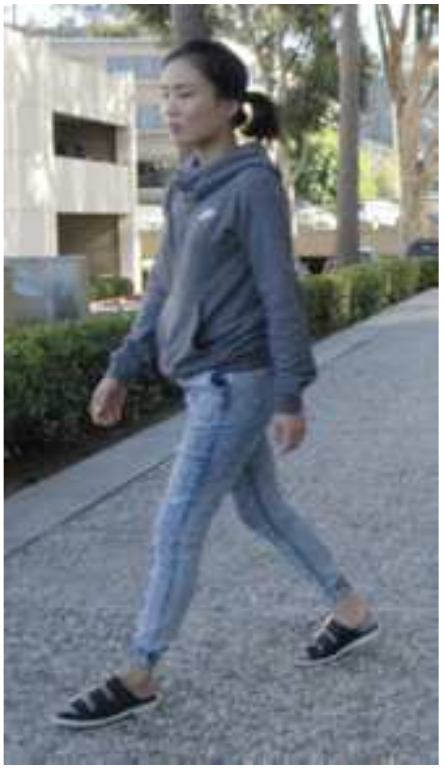}&
\hspace{-2ex}\includegraphics[width=0.23\linewidth]{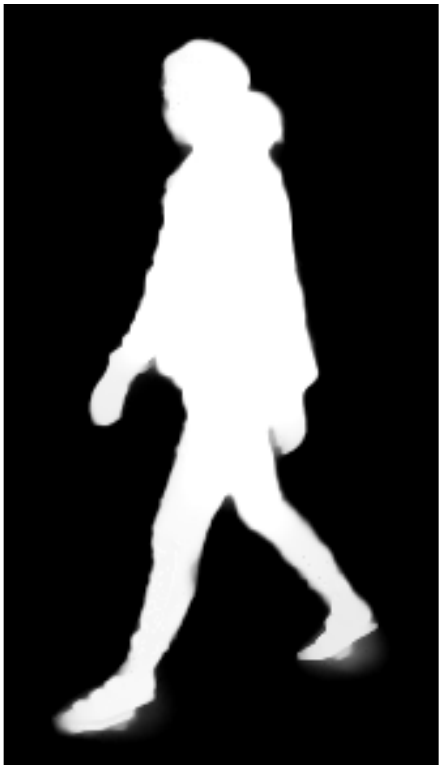}&
\hspace{-2ex}\includegraphics[width=0.23\linewidth]{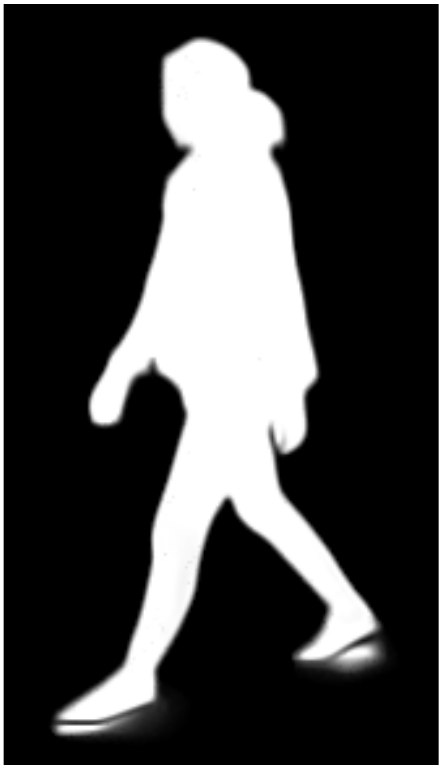}&
\hspace{-2ex}\includegraphics[width=0.23\linewidth]{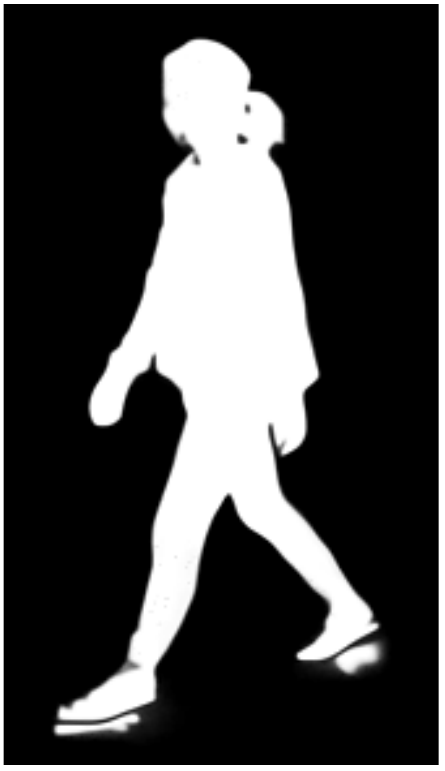}\\
(a) Input   &
\hspace{-2ex}(b) TV &
\hspace{-2ex}(c) BM3D &
\hspace{-2ex}(d) IRCNN \\
&
\hspace{-2ex}\cite{Chan_Khoshabeh_Gibson_2011} &
\hspace{-2ex}\cite{Dabov_Foi_Katkovnik_2007}  &
\hspace{-2ex}\cite{Zhang_Zuo_Gu_2017}
\end{tabular}
\caption{Comparison of different denoisers used in MACE. Shown are the results when MACE converges. The shadow near the foot is a typical place of false alarm, and many denoisers cannot handle.}
 \label{fig:different denoisers}
 \vspace{-4ex}
\end{figure}

A denoising agent is used in the MACE framework because we want to ensure smoothness of the resulting matte. The choice of the total variation denoising operation is a balance betweeen complexity and performance. Users can use stronger denoisers such as BM3D. However, these patch based image denoising algorithms rely on the patch matching procedure, and so they tend to under-smooth repeated patterns of false alarm / misses. Neural network denoisers are better candidates but they need to be trained with the specifically distorted alpha mattes. From our experience, we do not see any particular advantage of using CNN-based denoisers. \fref{fig:different denoisers} shows some comparison.

\subsection{Parameters and Runtime}
The typical values for parameters of the proposed method are presented in \tref{table:para values}. $\lambda_1$ and $\lambda_2$ are rarely changed, while $\lambda_3$ determines the denoising strength of Agent 3. $\gamma$ has a default value of 0.05. Inceasing $\gamma$ causes more binary results with clearer boundaries. $\tau_A$ and $\tau_\theta$ determine the edge term $\vr_e$ in Agent 2 and are fixed. $\sigma_\delta$ determines the color term $\vr_c$ in Agent 2. Large $\sigma_\delta$ produces less false negative but more false positive. Overall, the performance is reasonably stable to these parameters.

\begin{table}[H]
\centering
\label{table:para values}
\vspace{-3ex}
\caption{Typical values for parameters}
\begin{tabular}{cccccccc}
\hline
\hline
 Parameter      &$\lambda_1$& $\lambda_2$ &$\lambda_3$ & $\gamma$ & $\tau_A$ & $\tau_\theta$ &$\sigma_\delta$\\
\hline
 Value       & 0.01 & 2 & 4 & 0.05 & 0.01 &0.02 &10\\
\hline
\end{tabular}
\vspace{-1ex}
\end{table}

\begin{table*}[!]
\centering
\caption{Description of the video sequences used in our experiments.}
\footnotesize{
\begin{tabular}{c|c|c|c|c|c|c|c|c|c|c|c}
\hline
                                 &                   &             &          & time/Fr             &indoor/  &           &lighting    & Backgrd        &            &green   &ground\\
                                 &                   & resolution  & FGD \%   & (sec)   &outdoor  &shadow     &issues    &vibration   & camouflage       &screen  &truth \\
                          \hline
                                 & Book              & 540x960     & 19.75\%  &231  &outdoor  &           &          &$\checkmark$   & $\checkmark$ &         & $\checkmark$\\
                                 & Building          & 632x1012    & 4.03\%   &170.8    &outdoor  &$\checkmark$  &          &$\checkmark$   & $\checkmark$ &         &$\checkmark$      \\
                                 & Coach             & 790x1264    & 4.68\%   &396.1  &outdoor  &           &          &$\checkmark$   &           &$\checkmark$& $\checkmark$     \\
                      Purdue     & Studio            & 480x270     & 55.10\%  &58.3  &indoor   &           &          &            &           &         &  $\checkmark$     \\
                      Dataset    & Road              & 675x1175    & 1.03\%   &232.9  &outdoor  &$\checkmark$  &          &$\checkmark$   & $\checkmark$ &         & $\checkmark$       \\
                                 & Tackle            & 501x1676    & 4.80\%   &210.1  &outdoor  &$\checkmark$  &          &$\checkmark$   &           &$\checkmark$&  $\checkmark$    \\
                                 & Gravel            & 790x1536    & 2.53\%   &280.1  &outdoor  &$\checkmark$  &          &$\checkmark$   & $\checkmark$ &         &  $\checkmark$    \\
                                 & Office            & 623x1229    & 3.47\%   &185.3  &indoor   &$\checkmark$  &          &            & $\checkmark$ &         &   $\checkmark$    \\
        \hline
                                 &Bootstrap          & 480x640    & 13.28\%   &109.1 &indoor   &$\checkmark$  &$\checkmark$ &            & $\checkmark$ &         &$\checkmark$\\
                                 &Cespatx            & 480x640    & 10.31\%   &106.4 &indoor   &$\checkmark$  &$\checkmark$ &            & $\checkmark$ &         & $\checkmark$       \\
                       Public    &DCam               & 480x640    & 12.23\%   &123.6 &indoor   &$\checkmark$  &$\checkmark$ &            & $\checkmark$ &         & $\checkmark$       \\
                       Dataset   &Gen                & 480x640    & 10.23\%   &100.4 &indoor   &$\checkmark$  &$\checkmark$ &            & $\checkmark$ &         &  $\checkmark$       \\
                       \cite{Camplani_Maddalena_Alcover_2017}          &Multipeople        & 480x640    & 9.04\%    &99.5 &indoor   &$\checkmark$  &$\checkmark$ &            & $\checkmark$ &         &  $\checkmark$       \\
                                 &Shadow             & 480x640    & 11.97\%   &115.2 &indoor   &$\checkmark$  &$\checkmark$ &            &           &         &  $\checkmark$       \\
                                         \hline
\end{tabular}}
\label{table:dataset}
\vspace{-2ex}
\end{table*}

In terms of runtime, the most time-consuming part is Agent 1 because we need to solve a large-scale sparse least squares problem. Its runtime is determined by the number of foreground pixels. \tref{table:dataset} shows the runtime of the sequences we tested. In generating these results, we used an un-optimized MATLAB code on a Intel i7-4770k. The typical runtime is about 1-3 minutes per frame. From our experience working with professional artists, even with professional film production software, e.g., NUKE, it takes 15 minutes to label a ground truth label using the plate and temporal cues. Therefore, the runtime benefit offered by our algorithm is substantial. The current runtime can be significantly improved by using multi-core CPU or GPU. Our latest implementation on GPU achieves 5 seconds per frame for images of size $1280 \times 720$.

\section{Experimental Results}
\subsection{Dataset}
\begin{figure}[h]
\centering
\begin{tabular}{c}
\includegraphics[width=0.95\linewidth]{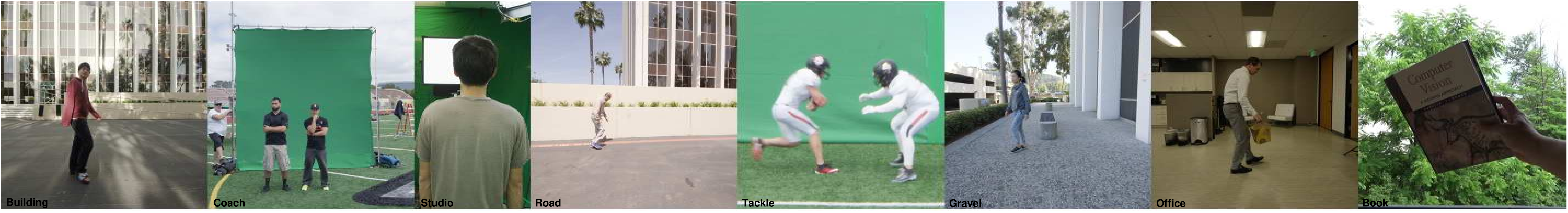}\\
\footnotesize{(a) Snapshots of the Purdue Dataset}\\
\includegraphics[width=0.95\linewidth]{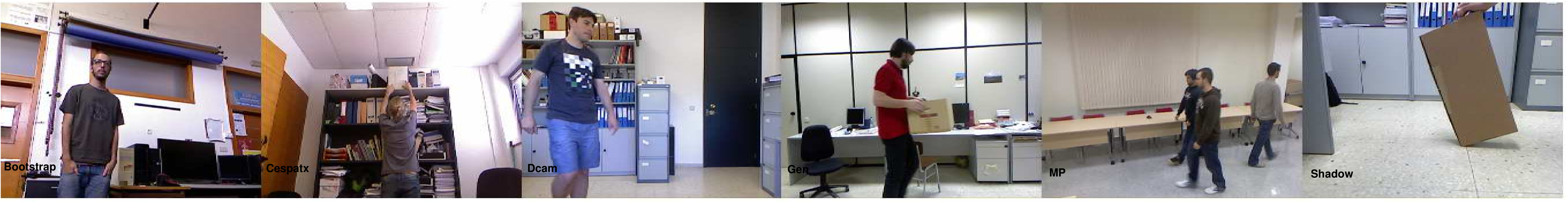}\\
\footnotesize{(b) Snapshots of a public dataset \cite{Camplani_Maddalena_Alcover_2017}}
\end{tabular}
\caption{Snapshots of the videos we use in the experiment. Top row: Building, Coach, Studio, Road, Tackle, Gravel, Office, Book. Bottom row:Bootstrap, Cespatx, Dcam, Gen, MP, Shadow.}
\label{fig:dataset}
\end{figure}

To evaluate the proposed method, we create a Purdue dataset containing 8 video sequences using the HypeVR Inc. 360 degree camera. The original image resolution is $8192 \times 4320$ at a frame rate of 48fps, and these images are then downsampled and cropped to speed up the matting process. In addition to these videos, we also include 6 videos sequences from a public dataset\cite{Camplani_Maddalena_Alcover_2017}, making a total of 14 video sequences. Snapshots of the sequences are shown in \fref{fig:dataset}. All video sequences are captured without camera motion. Plate images are available, either during the first or the last few frames of the video. To enable objective evaluation, for each video sequence we randomly select 10 frames and manually generate the ground truths. Thus totally there are 140 frames with ground truths.

The characteristics of the dataset is summarized in Table~\ref{table:dataset}. The Purdue dataset has various resolution, and the Public dataset has one resolution $480 \times 640$. The foreground percentage for the Purdue dataset videos ranges from $1.03\%$ to $55.10\%$, whereas that public dataset has similar foreground percentage around $10\%$. The runtime of the algorithm (per frame) is determined by the resolution and the foreground percentage. In terms of content, the Purdue dataset focuses on outdoor scenes whereas the public dataset are only indoor. The \texttt{shadow} column indicates the presence of shadow. \texttt{Lighting issues} include illumination change due to auto-exposure and auto-white-balance. The \texttt{background vibration} only applies to outdoor scenes where the background objects have minor movements, e.g., moving grass or tree branches. The \texttt{camouflage} column indicates the similarity in color between the foreground and background, which is a common problem for most sequences. The \texttt{green screen} column shows which of the sequences have green screens to mimic the common chroma-keying environment.

\subsection{Competing methods}
We categorize the competing methods into four different groups. The key ideas are summarized in \tref{table:methods}.

\begin{table*}[!]
\centering
\footnotesize{
\caption{Description of the competing methods.}
\vspace{-2ex}
\begin{tabular}{c|c|c|c|c}
\hline
&Methods                               &          &Supervised   &Key idea\\
\hline

   \multirow{4}{*}{\begin{tabular}{@{}c@{}}Unsupervised \\Video\\ Segmentation\end{tabular}}
   &NLVS  &\cite{Faktor_Irani_2014}             &no    &non-local voting \\
      &AGS  &\cite{Wang_Song_Zhao_2019}         &no    &visual attention \\
      &MOA  &\cite{Dehghan_Zhang_Siam_2019}     &no    &2-stream adaptation \\
       &PDB  &\cite{Song_Wang_Zhao_2018}    &no    &pyramid ConvLSTM \\
\hline
Alpha       &\multirow{2}{*}{Matting}& \multirow{2}{*}{\cite{Cho_Kim_Tai_2017}  }   &\multirow{2}{*}{trimap}    &Trimap generation  \\
  matting                       &               &                               &                           & + alpha matting \\

\hline
Background     &ViBe  &\cite{Barnich_Marc_2011} &no   &pixel model based \\

subtraction      &PBAS  &\cite{Hofmann_Tiefenbacher_Rigoll_2012} &no    &non-parametric \\
\hline
\multirow{2}{*}{Other}
&BSVS     &{\cite{Marki_Perazzi_Wang_2016}} &{key frame}    &bilateral space \\
&Grabcut  &{\cite{Rother_Kolmogorov_Blake_2004}} &{plate}    &iterative graph cuts \\
\hline
\end{tabular}}
\vspace{-2ex}
\label{table:methods}
\end{table*}

\vspace{-2ex}
\textcolor{black}{
\begin{itemize}
\item \textbf{Video Segmentation}: We consider four unsupervised video segmentation methods: Visual attention (AGS)\cite{Wang_Song_Zhao_2019}, pyramid dilated bidirectional ConvLSTM (PDB)\cite{Song_Wang_Zhao_2018}, motion adaptive object segmentation (MOA)\cite{Dehghan_Zhang_Siam_2019}, non-local consensus voting (NLVS) \cite{Faktor_Irani_2014}. These methods are fully-automatic and do not require a plate image. All algorithms are downloaded from the author's websites and are run under default configurations.
\item \textbf{Background Subtraction}: We consider two background subtraction algorithms Pixel-based adaptive segmenter (PBAS) \cite{Hofmann_Tiefenbacher_Rigoll_2012}, Visual background extractor (ViBe) \cite{Barnich_Marc_2011}. Both algorithms are downloaded from the author's websites and are run under default configurations.
\item \textbf{Alpha matting}: We consider one of the state-of-the-art alpha matting algorithm using CNN \cite{Cho_Tai_Kweon_2016}. The trimaps are generated by applying frame difference between the plate and color images, followed by morphological and thresholding operations.
\item \textbf{Others}: We consider the bilateral space video segmentation (BSVS) \cite{Marki_Perazzi_Wang_2016} which is a semi-supervised method. It requires the user to provide ground truth labels for key frames. We also modified the original Grabcut \cite{Rother_Kolmogorov_Blake_2004} to use the plate image instead of asking for user input.
\end{itemize}
}

\subsection{Metrics}
The following four metrics are used.

\noindent $\bullet$ \textcolor{black}{\textbf{Intersection-ver-union (IoU)} measures the overlap between the estimate mask and the ground truth mask:
\begin{equation*}
\mathrm{IoU} = \frac{\sum_{i} \min{\left( \widehat{x}_i,x_i \right)}}{\sum_{i} \max{\left( \widehat{x}_i,x_i \right)}},
\end{equation*}
where $\widehat{x}_i$ is the $i$-pixel of the estimated alpha matte, and $x_i$ is that of the ground truth. Higher IoU score is better.}

\noindent \textcolor{black}{\noindent $\bullet$ \textbf{Mean-absolute-error (MAE)} measures the average absolute difference between the ground truth and the estimate. Lower MAE is better.}

\noindent \textcolor{black}{\noindent $\bullet$ \textbf{Contour accuracy (F)}\cite{Perazzi_Pont_Mcwilliams_2016} measures the performance from a contour based perspective. Higher F score is better.}

\noindent \textcolor{black}{\noindent $\bullet$ \textbf{Structure measure (S)}\cite{Fan_Chen_Liu_2017} simultaneously evaluates region-aware and object-aware structural similarity between the result and the ground truth. Higher S score is better.}

\noindent \textcolor{black}{\noindent $\bullet$ \textbf{Temporal instability (T)}\cite{Perazzi_Pont_Mcwilliams_2016} that performs contour matching with polygon representations between two adjacent frames. Lower T score is better.}

\subsection{Results}

\vspace{2ex}
\noindent $\bullet$ \textbf{Comparison with video segmentation methods}: The results are shown in \tref{table:full chart}, where we list the average IoU, MAE, F, S and T scores over the datasets. In this table, we notice that the deep-learning solutions AGS \cite{Wang_Song_Zhao_2019}, MOA \cite{Dehghan_Zhang_Siam_2019} and PDB \cite{Song_Wang_Zhao_2018} are significantly better than classical optical flow based NLVS \cite{Faktor_Irani_2014} in all the metrics. However, since the deep-learning solutions are targeting for saliency detection, foreground but unsalient objects will be missed. AGS performs the best among the three with a F measure of 0.91, S measure of 0.94 and T measure of 0.19. PDB performs better than MOA in most metrics other than the T measure, with PDB scoring 0.2 while MOA scoreing 0.19.

\begin{figure}[h]
\centering
\begin{tabular}{ccc}
\hspace{-2ex}\includegraphics[width=0.3\linewidth]{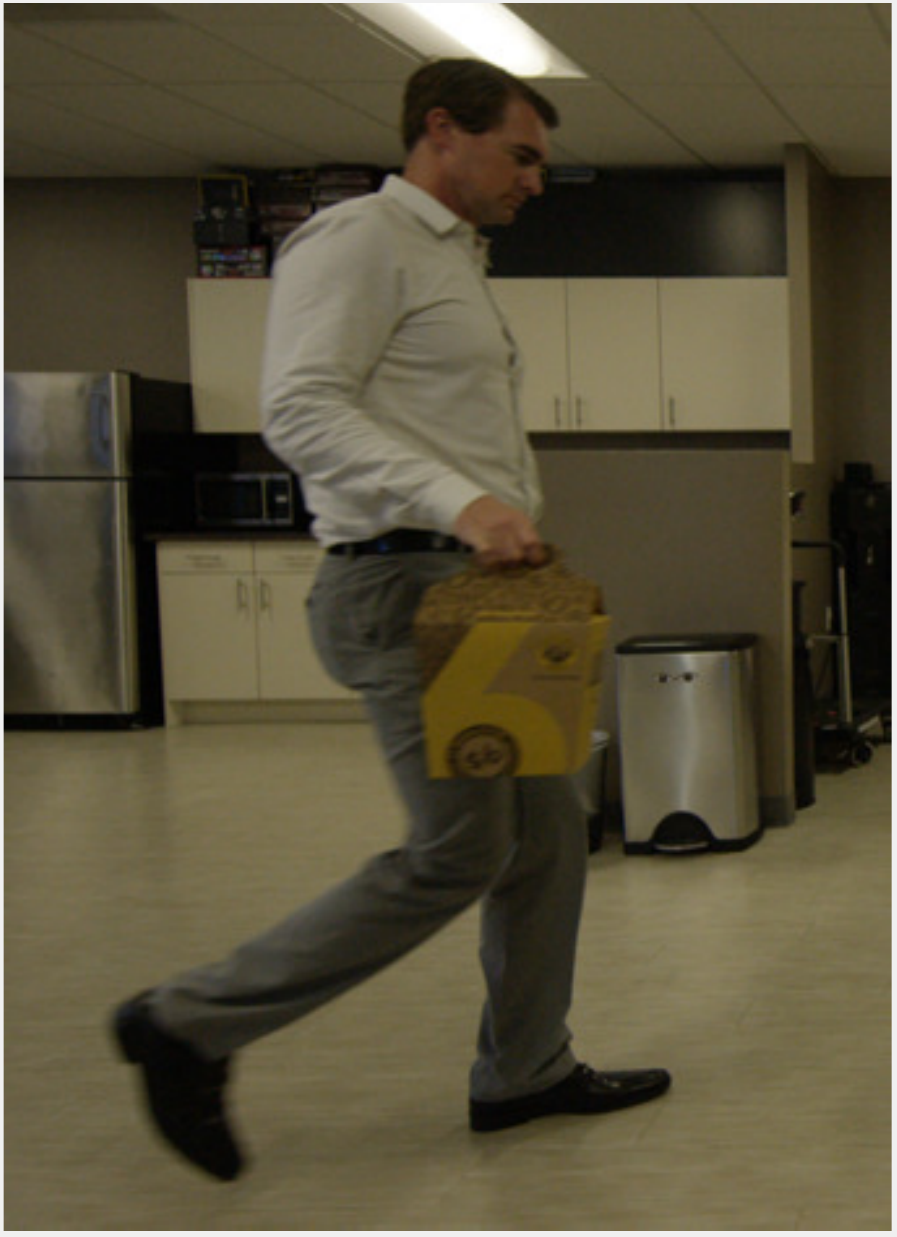}&
\hspace{-2ex}\includegraphics[width=0.3\linewidth]{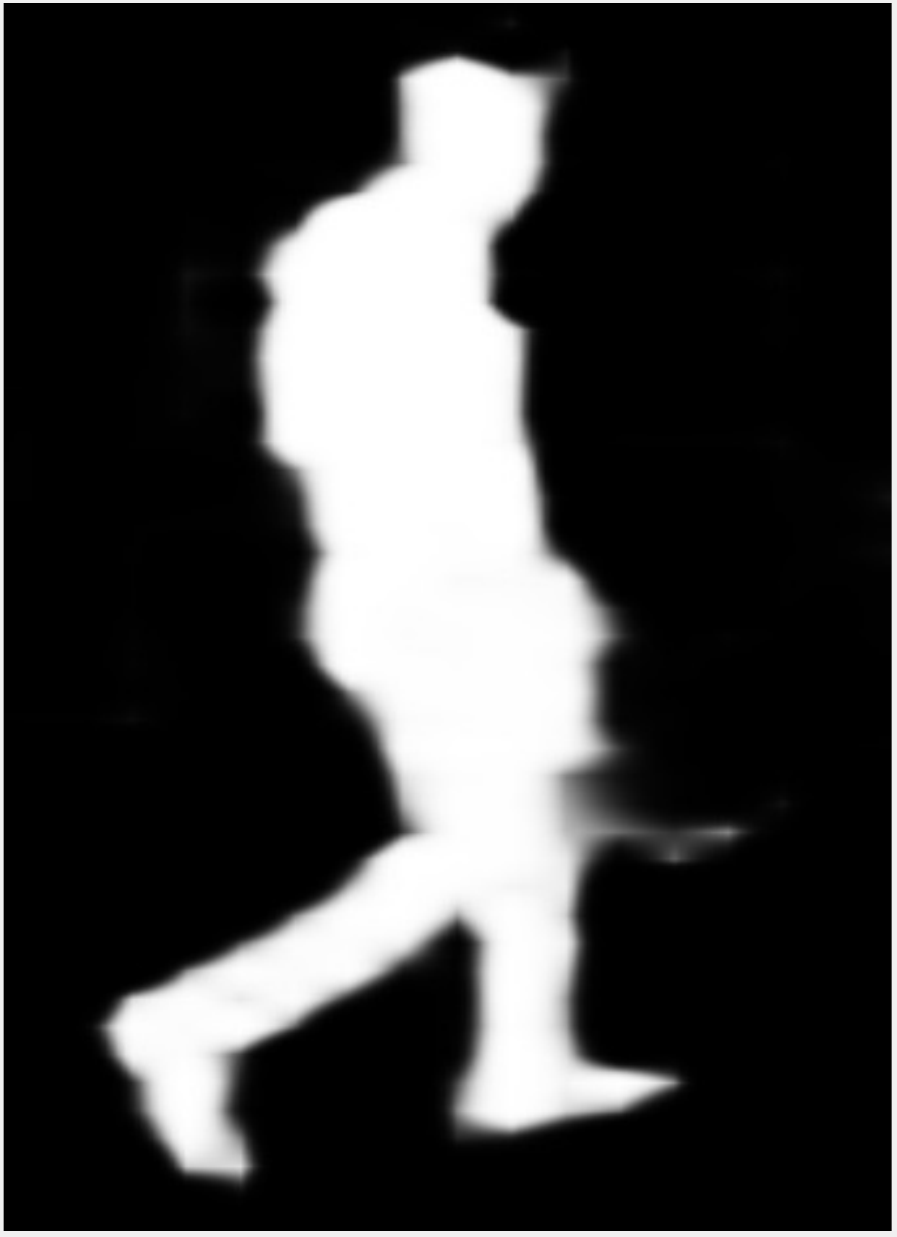}&
\hspace{-2ex}\includegraphics[width=0.3\linewidth]{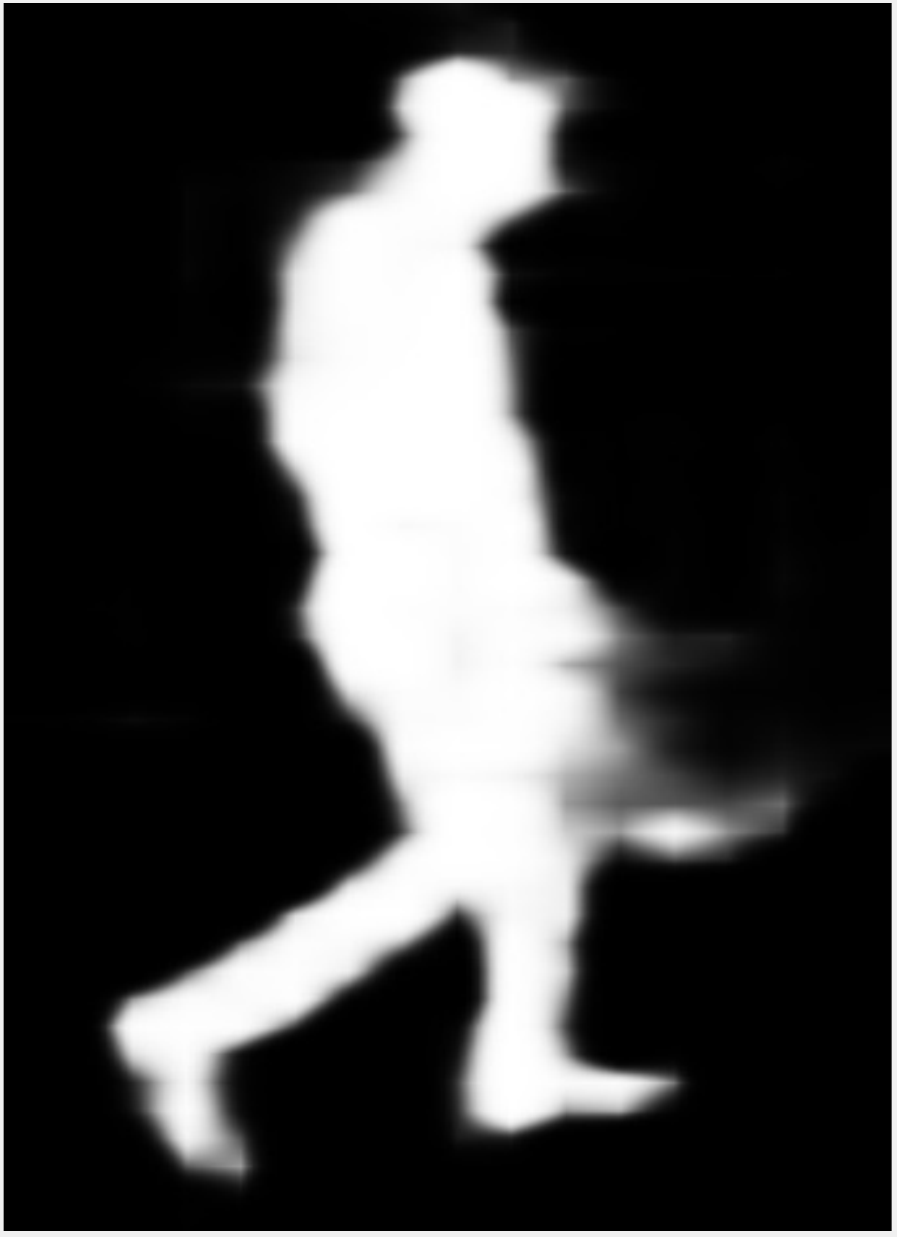}\\
\hspace{-2ex} \footnotesize{(a) Image} & \footnotesize{(b) AGS \cite{Wang_Song_Zhao_2019}, before} &\footnotesize{(c) PDB \cite{Song_Wang_Zhao_2018}, before}\\
\hspace{-2ex}\includegraphics[width=0.3\linewidth]{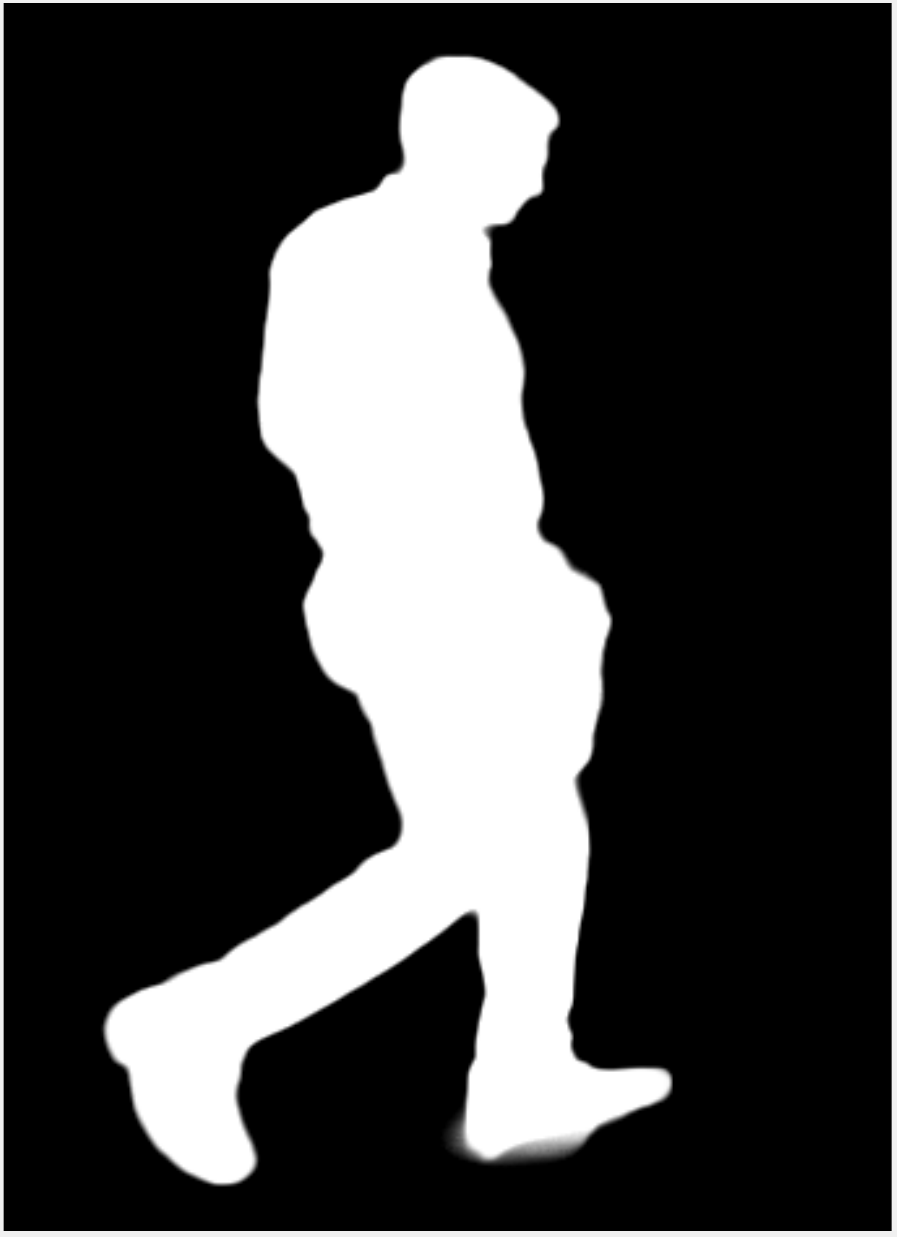}&
\hspace{-2ex}\includegraphics[width=0.3\linewidth]{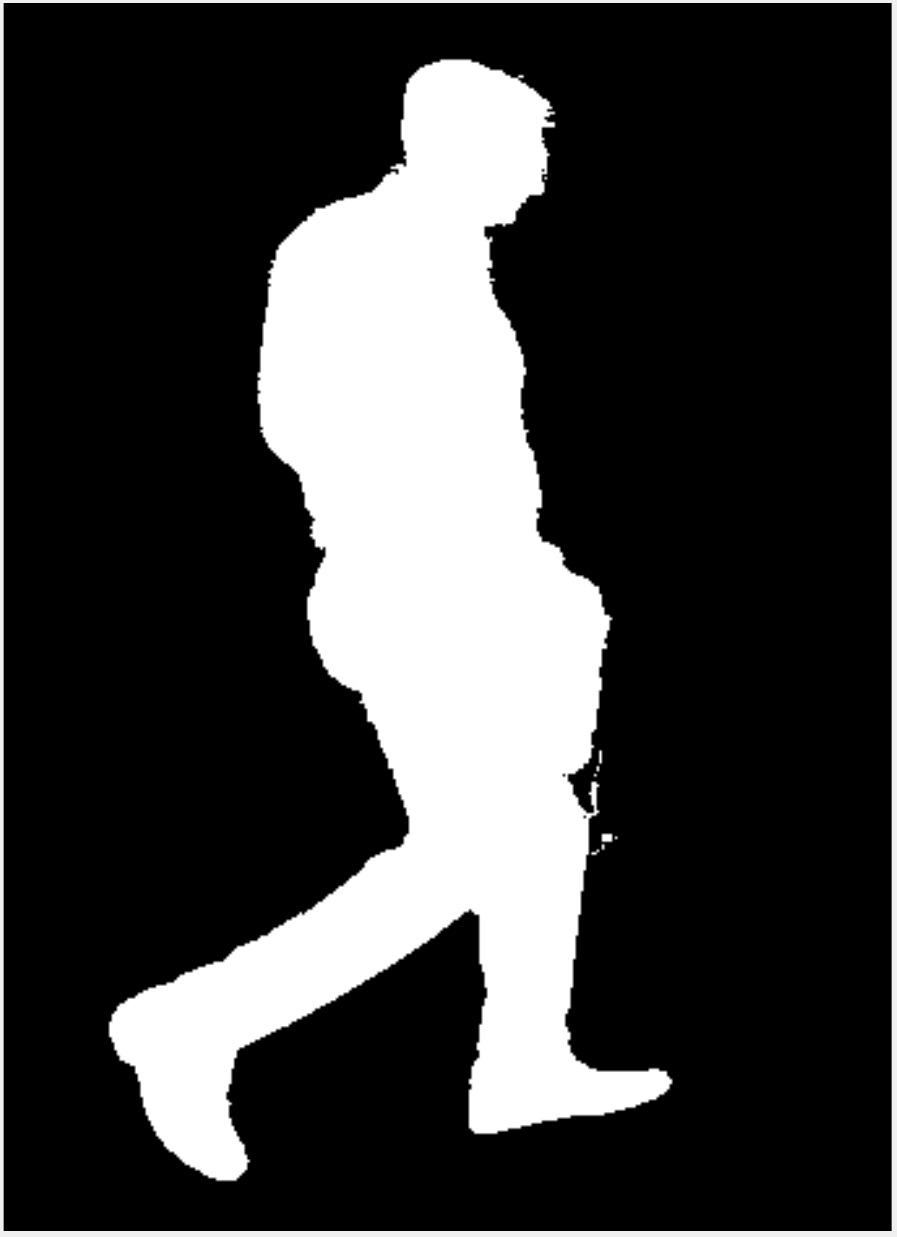}&
\hspace{-2ex}\includegraphics[width=0.3\linewidth]{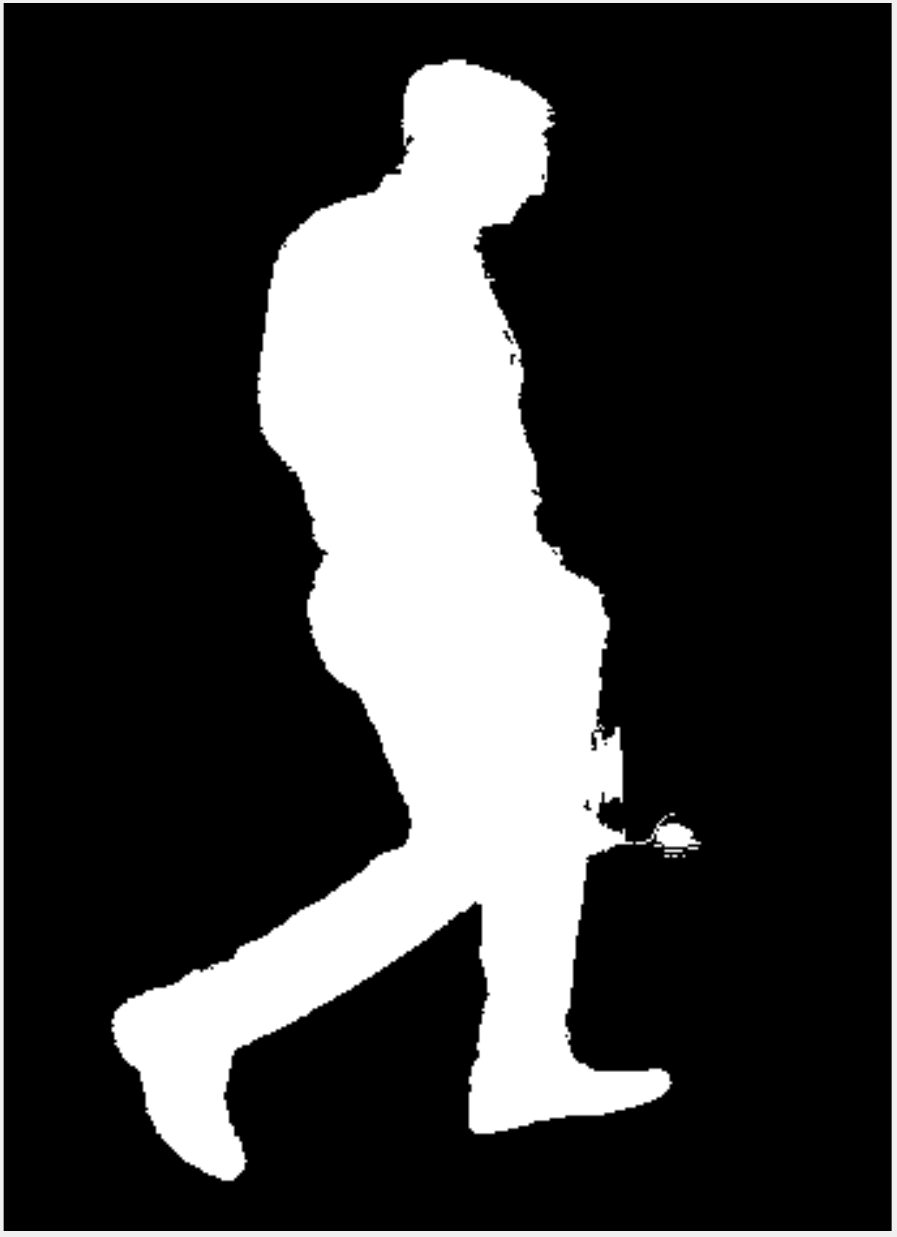}\\
\hspace{-2ex} \footnotesize{(d) Ours} & \footnotesize{(e) AGS \cite{Wang_Song_Zhao_2019}, after} & \footnotesize{(f) PDB \cite{Song_Wang_Zhao_2018}, after}\\
\end{tabular}
\caption{Dependency of conditional random field. (a) Input. (b) Raw output of the neural network part of AGS \cite{Wang_Song_Zhao_2019}. (c) Raw output of neural network part of PDB \cite{Song_Wang_Zhao_2018}. (d) Our result without post-processing. (e) Post-processing of AGS using conditional random field. (f) Post-processing of PDB using conditional random field. Notice the rough raw output of the deep neural network parts.}
\vspace{-4ex}
\label{fig:crf}
\end{figure}

We should also comment on the reliance on conditional random field of these deep learning solutions. In \fref{fig:crf} we show the raw outputs of AGS \cite{Wang_Song_Zhao_2019} and PDB \cite{Song_Wang_Zhao_2018}. While the salient object is correctly identified, the masks are coarse. Only after the conditional random field \cite{Krahenbuhl_Koltun_2011} the results become significantly better. In contrast, the raw output of our proposed algorithm is already high quality.

\begin{table*}[h]
\centering
\footnotesize{
\caption{Average results comparison with competing methods: AGS\cite{Wang_Song_Zhao_2019},PDB\cite{Song_Wang_Zhao_2018},MOA\cite{Dehghan_Zhang_Siam_2019}, NLVS \cite{Faktor_Irani_2014}, Trimap + DCNN \cite{Cho_Tai_Kweon_2016}, PBAS \cite{Hofmann_Tiefenbacher_Rigoll_2012}, ViBe \cite{Barnich_Marc_2011}, BSVS\cite{Marki_Perazzi_Wang_2016},   Grabcut\cite{Rother_Kolmogorov_Blake_2004}.  Higher intersection-over-union (IoU), higher Contour accuracy (F)\cite{Perazzi_Pont_Mcwilliams_2016}, higher Structure measure (S)\cite{Fan_Chen_Liu_2017}, lower MAE and lower Temporal instability (T)\cite{Perazzi_Pont_Mcwilliams_2016} indicate better performance.}
\renewcommand{\arraystretch}{1.8}
\begin{tabular}{c|c|cccc|c|cc|cc}
\hline
\hline
       &     & \multicolumn{4}{c|}{Unsupervised Video Segmentation} & Matting & \multicolumn{2}{c|}{Bkgnd Subtract.} & \multicolumn{2}{c}{Others} \\
Metric & Our & AGS~\cite{Wang_Song_Zhao_2019} &PDB~\cite{Song_Wang_Zhao_2018} &MOA~\cite{Dehghan_Zhang_Siam_2019} &NLVS~\cite{Faktor_Irani_2014} &Tmap~\cite{Cho_Tai_Kweon_2016} &PBAS~\cite{Hofmann_Tiefenbacher_Rigoll_2012} &ViBe~\cite{Barnich_Marc_2011}&BSVS~\cite{Marki_Perazzi_Wang_2016}  &Gcut~\cite{Rother_Kolmogorov_Blake_2004}    \\
& & CVPR '19 & ECCV '18 & ICRA '19 & BMVC '14 & ECCV '16 & CVPRW '12 & TIP '11 & CVPR '16 & ToG '04 \\
\hline
IoU     &\textbf{0.9321}	&0.8781	&0.8044	&0.7391	&0.5591	&0.7866	&0.5425	&0.6351	&0.8646	&0.6574\\
MAE	&\textbf{0.0058}	&0.0113	&0.0452	&0.0323	&0.0669	&0.0216	&0.0842	&0.0556	&0.0093 &0.0392\\
F	&\textbf{0.9443}	&0.9112	&0.8518	&0.7875	&0.6293	&0.7679	&0.6221	&0.5462	&0.8167 &0.6116\\
S	&\textbf{0.9672}	&0.938	&0.8867	&0.8581	&0.784	&0.9113	&0.7422	&0.8221	&0.9554 &0.8235\\
T &\textbf{0.165}	&0.1885	&0.2045	&0.1948	&0.229	&0.1852	&0.328	&0.2632	&0.2015 &0.232\\
\hline
\end{tabular}}
\vspace{1ex}
\label{table:full chart}
\end{table*}

\vspace{2ex}
\noindent $\bullet$ \textbf{Comparison with trimap + alpha-matting methods}: In this experiment we compare with several state-of-the-art alpha matting algorithms. The visual comparison is shown \fref{fig:failure example multi people}, and the performance of DCNN \cite{Cho_Tai_Kweon_2016} is shown in \tref{table:full chart}. In order to make this method work, careful tuning during the trimap generation stage is required.

\fref{fig:failure example multi people} and \tref{table:full chart} show that most alpha matting algorithms suffer from false alarms near the boundary, e.g., spectral matting \cite{Levin_Rav-Acha_Lischinski_2008},  closed-form mating \cite{Levin_Lischinski_Weiss_2008}, learning-based matting \cite{Zheng_Kambhamettu_2009} and comprehensive matting \cite{Shahrian_Rajan_Price_2013}. The more recent methods such as K-nearest neighbors matting \cite{Chen_Li_Tang_2013} and DCNN \cite{Cho_Tai_Kweon_2016} have equal amount of false alarm and miss. Yet, the overall performance is still worse than the proposed method and AGS. It is also worth noting that the matting approach achieves the second lowest T score (0.19), which is quite remarkable considering it is only a single-image method.

\vspace{2ex}
\noindent $\bullet$ \textbf{Comparison with background subtraction methods}: Background subtraction methods PBAS\cite{Hofmann_Tiefenbacher_Rigoll_2012} and ViBe\cite{Barnich_Marc_2011} are not able to obtain a score higher than 0.65 for IoU. Their MAE values are also significantly larger than the proposed method. Their temporal consistency is lagging by larger than 0.25 for T measure. Qualitatively, we observe that background subtraction methods perform most badly for scenes where the foreground objects are mostly stable or only have rotational movements. This is a common drawback of background subtraction algorithms, since they learn the background model in a online fashion and will gradually include non-moving objects into the background model. Without advanced design to ensure spatial and temporal consistency, the results also show errors even when the foreground objects are moving.

\vspace{2ex}
\noindent $\bullet$ \textbf{Comparison with other methods}: Semi-supervised BSVS \cite{Marki_Perazzi_Wang_2016} requires ground truth key frames to learn a model. After the model is generated, the algorithm will overwrite the key frames with the estimates. When conducting this experiment, we ensure that the key frames used to generate the model are not used during testing. The result of this experiment shows that despite the key frames, BSVS \cite{Marki_Perazzi_Wang_2016} still performs worse than the proposed method. It is particularly weak when the background is complex where the key frames fail to form a reliable model.

The modified Grabcut \cite{Rother_Kolmogorov_Blake_2004} uses the plate image as a guide for the segmentation. However, because of the lack of additional prior models the algorithm does not perform well. This is particularly evident in images where colors are similar between foreground and background. Overall, Grabcut scores badly in most metrics, only slightly better than the background subtraction methods.

\begin{figure*}[!]
\centering
\begin{tabular}{lllc}
(a) &                                                                                      & &\hspace{-2ex}\includegraphics[width=0.9\linewidth]{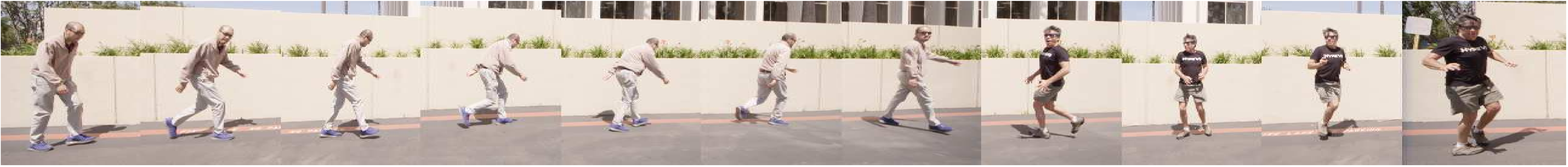}\\
(b) &\hspace{-3ex} \raisebox{1\normalbaselineskip}[0pt][0pt]{\rotatebox[origin=c]{90}{GT}} & &\hspace{-2ex}\includegraphics[width=0.9\linewidth]{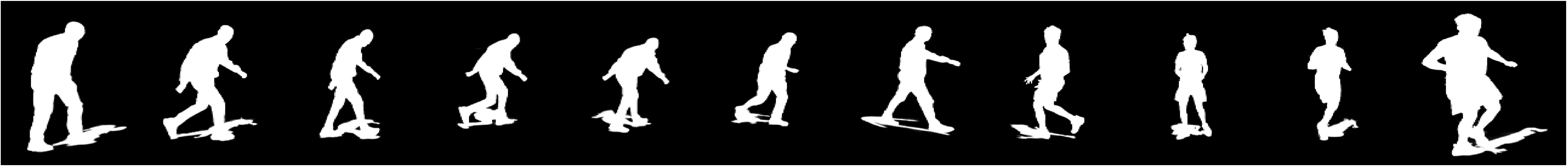}\\
(c) &\hspace{-2ex}\raisebox{1\normalbaselineskip}[0pt][0pt]{\rotatebox[origin=c]{90}{Our}} & &\hspace{-2.7ex} \includegraphics[width=0.9\linewidth]{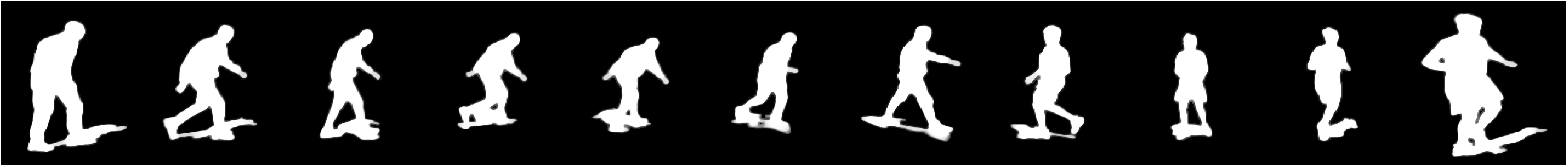}\\
(d) &\hspace{-2ex}\raisebox{1\normalbaselineskip}[0pt][0pt]{\rotatebox[origin=c]{90}{AGS}} &\hspace{-2ex}\raisebox{1\normalbaselineskip}[0pt][0pt]{\rotatebox[x=2mm,y=0mm]{90}{{\scriptsize  CVPR 2019}}}& \hspace{-2ex}\includegraphics[width=0.9\linewidth]{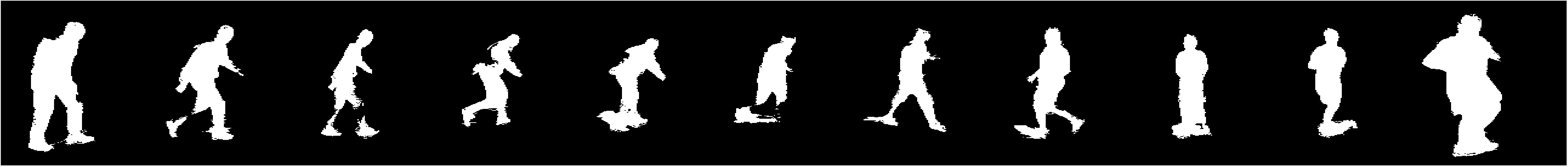}\\
(e) &\hspace{-2ex}\raisebox{1\normalbaselineskip}[0pt][0pt]{\rotatebox[origin=c]{90}{PDB}}&\hspace{-2ex}\raisebox{1\normalbaselineskip}[0pt][0pt]{\rotatebox[x=3mm,y=0mm]{90}{{\scriptsize ECCV 2018 }}}& \hspace{-2ex}\includegraphics[width=0.9\linewidth]{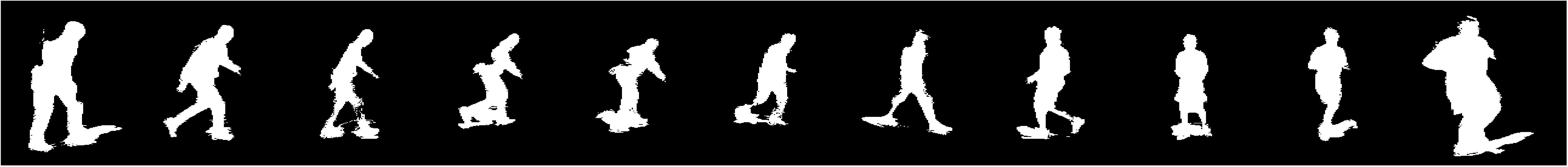}\\
(f) &\hspace{-2ex}\raisebox{1\normalbaselineskip}[0pt][0pt]{\rotatebox[origin=c]{90}{MOA}}&\hspace{-2ex}\raisebox{1\normalbaselineskip}[0pt][0pt]{\rotatebox[x=2mm,y=0mm]{90}{{\scriptsize ICRA 2019}}}&\hspace{-2.7ex} \includegraphics[width=0.9\linewidth]{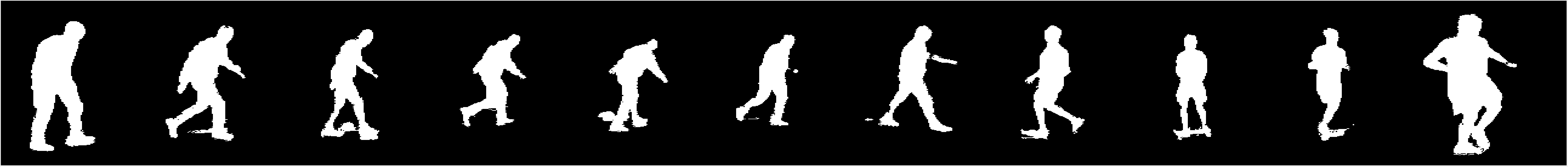}\\
(g) &\hspace{-2ex}\raisebox{1\normalbaselineskip}[0pt][0pt]{\rotatebox[origin=c]{90}{NLVS}}&\hspace{-2ex}\raisebox{1\normalbaselineskip}[0pt][0pt]{\rotatebox[x=3mm,y=0mm]{90}{{\scriptsize BMVC 2014 }}}&\hspace{-2.6ex} \includegraphics[width=0.9\linewidth]{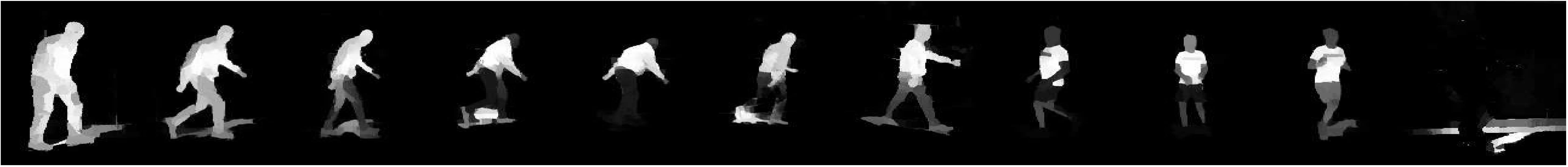}\\
(h) &\hspace{-2ex}\raisebox{1\normalbaselineskip}[0pt][0pt]{\rotatebox[origin=c]{90}{Tmap}}&\hspace{-2ex}\raisebox{1\normalbaselineskip}[0pt][0pt]{\rotatebox[x=2mm,y=0mm]{90}{{\scriptsize ECCV 2016}}}&\hspace{-2.5ex} \includegraphics[width=0.9\linewidth]{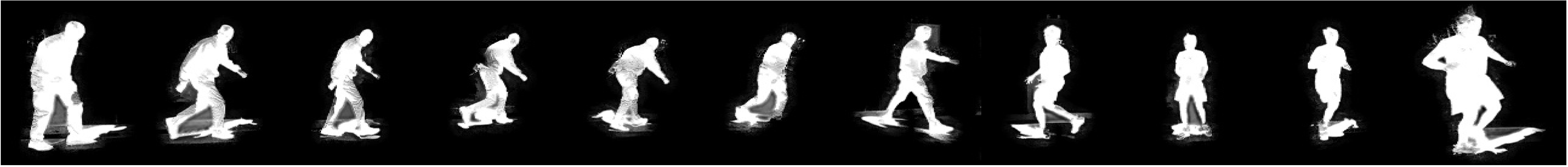}\\
(i) &\hspace{-2ex}\raisebox{1\normalbaselineskip}[0pt][0pt]{\rotatebox[origin=c]{90}{PBAS}}&\hspace{-2ex}\raisebox{1\normalbaselineskip}[0pt][0pt]{\rotatebox[x=3mm,y=0mm]{90}{{\scriptsize CVPRW 2012}}}&\hspace{-2.3ex} \includegraphics[width=0.9\linewidth]{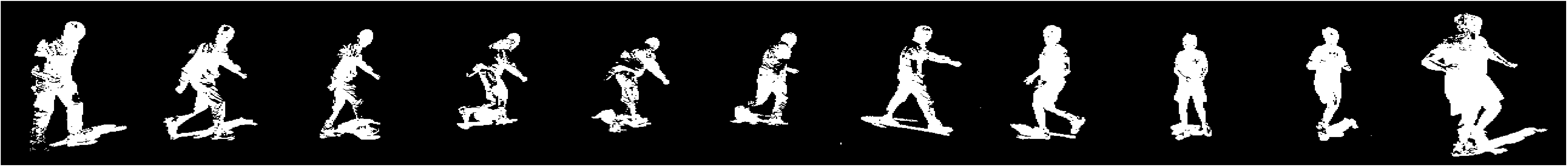}\\
(j) &\hspace{-2ex}\raisebox{1\normalbaselineskip}[0pt][0pt]{\rotatebox[origin=c]{90}{ViBe}}&\hspace{-2ex}\raisebox{1\normalbaselineskip}[0pt][0pt]{\rotatebox[x=2mm,y=0mm]{90}{{\scriptsize TIP 2011}}}&\hspace{-2ex} \includegraphics[width=0.9\linewidth]{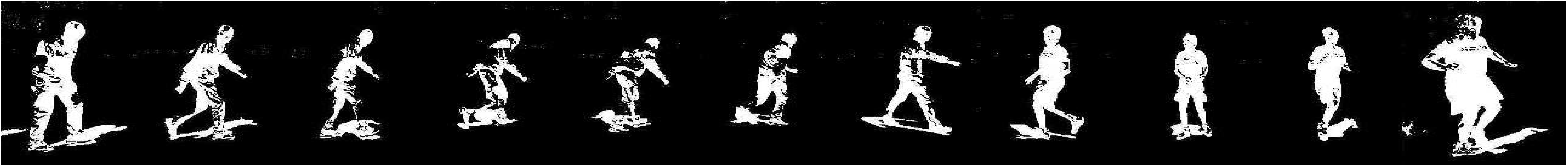}\\
(k) &\hspace{-2ex}\raisebox{1\normalbaselineskip}[0pt][0pt]{\rotatebox[origin=c]{90}{BSVS}}&\hspace{-2ex}\raisebox{1\normalbaselineskip}[0pt][0pt]{\rotatebox[x=3mm,y=0mm]{90}{{\scriptsize CVPR 2016}}}&\hspace{-2ex} \includegraphics[width=0.9\linewidth]{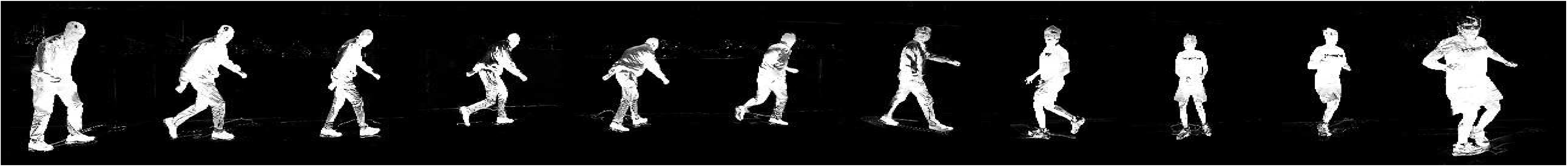}\\
(l) &\hspace{-2ex}\raisebox{1\normalbaselineskip}[0pt][0pt]{\rotatebox[origin=c]{90}{Gcut}}&\hspace{-2ex}\raisebox{1\normalbaselineskip}[0pt][0pt]{\rotatebox[x=4mm,y=0mm]{90}{{\scriptsize ACM ToG 2004}}}&\hspace{-2ex} \includegraphics[width=0.9\linewidth]{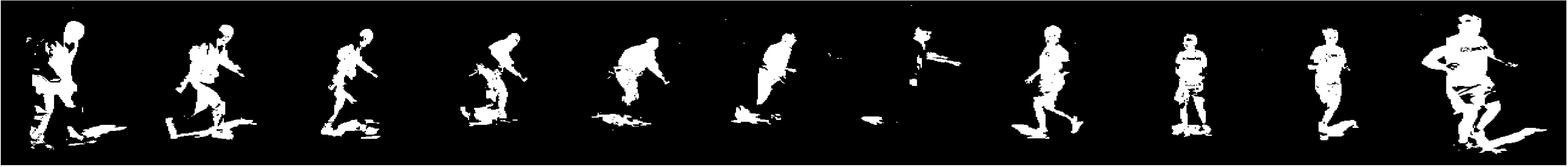}\\
\end{tabular}
\caption{Office sequence results. (a) Input. (b) Ground truth. (c) Ours. (d) AGS\cite{Wang_Song_Zhao_2019}. (e) PDB\cite{Song_Wang_Zhao_2018}. (f)MOA\cite{Dehghan_Zhang_Siam_2019}  (g)NLVS \cite{Faktor_Irani_2014}.  (h) Trimap + DCNN \cite{Cho_Tai_Kweon_2016}.  (i) PBAS \cite{Hofmann_Tiefenbacher_Rigoll_2012}. (j) ViBe \cite{Barnich_Marc_2011}. (k)BSVS \cite{Marki_Perazzi_Wang_2016}. (l)Gcut \cite{Rother_Kolmogorov_Blake_2004}. }
\label{fig:office sequence}
\end{figure*}

\subsection{Ablation study}

Since the proposed framework contains three different agents $F_1$, $F_2$ and $F_3$, we conduct an ablation study to verify the relative importance of the individual agents. To do so, we remove one of the three agents while keeping the other two fixed. The result is shown in \tref{table:variations}. For T score, results for w/o $F_1$ and w/o $F_3$ are omitted, as their results have many small regions of false alarms rendering untrackable amount of points on the polygon contours used in calculating T measure.

The matting agent $F_1$ has the most impact on the performance, followed by background estimator and denoiser. The drop in performance is most significant for hard sequences such as \texttt{Book} as it contains moving background, and \texttt{Road} as it contains strong color similarity between foreground and background. On average, we observe signiﬁcant drop in IoU from 0.93 to 0.72 when the matting agent is absent. The F measure decreases from 0.94 to 0.76 as the boundaries are more erroneous without the matting agent. The structure measure also degrades from 0.97 to 0.85. The amount of error in the results also cause the T measure to become untrackable.

In this ablation study, we also observe spikes of error for some scenes when $F_2$ is absent. This is because, without the $\valpha^{T}(1-\valpha)$ term in $F_2$, the result will look grayish instead of close-to-binary. This behavior leads to the error spikes. One thing worth noting is that the results obtained without $F_2$ do not drop significantly for S, F and T metric. This is due to the fact that IoU and MAE are pixel based metrics, whereas F, S and T are structural similarity. Therefore, even though the foreground becomes greyish without $F_2$, the structure of the labelled foreground is mostly intact.

For $F_3$, we observe that the total variation denoiser leads to the best performance for MACE. In a visual comparison shown in \fref{fig:different denoisers}, we observe that IRCNN\cite{Zhang_Zuo_Gu_2017} produces more detailed boundaries but fails to remove false alarms near the feet. BM3D\cite{Dabov_Foi_Katkovnik_2007} removes false alarms better but produces less detailed boundaries. TV on the other hand produces a more balanced result. As shown in \tref{table:variations}, BM3D performs similarly as IRCNN scoring similar values for most metrics except that IrCNN scores 0.93 in F measure with BM3D only scoring 0.77 meaning more accurate contours. In general, even with different denoisers, the proposed method still outperforms most competing methods.

\begin{table*}[h]
\centering
\caption{\textcolor{black}{Ablation study of the algorithm. We show the performance by eliminating one of the agents, and replacing the denoising agent with other denoisers. Higher intersection-over-union (IoU), higher Contour accuracy (F)\cite{Perazzi_Pont_Mcwilliams_2016}, higher Structure measure (S)\cite{Fan_Chen_Liu_2017}, lower MAE and lower Temporal instability (T)\cite{Perazzi_Pont_Mcwilliams_2016} indicate better performance.}}
\renewcommand{\arraystretch}{1.8}
\footnotesize{
\begin{tabular}{c|c|ccc|cc}
\hline
Metric & Our & w/o $F_1$ & w/o $F_2$ &w/o$F_3$  & BM3D & IrCNN\\
\hline
IoU & \textbf{0.9321}  &0.7161	&0.7529	&0.7775	&0.8533	&0.8585\\
MAE	&\textbf{0.0058}  & 0.0655	&0.0247	&0.0368	&0.0128	&0.0121\\
F	&\textbf{0.9443}  &0.7560	&0.9166	&0.6510	&0.7718	&0.8506\\
S	&\textbf{0.9672}  &0.8496	&0.9436	&0.8891	&0.9334	&0.9320\\
T	&\textbf{0.165} &too large	&0.1709	& too large &0.1911	&0.1817	\\
\hline
\end{tabular}}
\vspace{1ex}
\label{table:variations}
\end{table*}

\section{Limitations and Discussion}
\begin{figure}[h]
\centering
\begin{tabular}{cccc}
\hspace{-2ex}\includegraphics[width=0.23\linewidth]{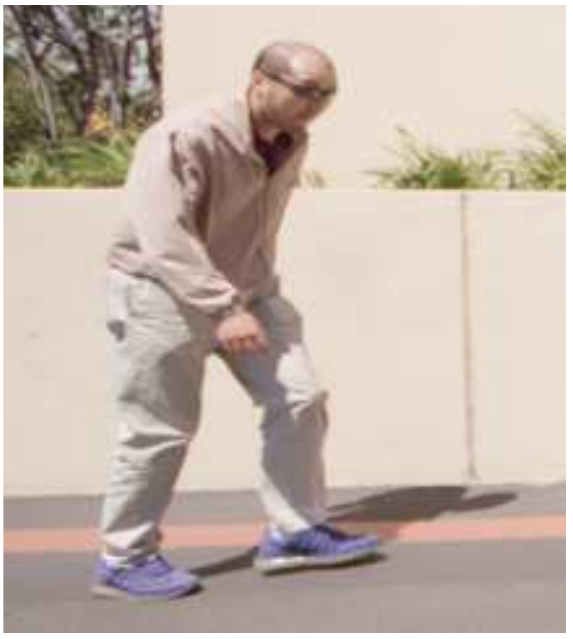}&
\hspace{-2ex}\includegraphics[width=0.23\linewidth]{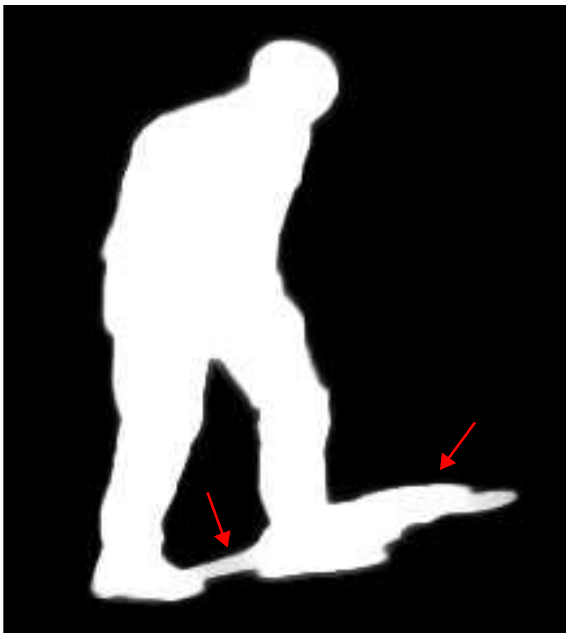}&
\hspace{-2ex}\includegraphics[width=0.22\linewidth]{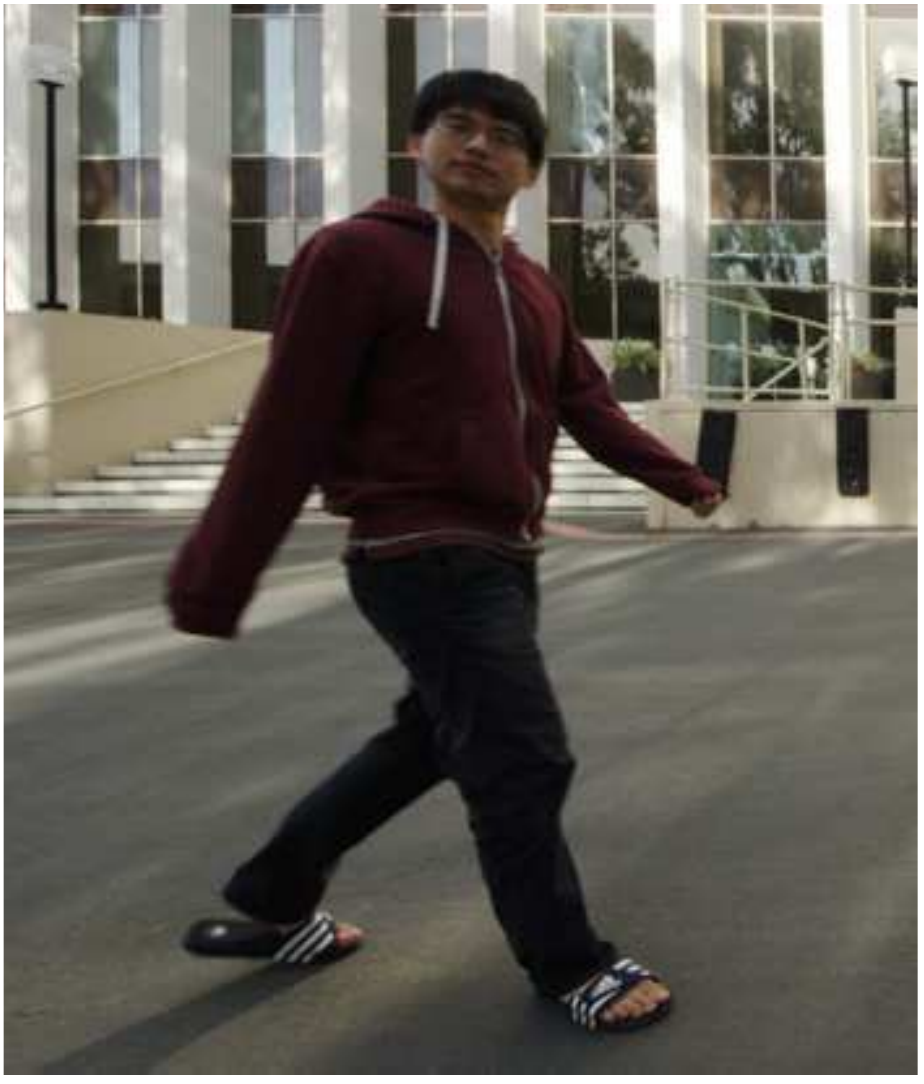}&
\hspace{-2ex}\includegraphics[width=0.22\linewidth]{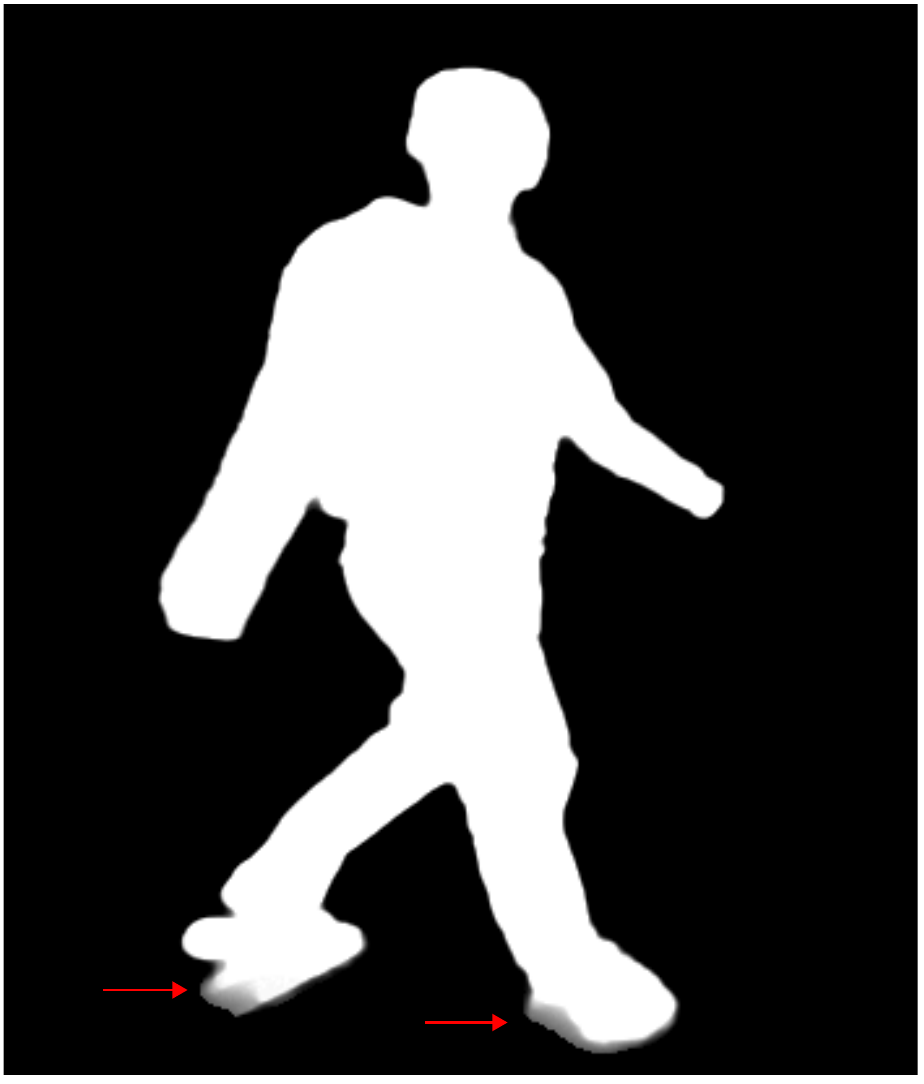}\\
\end{tabular}
\caption{Strong shadows. When shadows are strong, they are easily misclassified as foreground. }
\label{fig:shadow problems}
\end{figure}

While the proposed method demonstrates superior performance than the state-of-the-art methods, it also has several limitations.
\begin{itemize}[leftmargin=*]
\item \textbf{Quality of Plate Image}. The plate assumption may not hold when the background is moving substantially. When this happens, a more complex background model that includes dynamic information is needed. However, if the background is non-stationary, additional designs are needed to handle the local error and temporal consistency.
\item \textbf{Strong Shadows}. Strong shadows are sometimes treated as foreground, as shown in \fref{fig:shadow problems}. This is caused by the lack of shadow modeling in the problem formulation. The edge based initial estimate $\vr_{e}$ can resolve the shadow issue to some extent, but not when the shadow is very strong. We tested a few off-the-shelf shadow removal algorithms \cite{Guo_Dai_Hoiem_2011,Arbel_Hel-Or_2011,Liu_Gleicher_2008}, but generally they do not help because the shadow in our dataset can cast on the foreground object which should not be removed.
\end{itemize}

An open question here is whether our problem can be solved using deep neural networks since we have the plate. While this is certainly a feasible task because we can use the plate to replace the guided inputs (e.g., optical flow in \cite{Dehghan_Zhang_Siam_2019} or visual attention in \cite{Wang_Song_Zhao_2019}), an appropriate training dataset is needed. In contrast, the proposed method has the advantage that it is training-free. Therefore, it is less susceptible to issues such as overfit. We should also comment that the MACE framework allows us to use deep neural network solutions. For example, one can replace $F_1$ with a deep neural network, and $F_2$ with another deep neural network. MACE is guaranteed to find a fixed point of these two agents if they do not agree.

\vspace{-2ex}
\section{Conclusion}
This paper presents a new foreground extraction algorithm based on the multi-agent consensus equilibrium (MACE) framework. MACE is an information fusion framework which integrates multiple weak experts to produce a strong estimator. Equipped with three customized agents: a dual-layer closed form matting agent, a background estimation agent and a total variation denoising agent, MACE offers substantially better foreground masks than state-of-the-art algorithms. MACE is a fully automatic algorithm, meaning that human interventions are not required. This provides significant advantage over semi-supervised methods which require trimaps or scribbles. In the current form, MACE is able to handle minor variations in the background plate image, illumination changes and weak shadows. Extreme cases can still cause MACE to fail, e.g., background movement or strong shadows. However, these could potentially be overcome by improving the background and shadow models.

\section{Appendix}
\subsection{Proof of Theorem 2}
\begin{proof}
We start by writing \eref{eq:cfm_obj_2} in the matrix form
\begin{align*}
\widetilde{J}(\valpha^{\mI},\valpha^{\mP},\va,\vb)=\sum_{k \in \mI}\left|\left|
\begin{bmatrix}
\mH_{k} &\vec{1}\\
\sqrt{\eta}\mG_{k} &\sqrt{\eta}\vec{1}\\
\sqrt{\epsilon}\mI_{3\times 3} &\vec{0}
\end{bmatrix}
\begin{bmatrix}
{\va}_{k}\\
b_{k}
\end{bmatrix}
-
\begin{bmatrix}
{\valpha}_{k}^{\mI}\\
{\valpha}_{k}^{\mP}\\
\vec{0}
\end{bmatrix}
\right|\right|^{2}
\end{align*}
where
\begin{align*}
&\mH_{k}=\begin{bmatrix}
\vdots &\vdots &\vdots \\
I^{r}_{i} &I^{g}_{i} &I^{b}_{i} \\
\vdots &\vdots &\vdots \\
\end{bmatrix},
\quad
\mG_{k}=
\begin{bmatrix}
\vdots &\vdots &\vdots \\
P^{r}_{i} &P^{g}_{i} &P^{b}_{i} \\
\vdots &\vdots &\vdots \\
\end{bmatrix},\\
&{\va}_{k}=
\begin{bmatrix}
a_{k}^{r}\\
a_{k}^{g}\\
a_{k}^{b}
\end{bmatrix},
\quad
{\valpha}_{k}^{\mI}=
\begin{bmatrix}
\vdots\\
\alpha_{i}^{\mI}\\
\vdots
\end{bmatrix},
\quad
{\valpha}_{k}^{\mP}=
\begin{bmatrix}
\vdots\\
\alpha_{i}^{\mP}\\
\vdots
\end{bmatrix},
\end{align*}
and $i$ denotes the index of the $i$-th pixel in the neighborhood $w_k$. The difference with the classic closed-form matting \cite{Levin_Lischinski_Weiss_2008} is the new terms $\mG_k$, $\vone$ and $\valpha_k^{\mP}$ (i.e., the second row of the quadratic function above.)

Denote
\begin{equation}
\mB_k \bydef \begin{bmatrix}
\mH_{k} &\vec{1}\\
\sqrt{\eta}\mG_{k} &\sqrt{\eta}\vec{1}\\
\sqrt{\epsilon}\mI_{3 \times 3} &\vec{0}
\end{bmatrix},
\end{equation}
and use the fact that $\valpha^{\mP}=\vec{0}$, we can find out the solution of the least-squares optimization:
\begin{align}
\begin{bmatrix}
{\va}_{k}\\
b_{k}
\end{bmatrix}
=
(\mB_{k}^{T}\mB_{k})^{-1}\mB_{k}^{T}
\begin{bmatrix}
{\valpha}_{k}^{\mI}\\
\vec{0}\\
\vec{0}
\end{bmatrix}
\label{eq: a_k}
\end{align}
We now need to simplify the term $\mB_k^T\mB_k$. First, observe that
\begin{align*}
\mB_{k}^{T}\mB_{k}
&=\begin{bmatrix}
\mH_{k}^{T}\mH_{k}+\eta\mG_{k}^{T}\mG_{k}+\epsilon\mI_{3 \times 3} &\mH_{k}^{T}\vec{1}+\eta\mG_{k}^{T}\vec{1}\\
(\mH_{k}\vec{1}+\eta\mG_{k}^{T}\vec{1})^{T} &n(1+\eta)
\end{bmatrix}\\
&= \begin{bmatrix}
\mSigma_{k}                             &                  \vmu_{k}\\
\vmu_{k}^{T}                            &                  c
\end{bmatrix}
\end{align*}
where we define the terms $\mSigma_{k}\bydef \mH_{k}^{T}\mH_{k}+\eta\mG_{k}^{T}\mG_{k}+\epsilon\mI$, $\vmu_{k} \bydef \mH_{k}^{T}\vec{1}+\eta\mG_{k}^{T}\vec{1}$ and $c \bydef n(1+\eta)$. Then, by applying the block inverse identity, we have
\begin{align}
(\mB_{k}^{T}\mB_{k})^{-1}=
\begin{bmatrix}
\mT_{k}^{-1} &-\mT_{k}^{-1}\widehat{\vmu}_{k}\\
-(\mT_{k}^{-1}\widehat{\vmu}_{k})^{T} &\frac{1}{c}+\widehat{\vmu}_{k}\mT_{k}^{T}\widehat{\vmu}_{k}
\end{bmatrix}
\label{eq: inv(B'B)}
\end{align}
where we further define $\mT_{k}=\mSigma_{k}-\frac{\vmu_{k}\vmu_{k}^{T}}{c}$ and $\widehat{\vmu}_{k}=\frac{\vmu_{k}}{c}$.

Substituting \eref{eq: a_k} back to $\Jtilde$, and using \eref{eq: inv(B'B)}, we have
\begin{align*}
\Jtilde(\valpha^{\mI})
&=\sum_{k}\left|\left|(\mI_{3 \times 3}-\mB_{k}(\mB_{k}^{T}\mB_{k})^{-1}\mB_{k}^{T})
\begin{bmatrix}
{\valpha}_{k}^{\mI}\\
\vec{0}\\
\vec{0}
\end{bmatrix}\right|\right|^{2}\\
&= (\valpha_{k}^{\mI})^{T}\mL_{k}{\valpha}_{k}^{\mI},
\end{align*}
where
\begin{align}
\mL_{k} &= \mI_{3\times 3}-\bigg(\mH_{k}\mT_{K}^{-1}\mH_{k}^{T}-\mH_{k}\mT_{k}^{-1}\widehat{\vmu}_{k}\vec{1}^{T} \notag\\
&-\vec{1}^{T}(\mT_{k}^{-1}\widehat{\vmu}_{k})^{T}\mH_{k}+\frac{1}{c}\vec{1}^{T}\widehat{\vmu}_{k}\mT_{k}^{-1}\widehat{\vmu}_{k}\vec{1} \bigg)
\end{align}
The $(i,j)$-th element of $\mL_k$ is therefore
\begin{align}
\mL_{k}(i,j)=&\delta_{ij}-(\mI_{ki}^{T}\mT_{k}^{-1}\mI_{kj}-\mI_{ki}^{T}\mT_{k}^{-1}\widehat{\vmu}_{k} \notag \\
&-\widehat{\vmu}_{k}^{T}\mT_{k}^{-1}\mI_{kj}+\frac{1}{c}+\widehat{\vmu}_{k}^{T}\mT_{k}^{-1}\widehat{\vmu}^{k})\notag \\
=&\delta_{ij}-(\frac{1}{c}+(\mI_{ki}-\widehat{\vmu}_{k})^{T}\mT_{k}^{-1}(\mI_{kj}-\widehat{\vmu}_{k}))
\end{align}
Adding terms in each $w_k$, we finally obtain
\begin{align*}
\widetilde{L}_{i,j} &= \sum_{k | (i,j) \in w_{k}} \bigg\{ \delta_{ij}-(\frac{1}{c}+(\mI_{ki}-\widehat{\vmu}_{k})^{T}\mT_{k}^{-1}(\mI_{kj}-\widehat{\vmu}_{k})) \bigg\}.
\end{align*}

\end{proof}

\subsection{Proof: $\mLtilde$ is positive definite}
\begin{proof}
Recall the definition of  $\widetilde{J}(\valpha^I,\valpha^P,\va,\vb)$:
\begin{align*}
\widetilde{J}(\valpha^I,\valpha^P,\va,\vb)=\sum_{j\in I} \Bigg\{ \sum_{i\in w_{j}} \left(\alpha_{i}^I-\sum_{c} a^{c}_{j}I^{c}_{i}-b_{j} \right)^{2} \notag \\
+ \eta \sum_{i\in w_{j}}\left(\alpha_{i}^P-\sum_{c} a^{c}_{j}P^{c}_{i}-b_{j} \right)^{2} +\epsilon\sum_{c}(a^{c}_{j})^{2} \Bigg\}
\end{align*}
Based on Theorem 2 we have,
\begin{align}
\widetilde{J}(\valpha) \bydef \min_{\va,\vb}\; \widetilde{J}(\valpha,\vzero,\va,\vb)=\valpha^{T}\widetilde{\mL}\valpha.
\end{align}

We consider two cases: (i) $a^{c}_{j}=0$ $\forall j$ and $\forall c$, (ii) there exists some $j$ and $c$ such that $a^{c}_{j}\neq 0$. For the second case, $\Jtilde$ is larger than $0$.
For the first case, $\Jtilde$ can be reduced into
\begin{align}
\widetilde{J}(\valpha,0,\va,\vb)=\sum_{j\in I} \Bigg\{ \sum_{i\in w_{j}} \left(\left(\alpha_{i}-b_{j} \right)^{2}
+ \eta \left(-b_{j} \right)^{2} \right)  \Bigg\}\label{eq:2}
\end{align}
For any vector $\alpha\neq 0$, there exists at least one $\alpha_{i} \neq0$. Then by completing squares we can show that
\begin{align*}
&(\alpha_{i}-b_{j})^{2}+\eta b_{j}^{2}\\
&=\alpha_{i}^{2}-2\alpha_{i}b_{j}+(1+\eta)b_{j}^{2}\\
&=\left( \sqrt{\frac{1}{1+\eta}}\alpha_{i}-\sqrt{1+\eta}b_{j}\right)^{2}+\frac{\eta}{1+\eta}\alpha_{i}^{2}>0
\end{align*}
Therefore, $\widetilde{J}(\valpha,0,\va,\vb)>0$ for any non-zero vector $\alpha$. As a result, $\widetilde{J}(\valpha,0,\va,\vb)=\valpha^{T}\widetilde{\mL}\valpha>0$ for both cases, and $\widetilde{\mL}$ is positive definite.
\end{proof}

\subsection{Proof of Proposition 1}
\begin{proof}
Let $\underline{\vx} \in \R^{nN}$ and $\underline{\vy} \in \R^{nN}$ be two super-vectors.

\vspace{2ex}
\noindent(i). If the $F_{i}$'s are non-expansive, then
\begin{align*}
&\|\calF(\underline{\vx})-\calF(\underline{\vy})\|^{2}+\|\underline{\vx}-\underline{\vy}-(\calF(\underline{\vx})-\calF(\underline{\vy}))\|^{2}\\
&=\sum_{i=1}^{N}\left(\|F_i(\vx_i)-F_i(\vx_i)\|^{2}+\|\vx_i-\vy_i-(F_i(\vx_i)-F_i(\vy_i))\|^{2}\right)\\
&\stackrel{(c)}{\leq} \sum_{i=1}^{N} \|\vx_i-\vy_i\|^{2}=\|\underline{\vx}-\underline{\vy}\|^{2}
\end{align*}
where $(c)$ holds because each $F_i$ is firmly non-expansive. As a result, $\calF$ is also firmly non-expansive.

\vspace{2ex}
\noindent(ii). To prove that $\calG$ is firmly non-expansive, we recall from Theorem 1 that $2\calG-\calI$ is self-inverse. Since $\calG$ is linear, it has a matrix representation. Thus, $\|(2\calG-\calI)\vx\|^2 = \vx^T (2\calG-\calI)^T (2\calG-\calI) \vx$. Because $\calG$ is an averaging operator, it has to be symmetric, and hence $\calG^T = \calG$. As a result, we have $\|(2\calG-\calI)\vx\|^2 = \|\vx\|^2$ for any $\vx$, which implies non-expansiveness.

\vspace{2ex}
\noindent(iii). If $\calF$ and $\calG$ are both firmly non-expansive, we have
\begin{align*}
&\|(2\calG-\calI)[(2\calF-\calI)(\underline{\vx})]-(2\calG-\calI)[(2\calF-\calI)(\underline{\vy})]\|^{2}\\
&\stackrel{(a)}{\leq}\|(2\calF-\calI)(\underline{\vx})-(2\calF-\calI)(\underline{\vy})\|^{2} \stackrel{(b)}{\leq}\|\underline{\vx}-\underline{\vy}\|^{2}
\end{align*}
where $(a)$ is true due to the firmly non-expansiveness of $\calG$ and $(b)$ is true due to the non-expansiveness of $\calG$. Thus, $\calT \bydef (2\calG-\calI)(2\calF-\calI)$ is non-expansive. This result also implies convergence of the MACE algorithm, due to \cite{Buzzard_Chan_Sreehari_2017}.
\end{proof}

\bibliographystyle{IEEEbib}
\bibliography{refs}
\end{document}